\documentclass{article}

\usepackage[preprint, nonatbib]{neurips_2026}

\usepackage[utf8]{inputenc} 
\usepackage[T1]{fontenc}    
\usepackage{hyperref}       
\usepackage{url}            
\usepackage{booktabs}       
\usepackage{multirow}
\usepackage{multicol}
\usepackage{makecell}
\usepackage{nicefrac}       
\usepackage{microtype}      
\usepackage{xcolor}         
\usepackage{amsmath}
\usepackage{amsthm}
\usepackage{amssymb}
\usepackage{amsfonts}
\usepackage{caption}
\usepackage{subcaption}
\usepackage{algorithm}
\usepackage{longtable}

\usepackage{booktabs}
\usepackage[
    textcolor=red,
    linecolor=red,
    bordercolor=red,
    backgroundcolor=white,
]{todonotes}
\usepackage{algpseudocode}
\usepackage[table]{xcolor}
\usepackage{cleveref}
\usepackage{mathtools}
\usepackage{wrapfig}

\usepackage{tikz}
\usetikzlibrary{arrows.meta, positioning, calc, fit}
\usepackage{pgfplots}

\definecolor{plantblue}{RGB}{35,92,170}
\definecolor{ctrlgold}{RGB}{176,118,0}
\definecolor{memgreen}{RGB}{42,135,66}
\definecolor{softgray}{RGB}{245,245,245}
\definecolor{midgray}{RGB}{150,150,150}

\tikzset{
  >={Latex[length=2.8mm]},
  base/.style={
    rounded corners=3pt,
    draw,
    very thick,
    align=center,
    inner sep=6pt,
    font=\small
  },
  sidebox/.style={
    base,
    draw=black!55,
    fill=softgray,
    text=black,
    minimum height=2.55cm
  },
  plantbox/.style={
    base,
    draw=plantblue,
    fill=plantblue!8,
    text=plantblue
  },
  ctrlbox/.style={
    base,
    draw=ctrlgold,
    fill=ctrlgold!10,
    text=ctrlgold!75!black
  },
  membox/.style={
    base,
    draw=memgreen,
    fill=memgreen!8,
    text=memgreen!70!black
  },
  groupframe/.style={
    rounded corners=4pt,
    draw=plantblue!45,
    dashed,
    line width=0.9pt,
    inner sep=10pt
  },
  dataarrow/.style={draw=black!85, line width=1.0pt, -{Latex[length=2.7mm]}},
  yarrow/.style={draw=plantblue, line width=1.2pt, -{Latex[length=2.7mm]}},
  uarrow/.style={draw=ctrlgold!85!black, line width=1.2pt, -{Latex[length=2.7mm]}},
  marrow/.style={draw=memgreen!85!black, line width=1.2pt, -{Latex[length=2.7mm]}},
  metaarrow/.style={draw=midgray!90, dashed, line width=1.0pt, -{Latex[length=2.7mm]}},
  regarrow/.style={draw=memgreen!75!black, dashed, line width=0.95pt, -{Latex[length=2.7mm]}},
  stepnum/.style={
    circle,
    fill=black!85,
    text=white,
    font=\bfseries\small,
    minimum size=5.8mm,
    inner sep=0pt
  }
}

\newtheorem{theorem}{Theorem}[section]

\newtheorem{proposition}{Proposition}[section]

\newtheorem{remark}{Remark}[section]

\usepackage{pifont}      
\newcommand{\cmark}{\ding{51}}  
\newcommand{\xmark}{\ding{55}}  
\newcommand{\pmark}{\ding{108}} 

\providecommand{\best}[1]{\textbf{#1}}
\providecommand{\second}[1]{\underline{#1}}

\title{When Descent Is Too Stable: Event-Triggered Hamiltonian Learning to Optimize}

\author{%
Yi Wang{$^1$} \\
\texttt{panzer.wy@utexas.edu}
\\~\\
$^1${Oden Institute} \\
The University of Texas at Austin \\Austin, Texas 78712 
 \And
{Chandrajit Bajaj}{$^{1,2}$} \\\texttt{bajaj@cs.utexas.edu} \\~\\
$^2${Department of Computer Science} \\ The University of Texas at Austin \\Austin, Texas 78712
}

\begin{document}

\maketitle

\begin{abstract}
Fixed-budget nonconvex optimization can fail not because local descent is unstable, but because it is too stable: after reaching a nearby stationary point, an optimizer may spend the remaining evaluations refining an uninformative local minimum. We formulate this failure mode as a control problem over optimizer dynamics, where the learner must decide when to descend, when to exploit a promising basin, and when stagnation should trigger movement elsewhere. We introduce SHAPE, a structured adaptive port-Hamiltonian task-family optimizer for event-triggered minima hunting under local information. Starting from gradient-descent dynamics, SHAPE lifts optimization to an augmented phase space $(q, p)$, where the primal state $q$ represents the candidate solution, the cotangent variable $p$ carries directional sensitivity, and a controller $u$ provides processed information from current gradient oracle. Within each stage, a learned Hamiltonian vector field induces structured local descent; across stages, a fixed event clock in the implementation updates ports and memory when local equilibria are detected, with stage-dependent horizons treated in the analysis as a direct generalization. This design preserves a passivity-compatible structure while allowing the same trained policy to use clean, stochastic, or estimated gradient inputs. Experiments on fixed-budget nonconvex optimization tasks show that SHAPE improves best-so-far performance compared with fixed-policy optimizers. These results suggest that adaptive Hamiltonian energy shaping provides a principled mechanism for balancing descent, exploration, and budget allocation in difficult optimization landscapes.
\end{abstract}

\section{Introduction}
\label{sec:intro}

We study fixed-budget minimization problems of the form
\begin{equation}
q^* \in \operatorname*{arg\,min}_{q\in\mathcal Q} f(q),
\tag{P}
\label{eq:minimization:erm}
\end{equation}
where $q\in\mathcal Q\subset\mathbb R^d$ is the decision variable and
$f:\mathcal Q\to\mathbb R$ is the objective observed only through an oracle.
A first-order optimizer may be written abstractly as
\(
    q_{k+1}=\mathcal G(q_k,g(q_k);\psi),
\)
where $g(q_k)$ is the acquired local force, equal to $\nabla f(q_k)$ in the
clean-gradient case, and $\psi$ denotes either hand-tuned hyperparameter or
learned update-rule parameters. Descent update rule $\mathcal{G}$ is proved to be stable and can find the desired global minima if $f$ is convex or strongly convex \cite{nocedal2006numerical, polyak1964some, karimi2016linear}. Nevertheless, such a form hides a central difficulty of
nonconvex optimization under a limited budget: local descent can be \emph{too
stable}.  Once the update reaches a nearby attractive critical point, the
remaining evaluations may be spent refining a basin that is irrelevant to the
best solution available under the same budget. This failure mode suggests a navigation view of optimization. A fixed gradient descent scheme is not able to decide when local descent is useful, when momentum should be dissipated,
when a stable basin should be recorded, and when the next stage should be
redirected elsewhere.  Classical adaptive optimizers such as momentum~\cite{polyak1964some}, NAG~\cite{nesterov1983method},
 RMSProp~\cite{tieleman2012lecture}, Adam~\cite{kingma2014adam} and related variance-reduction schemes \cite{johnson2013svrg} provide important
first-order mechanisms that adapt to online gradient descent scenario, but their deployment-time update laws are still largely fixed. Thus, a learning-to-optimize pipeline is necessary if one can learn from repetitive past descent records and enrich the search space that can traverse in the target potential landscape.

\paragraph{SHAPE: Structured Hamiltonian Adaptive Port Evaluation}
Notably, gradient-based methods have a  continuous-time interpretation as
dynamical systems \cite{jordan2018dynamical}. This observation motivates us to propose SHAPE, a learnable controlled Hamiltonian system that navigate on target energy landscape. SHAPE lifts the optimizer state from $q\in\mathcal Q$ to
$x=(q,p)\in T^*\mathcal Q$.  The momentum $p$ is a covector conjugate to $q$,
so the optimizer evolves in a phase space rather than by a primal update alone.
At event stage $s$, memory $m_s$ and a local anchor $\bar q_s$ define a shaped
potential
\[
    U^{\rm shp}_s(q)
    =
    U_\eta(q;m_s)
    +\frac{\kappa_s}{2}\|q-\bar q_s\|^2
    +V_{{\rm bar},s}(q;m_s,\mu_s),
\]
and the local Hamiltonian is
\(
    H_{s,k}(q,p)=f(q)+U^{\rm shp}_s(q)+\frac12p^\top M_k^{-1}p .
\)
The memory term summarizes previously visited basins, the quadratic term gives
a local stage anchor, and the optional barrier term discourages repeated
refinement of excluded regions.  A learned port-Hamiltonian controller chooses
the skew channel, dissipation, metric, and bounded port input.  The port input
is written
$u^{\rm port}=u^{\rm shp}-K^d y$: $u^{\rm shp}$ is the active shaping or escape
input, while $K^d y$ is passive damping injection through the power-conjugate
output $y$.
\begin{figure}[t]
    \centering
    \begin{subfigure}{0.42\textwidth}
    \centering
    \includegraphics[width=\linewidth]{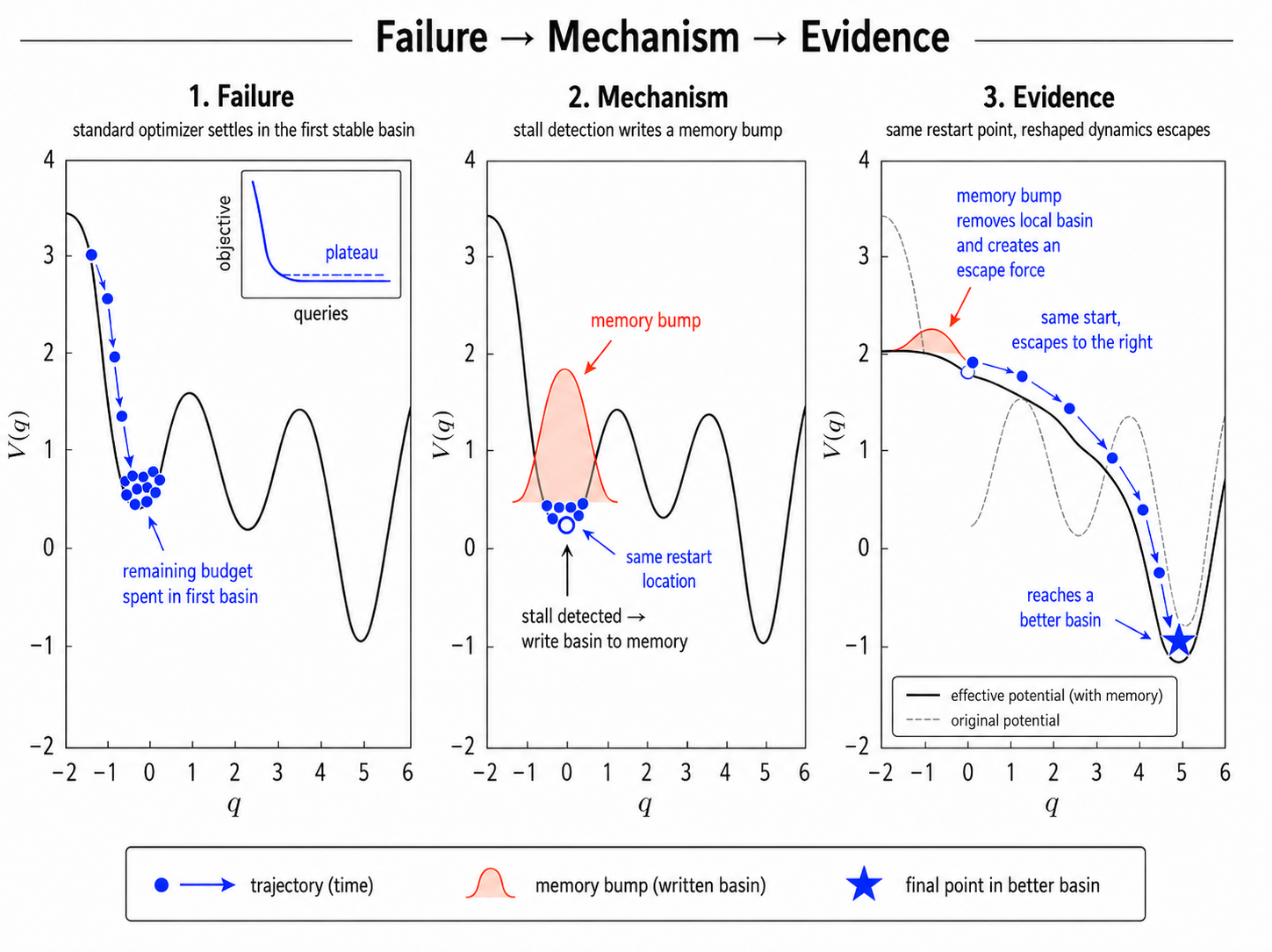}
    \caption{}
    \label{fig:teaser:left}
    \end{subfigure}
    \begin{subfigure}{0.54\textwidth}
    \centering
    \includegraphics[width=\linewidth]{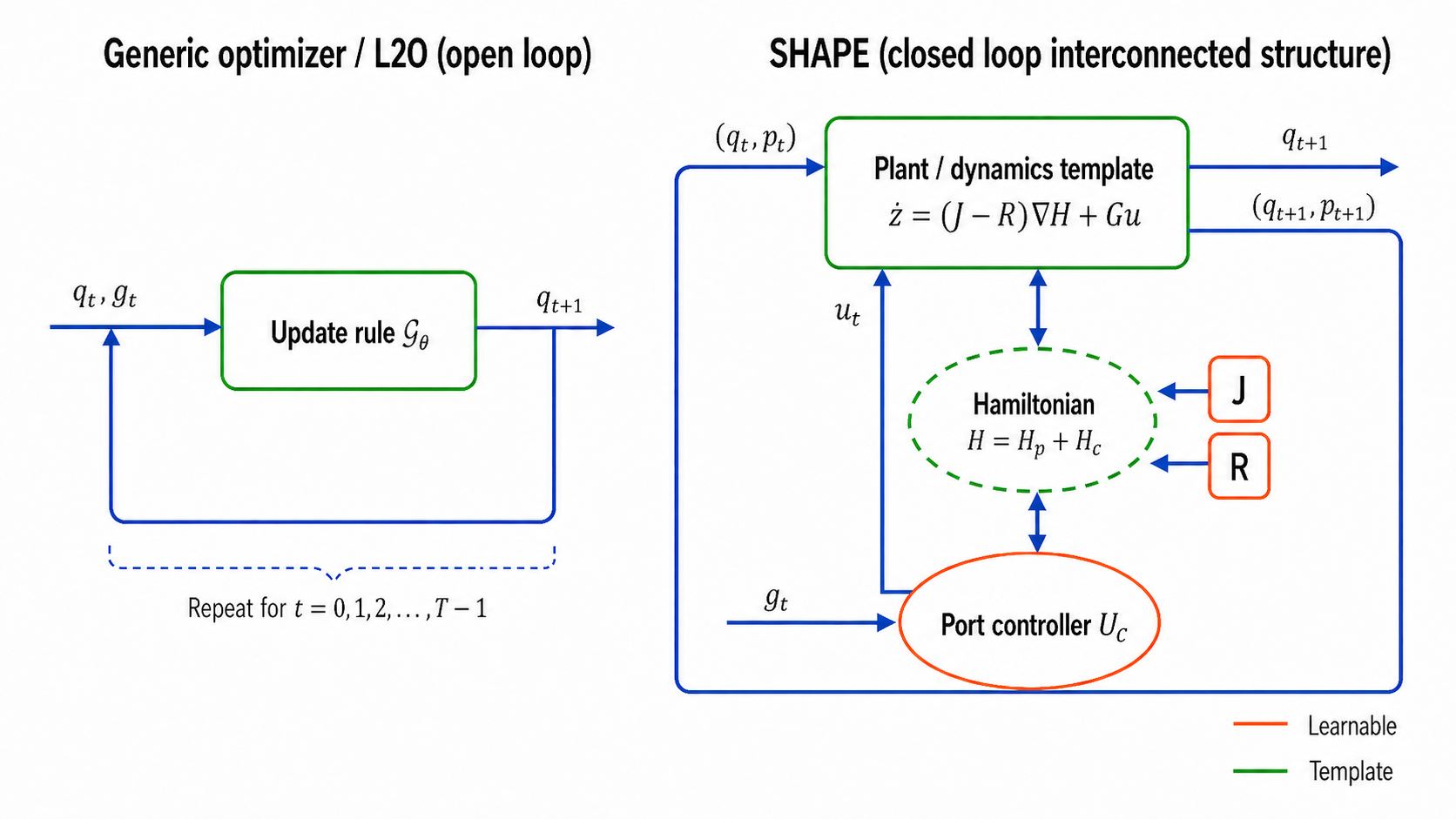}
    \caption{}
    \label{fig:teaser:right}
    \end{subfigure}
    \caption{(a) A local descent method may terminate at the first stable critical point encountered from a short-sighted trajectory.  SHAPE records such events in memory and uses the resulting shaped energy to discourage repeated refinement of already explored basins.  (b) A generic learned optimizer maps local oracle information directly to the next iterate, whereas SHAPE implements a closed-loop port-Hamiltonian interconnection.  The plant state $(q_k,p_k)$ evolves under the shaped Hamiltonian \eqref{eq:shape_local_hamiltonian}; the learned controller selects the structured operator $\mathcal A_{s,k}$ in \eqref{eq:shape_ph_vector_field}, damping/interconnection gains, and bounded port input.}
    \label{fig:teaser}
\end{figure}
The architecture separates two time scales.  Within a stage, the frozen shaped
Hamiltonian induces dissipative phase-space transport.  Across stages, an
event interface updates memory, mode, anchor, and budget.  This differs from an
open-loop learned optimizer $x_{k+1}=\mathcal G(x_k, g(q_k);\psi)$:
SHAPE couples a plant and controller through power-conjugate ports, as shown in Figure \ref{fig:teaser}. An ideal dynamic interconnection can be written as
\[
\dot x_p=(J_p-R_p)\nabla H_p+G_pu_p,
\quad 
\dot x_c=(J_c-R_c)\nabla H_c+G_cu_c,
\quad 
(u_p,u_c)=(-y_c,y_p),
\]
where
\(
    y_p=G_p^\top\nabla H_p,~
    y_c=G_c^\top\nabla H_c .
\)
This interconnection is power-preserving before damping injection.  In the implemented optimizer, we use the corresponding feedback-reduced form: the learned controller and memory observe the plant port output and return a bounded port input, damping injection, and shaping terms.  Thus the learned update acts through a structured port channel rather than through an unconstrained coordinate update.

\paragraph{Contributions.}
Our contributions are threefold.
(1) We formulate fixed-budget nonconvex optimization as an event-triggered
task-family minima hunter on $T^*\mathcal Q$, with a unified shaped potential
$U_s^{\rm shp}$ and an explicit split between shaping input $u^{\rm shp}$,
port input $u^{\rm port}$, and damping injection.
(2) We give a practical SHAPE optimizer for clean-gradient, stochastic-gradient, and value-only oracle inputs, using the local port-Hamiltonian template and energy-balance diagnostics for noise, port work, and discretization defects. (3) We connect the method to finite-budget progress through supporting results on frozen-stage hypocoercive contraction, discrete contraction, hybrid memory-assisted improvement, and stochastic-oracle energy perturbations, and evaluate the resulting fixed-budget policy on synthetic, physics-based, and control-oriented nonconvex task families. SHAPE shows that finite-budget optimization benefits from treating stagnation as a learning-to-optimize event: stable descent remains useful, but the optimizer must decide when a basin has become uninformative.

\section{Preliminaries}
\label{sec:related}

\paragraph{Geometry of dynamics for optimization.}
Many optimizers choose a fixed-form update $\mathcal{G}$. For instance, momentum methods can be understood as discretizations of a second-order (underdamped) flow. In particular, they are conservative or dissipative Hamiltonian dynamics with constant mass $M\succ 0$ and linear damping. Given Hamiltonian
$H(q,p)=f(q)+\tfrac12 p^\top M^{-1}p$ and dissipation acting on momentum, these dynamics reduce to the classical
heavy-ball system
\begin{equation}
\dot q = M^{-1}p,
\qquad
\dot p = -\nabla f(q) - D\,M^{-1}p,
\label{eq:hb_ode_related}
\end{equation}
whose standard splittings yield Polyak momentum/momentum SGD \cite{polyak1964some} and related variants such as Nesterov acceleration
\cite{nesterov1983method}. Stochastic counterparts (underdamped Langevin\cite{welling2011bayesian}, SGHMC\cite{chen2014stochastic}, etc.) further add dissipation-consistent
noise to regulate exploration, but typically operate with a fixed geometry and without an explicit mechanism for bounded actuation under
finite oracle budgets.

\vspace{-0.5em}
\begin{wraptable}{r}{6.8cm}
\centering
\caption{Compact pH-lens summary of optimizer mechanisms. Symbols: \cmark\ = explicit/present; \xmark\ = absent; \pmark\ = implicit, optional, or embedding-dependent. Columns: Mom. = explicit phase/momentum state; Aux. = persistent auxiliary state such as moments, curvature estimates, or slow copies; Prec. = adaptive metric/preconditioner; $H_{\rm shp}$ = objective/energy shaping beyond the base objective $f(q)$; Port = explicit bounded control/interconnection port; Diff. = stochastic diffusion/noise; $+\nabla$ = extra gradient queries per update. $(0^\ast)$ denotes extra computation or auxiliary statistics but no additional gradient oracle call.}
\label{tab:summary:short}
\scriptsize
\setlength{\tabcolsep}{2.6pt}
\renewcommand{\arraystretch}{1.03}
\begin{tabular}{@{}lccccccc@{}}
\toprule
Method & Mom. & Aux. & Prec. & $H_{\rm shp}$ & Port & Diff. & $+\nabla$\\
\midrule
Heavy-ball     & \cmark & \xmark & \xmark & \xmark & \xmark & \xmark & $0$ \\
NAG            & \cmark & \xmark & \xmark & \xmark & \xmark & \xmark & $0$ \\
AdaGrad        & \xmark & \cmark & \cmark & \xmark & \xmark & \xmark & $0$ \\
RMSProp        & \xmark & \cmark & \cmark & \xmark & \xmark & \xmark & $0$ \\
Adam/AdamW     & \xmark & \cmark & \cmark & \xmark & \xmark & \xmark & $0$ \\
SGHMC\cite{chen2014stochastic}
               & \cmark & \xmark & \pmark & \xmark & \xmark & \cmark & $0$ \\
Lookahead\cite{zhang2019lookahead}
               & \xmark & \cmark & \xmark & \xmark & \pmark & \xmark & $0$ \\
SAM\cite{foret2021sharpnessaware}
               & \xmark & \xmark & \xmark & \cmark & \xmark & \xmark & $1$ \\
Shampoo\cite{gupta2018shampoo}
               & \xmark & \cmark & \cmark & \xmark & \xmark & \xmark & $0$ \\
SOAP\cite{vyas2024soap}
               & \xmark & \cmark & \cmark & \xmark & \xmark & \xmark & $0^\ast$ \\
Sophia\cite{liu2023sophia}
               & \xmark & \cmark & \cmark & \xmark & \xmark & \xmark & $0^\ast$ \\
Lion\cite{chen2024lion}
               & \pmark & \cmark & \pmark & \xmark & \xmark & \xmark & $0$ \\
Muon\cite{jordan2024muon}
               & \pmark & \cmark & \cmark & \xmark & \xmark & \xmark & $0^\ast$ \\
\textbf{SHAPE (ours)}
               & \cmark & \cmark & \cmark & \cmark & \cmark & \pmark & $0^\ast$ \\
\bottomrule
\end{tabular}
\end{wraptable}


\paragraph{Adaptive metrics and state augmentation.}
Many widely used optimizers can be interpreted as modifying \eqref{eq:hb_ode_related} by introducing a state- or time-dependent metric
(preconditioner) $M$ estimated from online gradient statistics, including AdaGrad \cite{duchi2011adaptive}, RMSProp \cite{tieleman2012lecture},
Adam/AdamW \cite{kingma2014adam,loshchilov2018decoupled}, and related variants. Other methods augment the state with additional copies or
filters: Lookahead \cite{zhang2019lookahead} introduces a slow parameter copy coupled to fast updates, while various look-ahead schemes
evaluate the force at a short-horizon prediction rather than the current iterate. These modifications can be viewed through a port-Hamiltonian
lens as changing the effective energy, the metric, or the measurement location.
\paragraph{Measurement-driven corrections and robustified objectives.}
Several recent approaches explicitly use gradients to reshape the effective descent direction or objective.
SAM \cite{foret2021sharpnessaware} replaces the empirical loss by a local robustification,
and Sophia \cite{liu2023sophia} incorporates curvature-weighted preconditioning with clipping to bound sharp-direction updates.
More recently, Lion-K \cite{chen2024lion} and Muon \cite{jordan2024muon} can be interpreted as port-Hamiltonian flows on augmented
state variables where the ``kinetic'' term encodes mirror geometry or implicit constraints. We summarize these families in
Table~\ref{tab:summary:short} and defer extended Hamiltonian system mappings to Appendix~\ref{app:disc_ph_optimizers}.
\paragraph{Hamiltonian dynamics and symplectic integration}
For a Hamiltonian system, its time integration scheme is relevant to the numerical stability of Hamiltonian dynamics \cite{leimkuhler2004simulating, girolami2011riemann, fu2025hamiltonian, van2006port, van2024port, CORDONI2022stochasticport, hong_sun_2023_symplecticSHS}.
The advantage of constructing Hamiltonian dynamics and applying symplectic integration is twofold. On the one hand, one can introduce a pseudo-Hamiltonian system for descent dynamics with an associated kinetic energy. On the other hand, one can use symplectic or conformal-symplectic integration schemes, depending on the differential form being discretized \cite{van2006port, hairer2006geometric}. We defer a more detailed discussion of Hamiltonian dynamics and time integration to Appendix \ref{sec:more:related:works}.
\paragraph{Learning to optimize (L2O) and learned update rules.}
Learning to optimize (L2O) learns an optimizer from a distribution of
optimizee problems, typically by parameterizing an iterative update rule and training it to reduce the downstream optimization loss
\cite{chen2022L2Osurvey,liu2023towards}.  The broader view of optimization problems as data points also appears in machine learning for combinatorial optimization, where learning is used to automate expensive branching, selection, or heuristic decisions over a task distribution~\cite{bengio2021machine}. \cite{metz2022practical} further show that learned optimizers must be evaluated not only by final loss but also by memory and compute overhead.  A related but distinct line is model-based deep learning and deep unfolding, which embeds an algorithmic solver into a trainable architecture \cite{shlezinger2022model}; for example, \cite{oshin2026deep} accelerate a feasible Quadratic Programming solver through deep unfolding.

\section{Method}
\label{sec:method}

SHAPE follows the same high-level goal of learning optimization behavior from task families, but differs in what is learned and how it is constrained.  Rather than regressing an unconstrained black-box update \((q_k,g(q_k))\mapsto q_{k+1}\), SHAPE learns a closed-loop phase-space optimizer whose fast dynamics retain port-Hamiltonian semantics, as shown in Figure~\ref{fig:teaser} and Algorithm~\ref{alg:streaming_shape_compact}.  Gradients enter as oracle measurements through ports, while the learned controller modifies the interconnection, damping, and shaping terms subject to skew-symmetry, dissipation, and soft passivity regularization. SHAPE is a two-timescale optimizer as a navigation process.  At test time it receives an oracle stream
\(
    \mathcal O(q)=\bigl(f(q),g(q)\bigr),
\)
where $g(q)=\nabla f(q)$ in the clean first-order setting and may be a
stochastic or estimated force under other oracle models.  The main text uses
$f(q)$ for the fixed task objective.  We denote $f_\tau$ for a distribution of task instances with flexible task parameter $\tau$.

\subsection{From local oracle data to a frozen stage}

At event stage $s$, memory $m_s$ summarizes regions visited by previous
stages.  The planner reads the current state, oracle values, and memory readout
through
\(
\label{eq:shape_stage_observation}
    o_s=
    \bigl(q_k,p_k,f(q_k),g(q_k),\mathrm{Read}(m_s,q_k)\bigr),
\)
and selects a stage action
\(
\label{eq:shape_stage_action}
    a_s=
    (\ell_s,\bar q_s,\alpha_s^J,\alpha_s^R,\kappa_s,H_s)
    =\pi_\phi(o_s).
\)
Here $\ell_s$ is a discrete mode, $\bar q_s$ is a local anchor,
$\alpha_s^J$ and $\alpha_s^R$ modulate skew and dissipative channels,
$\kappa_s$ is the anchor strength, and $H_s$ is the stage horizon.  These quantities are held fixed during the short local traversal so that the fast motion has a frozen port-Hamiltonian interpretation. The anchor $\bar q_s$ is not assumed to be a known goal.  At test time it is the output of the slow planner $\pi_\phi$ from the current local oracle and memory context.  In the implementation used for the experiments, $\pi_\phi$ is trained with a probe-based teacher: a small trust-region candidate set around
$q_k$ is generated from the descent direction, momentum direction, memory-novel
directions, and random orthogonal directions; candidates are scored by predicted
one-step improvement, novelty, and excursion risk; the best candidate becomes
the supervised anchor label.  Appendix~\ref{subsec:dp_training} gives the exact
candidate and scoring rule. The shaped potential used by the stage is
\begin{equation}
\label{eq:shape_shaped_potential}
    U^{\rm shp}_{s,k}(q)
    =
    U_\eta(q;m_s)
    +\frac{\kappa_s}{2}\|q-\bar q_s\|^2
    +V_{{\rm bar},s}(q;m_s,\ell_s).
\end{equation}
The first term is the memory-induced reshaping potential, the second is the local anchoring potential, and the third is an optional barrier/exclusion potential activated by the stage mode.  With this convention, all reshaping terms are contained in $U^{\rm shp}_{s,k}$ and the task objective remains $f(q)$. The fast controller evaluates
\begin{equation}
\label{eq:shape_fast_controller}
    (M_k,\Omega_k,D_k,u_k^{\rm shp},K_k^d)
    =
    \pi_\psi\bigl(q_k,p_k,g(q_k),\mathrm{Read}(m_s,q_k),a_s\bigr).
\end{equation}
Here $M_k\succ0$ defines the local kinetic metric, $\Omega_k^\top=-\Omega_k$ is lossless,
$D_k\succeq0$ is dissipative, $u_k^{\rm shp}$ is a bounded active shaping input, and
$K_k^d\succeq0$ injects additional passive damping through the port output.  The local
Hamiltonian is
\begin{equation}
\label{eq:shape_local_hamiltonian}
    H_{s,k}(q,p)
    =
    f(q)+U^{\rm shp}_{s,k}(q)+\frac12p^\top M_k^{-1}p .
\end{equation}
\subsection{Port-Hamiltonian(pH) stage update}
The effort variables associated with \eqref{eq:shape_local_hamiltonian} are
\begin{equation}
\label{eq:shape_efforts}
    e_k^q
    =
    g(q_k)+\nabla_qU^{\rm shp}_{s,k}(q_k),
    \qquad
    e_k^p=M_k^{-1}p_k.
\end{equation}
A port output is defined by
\begin{equation}
\label{eq:shape_port_output}
    y_k=G^\top e_k^p,
    \qquad
    u_k^{\rm port}=u_k^{\rm shp}-K_k^dy_k.
\end{equation}
Thus $u^{\rm port}$ is the actual input applied to the plant, while
$u^{\rm shp}$ is the learnable energy-injection component.  The term
$-K^dy$ is passive damping injection because $y^\top K^dy\ge0$. With the controller substituted through the feedback-reduced port law, the
local stage evolves as a closed-loop pH-like system
\begin{equation}
\label{eq:shape_ph_vector_field}
    \begin{bmatrix}
    \dot q\\
    \dot p
    \end{bmatrix}
    =
    \underbrace{
    \begin{bmatrix}
    0&I\\
    -I&\alpha_s^J\Omega_k-\alpha_s^RD_k
    \end{bmatrix}}_{\mathcal A_{s,k}}
    \nabla H_{s,k}(q,p)
    +
    \begin{bmatrix}
    0\\
    G u_k^{\rm port}
    \end{bmatrix}.
\end{equation}
This equation identifies the energy channels: the skew part transports energy, the dissipative part removes energy, and only the bounded shaping component can inject energy through the port.  The implementation uses the semi-implicit first-order step
\begin{align}
    p_{k+1}
    &=
    p_k
    -h e_k^q
    +h\alpha_s^J\Omega_k e_k^p
    -h\alpha_s^R D_k e_k^p
    +hGu_k^{\rm port},
    \label{eq:shape_p_update}
    \\
    q_{k+1}
    &=
    q_k+hM_k^{-1}p_{k+1}.
    \label{eq:shape_q_update}
\end{align}
Appendix~\ref{subsec:dp_algorithm_sketch} discusses details regarding this discretization. Therefore the continuous model is not used to claim that the code solves the ODE exactly; it is used to
organize the learned update into interpretable pH energy channels. Along a frozen stage with fixed $H_{s,k}$ and port output $y_k$, the formal energy balance is
\begin{equation}
\label{eq:idapbc_energy_balance}
    \dot H_{s,k}
    = 
    - (e^p)^\top \alpha_s^RD_k e^p
    - y^\top K_k^d y
    + y^\top u_k^{\rm shp},
\end{equation}
because the skew contribution satisfies $(e^p)^\top\Omega_ke^p=0$.  This is the effective interconnection inspired structure throughout the paper: $U^{\rm shp}$ changes the energy landscape, $\Omega$ changes lossless transport, $D$ and $K^d$
provide dissipation, and $u^{\rm shp}$ is the only deliberate active injection. 

\begin{algorithm}[t]
\caption{SHAPE optimizer (compact)}
\label{alg:streaming_shape_compact}
\begin{algorithmic}[1]
\State \textbf{Input:} oracle $\mathcal O(q)=(f(q),g(q))$, initial state $(q_0,p_0)$, memory $m_0$, step size $h$
\State $k\gets0$, $\widehat q\gets q_0$
\For{event stages $s=0,\ldots,S-1$}
    \State Query $(f_k,g_k)=\mathcal O(q_k)$ and read $r_s=\mathrm{Read}(m_s,q_k)$
    \State Plan $a_s=(\ell_s,\bar q_s,\alpha_s^J,\alpha_s^R,\kappa_s,H_s)=\pi_\phi(q_k,p_k,f_k,g_k,r_s)$
    \For{local steps $j=0,\ldots,H_s-1$}
        \State Read $r_k=\mathrm{Read}(m_s,q_k)$
        \State Compute $(M_k,\Omega_k,D_k,u_k^{\rm shp},K_k^d)=\pi_\psi(q_k,p_k,g_k,r_k,a_s)$
        \State Form $U^{\rm shp}_{s,k}$ by \eqref{eq:shape_shaped_potential}, efforts by \eqref{eq:shape_efforts}, and $u_k^{\rm port}$ by \eqref{eq:shape_port_output}
        \State Update $(q_{k+1},p_{k+1})$ by \eqref{eq:shape_p_update}--\eqref{eq:shape_q_update}
        \State Query $(f_{k+1},g_{k+1})=\mathcal O(q_{k+1})$ and update $\widehat q\gets\arg\min\{f(\widehat q),f(q_{k+1})\}$
        \State $k\gets k+1$
        \If{event trigger detects convergence, stall, or budget exhaustion}
            \State \textbf{break}
        \EndIf
    \EndFor
    \State Update memory $m_{s+1}=\mathcal U(m_s,\{q_i,p_i,f(q_i),g(q_i)\}_{i\in\mathcal I_s})$
\EndFor
\State \textbf{return} best-so-far iterate $\widehat q$
\end{algorithmic}
\end{algorithm}

\paragraph{Oracle-independent force models}
The same pH template is used for clean gradients, stochastic gradients, and value-only access by replacing $\nabla f(q_{s,n})$ with an estimator $\widetilde g_{s,n}$.  The local force convention is
\begin{equation}
\label{eq:oracle_generic_force}
    \widetilde g_{s,n}\approx\nabla f(q_{s,n}),
    \qquad
    e_{s,n}^q=\widetilde g_{s,n}+\nabla U^{\rm shp}_{s,n}(q_{s,n}).
\end{equation}
For deterministic first-order access,
\begin{equation}
\label{eq:oracle_clean_gradient}
    \widetilde g_{s,n}=\nabla f(q_{s,n}).
\end{equation}
For stochastic or mini-batched gradients,
\begin{equation}
\label{eq:oracle_stochastic_gradient}
    \widetilde g_{s,n}=\nabla f(q_{s,n})+\epsilon_{s,n},
    \qquad
    \mathbb E[\epsilon_{s,n}\mid\mathcal F_{s,n}]=0,
    \qquad
    \mathbb E[\epsilon_{s,n}\epsilon_{s,n}^{\top}\mid\mathcal F_{s,n}]=\Sigma_{s,n}.
\end{equation}
For value-only access, we use a directional finite-difference estimator
\begin{equation}
\label{eq:oracle_zo_gradient}
    \widetilde g_{s,n}
    =
    \frac1{K_s}
    \sum_{i=1}^{K_s}
    \frac{f(q_{s,n}+\varepsilon_su_i)-f(q_{s,n}-\varepsilon_su_i)}{2\varepsilon_s}
    u_i,
    \qquad
    \|u_i\|=1.
\end{equation}
The deterministic energy increment follows \eqref{eq:idapbc_energy_balance}
up to discretization error.  Under stochastic gradients, the expected increment contains the covariance defect. For zeroth-order estimation, the same display holds with additional bias and
variance defects from the finite-difference estimator; the detailed statement is deferred to Appendix~\ref{app:dp_stochastic_extension}.

\subsection{Event-level policy and training}
\label{subsec:event_policy_training}

The oracle model changes the local force estimator but not the outer loop.  At
event time $s$, the planner forms a history-dependent context
\(
    \omega_s=\Xi\bigl(q_s^0,p_s^0,m_s^0,\mathcal S_{0:s-1}\bigr),
\)
and outputs
\begin{equation}
\label{eq:outer_policy_action}
    a_s=\pi_\phi^{\mathrm G}(\omega_s)
    =
    \bigl(\bar q_s,\mu_s,\bar\alpha_s^J,\bar\alpha_s^R,
    \bar\kappa_s,b_s\bigr).
\end{equation}
Here $\mu_s\in\{\mathrm{settle},\mathrm{refine},\mathrm{escape}\}$ is the stage
mode and $b_s$ is the stage oracle budget.  Given $a_s$, the local solver produces discrete update as realization of diffeomorphism from local Hamiltonian:
\begin{equation}
\label{eq:outer_stage_rollout}
    x_{s,n+1}
    =
    \Phi^{h_s}_{\psi}
    \bigl(x_{s,n};a_s,m_s^0,\widetilde g_{s,n}\bigr),
    \qquad n=0,\ldots,N_s-1.
\end{equation}
The accepted point is updated by the best-so-far rule
\(
    \widehat q_{s+1}
    \in
    \arg\min\Bigl\{
        f(\widehat q_s),
        \min_{0\le n\le N_s}f(q_{s,n})
    \Bigr\}.
\)
Finally, the event map updates memory and initializes the next stage:
\(
    m_{s+1}^0=\mathcal U_\eta^{\rm mem}(m_s^0,\mathcal S_s),
    ~
    q_{s+1}^0=q_s^+,
    ~
    p_{s+1}^0=p_s^+.
\)
Thus the plant state is continuous across stages, while memory, mode, anchor,
and structural gains are refreshed only on the slow event clock.

\paragraph{Training loss.}
Our training of SHAPE optimizer is stagewise.  The local controller is first initialized by a local pretraining loss \(\mathcal L_{\rm loc}\).  The subsequent joint
planner--controller rollout phase uses the code-faithful objective
{\small
\begin{equation}
\label{eq:main_training_objective}
\begin{aligned}
\mathcal L_{\rm roll}
=
\underbrace{\lambda_{\rm term}\mathcal L_{\rm term}
+\lambda_{\rm best}\mathcal L_{\rm best}
+\lambda_{\rm prog}\mathcal L_{\rm prog}}_{\text{L2O type of loss}} +\underbrace{\mathcal L_{\mathrm{plan}}^{\mathrm{sup}}}_{\text{planner loss}}
+\underbrace{\lambda_{\rm ctrl}\mathcal L_{\rm ctrl}
+\lambda_{JR}\mathcal L_{JR}
+\lambda_{\rm port}\mathcal L_{\rm port}}_{\text{structure preserving loss}}.
\end{aligned}
\end{equation}
}
Here \(\mathcal L_{\rm term}\) penalizes the normalized terminal distance,
\(\mathcal L_{\rm best}\) penalizes the best-seen normalized distance along the
rollout, and \(\mathcal L_{\rm prog}\) encourages progress toward the current
stage waypoint \(\bar q_s\).  The two planner-supervision terms respectively
train the planner mode and the proposed waypoint.  The remaining terms penalize
control magnitude, large learned structure/damping/port parameters, and
positive port-power violations. Detailed
form of losses are stated in Appendix~\ref{subsec:dp_training} and the detailed schedule, weights, and training epochs are reported in Appendix~\ref{app:exp:shape_training_reproducibility}.

\subsection{Theoretical analysis}
\label{subsec:main_theory_summary}

We summarize the main proposition here and defer its proof in Appendix \ref{app:dp_finite_budget_extension}.

\begin{proposition}[Budgeted progress of the event-triggered outer loop]
\label{prop:outer_budgeted_progress}
Let
\(
    F_s:=f(\widehat q_s)
\)
denote the accepted best-so-far value after stage $s$.  Suppose that,
conditional on the history $\mathcal F_s$, the planner--memory update proposes
a stage action $a_s$ whose induced frozen stage has a reachable lower basin
with probability at least $\vartheta_s$.  More precisely, on this proposal event,
the next frozen-stage equilibrium $q_{s+1}^{\star}$ satisfies
\(
    f(q_{s+1}^{\star})
    \le
    F_s-\delta-\Delta_s,
\)
where $\delta>0$ is the desired improvement margin and $\Delta_s\ge0$
dominates the stage residual, shaping-distortion, and numerical-defect terms
appearing in Appendix~\ref{app:shape_theory}.  Suppose further that, under the
chosen oracle model and allocated budget $b_s$, the local IDA-PBC rollout
reaches this basin with probability at least $r_s$.  Then
\begin{equation}
\label{eq:outer_one_stage_success}
    \mathbb P(F_{s+1}\le F_s-\delta\mid\mathcal F_s)
    \ge
    \vartheta_s r_s .
\end{equation}
Consequently, after $S$ stages,
\begin{equation}
\label{eq:outer_no_improvement_probability}
    \mathbb P(F_S>F_0-\delta)
    \le
    \prod_{s=0}^{S-1}(1-\vartheta_s r_s).
\end{equation}
In particular, if $\vartheta_s r_s\ge\rho>0$ for all $s<S$, then
\begin{equation}
\label{eq:outer_exponential_budget_progress}
    \mathbb P(F_S>F_0-\delta)
    \le
    \exp(-\rho S).
\end{equation}
\end{proposition}

The factor \(\vartheta_s\) describes the slow planner--memory mechanism:
conditioned on the past, it is the probability that the next stage action
exposes a basin whose local equilibrium has sufficiently lower value for the
original objective.  The factor \(r_s\) describes the local IDA-PBC rollout: it is the probability that the frozen-stage dynamics reaches the corresponding basin within the allocated stage budget. In particular, the local success factor \(r_s\) is supported by the
continuous-time frozen-stage contraction
(Theorem~\ref{lem:dp_canonical_frozen_stage_rate}), its implemented
semi-implicit counterpart with truncation, port-work, and projection defects
(Theorem~\ref{lem:dp_discrete_frozen_stage_contraction}), and the
approximate-equilibration estimate
(Proposition~\ref{prop:dp_frozen_stage_conv}).  The stochastic-oracle case
modifies the same defect floor through the covariance-weighted term in
\eqref{eq:dp_stochastic_discrete_ineq}.  Finally, the hybrid
memory-assisted improvement result
(Proposition~\ref{prop:dp_hybrid_memory_improvement}) shows how local
equilibration in the shaped potential transfers back to best-so-far
improvement in \(f\), up to the explicit residual, shaping-distortion, and
numerical-error terms.

\section{Numerical Experiments}
\label{sec:experiments}
We evaluate the proposed dual-policy optimizer on synthetic and differentiable scientific/engineering objectives under matched oracle-query budgets. The experiments test three questions: (i) whether the fast port-Hamiltonian (pH) / IDA-PBC controller improves local stabilization, (ii) whether the slow planner and memory state help escape poor basins, and (iii) whether the learned policy transfers zero-shot to held-out task instances. The main text reports the compact protocol and metric tables; full task definitions, oracle variants, hyperparameters, full numerical tables, and ablation details are deferred to \cref{app:exp:details}.
\paragraph{Protocol.}
Each task instance is written as a differentiable finite-budget optimization problem
\(
  \min_{q\in\mathcal X_\tau} f_\tau(q),
  \) with task specific parameter $\tau$. All learned components are trained with the same curriculum stated in Appendix~\ref{subsec:dp_training}: Phase~I emphasizes short-horizon local reachability, Phase~II trains coarse exploration and memory updates, and Phase~III performs joint end-to-end rollout training under the final task loss. At test time, all learned parameters are frozen and evaluated zero-shot on held-out tasks. Classical baselines are run under matched oracle-query budgets.

\paragraph{Metrics.}
Whenever the task admits a reference optimum or terminal proxy, we report the final gap, best gap and auc gap computed denoted as
{\small
\begin{align}
  \mathrm{FinalGap} = f_\tau(q_N)\!-\!f_\tau^\star, ~ 
  \mathrm{BestGap}  = \min_{0\le k\le N} f_\tau(q_k)\!-\!f_\tau^\star, ~
  \mathrm{AUCGap}  = \frac{1}{N}\!\sum_{k=1}^{N} \bigl(f_\tau(q_k)-f_\tau^\star\bigr).
  \label{eq:exp:metrics_main}
\end{align}
}
We also report hit rate under family-specific tolerances and, when meaningful, final distance to the known minimizer or to an equivalence class of minimizers. For tasks with symmetries, such as sign ambiguity in phase retrieval or permutation/rotation invariance in cluster energies, distance metrics are computed after quotient alignment.

\begin{figure}[t]
    \centering
    \includegraphics[width=0.48\linewidth]{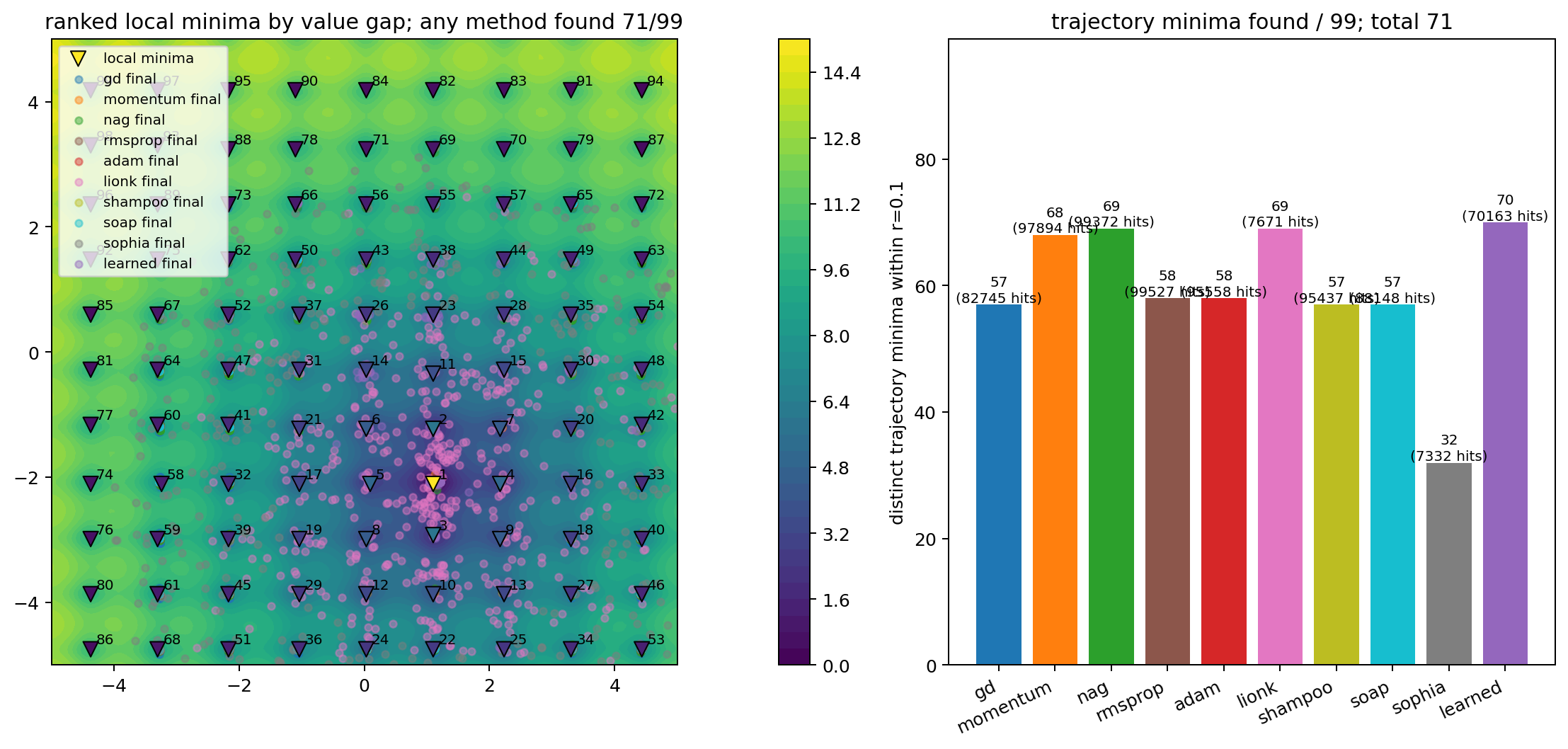}
    \includegraphics[width=0.48\linewidth]{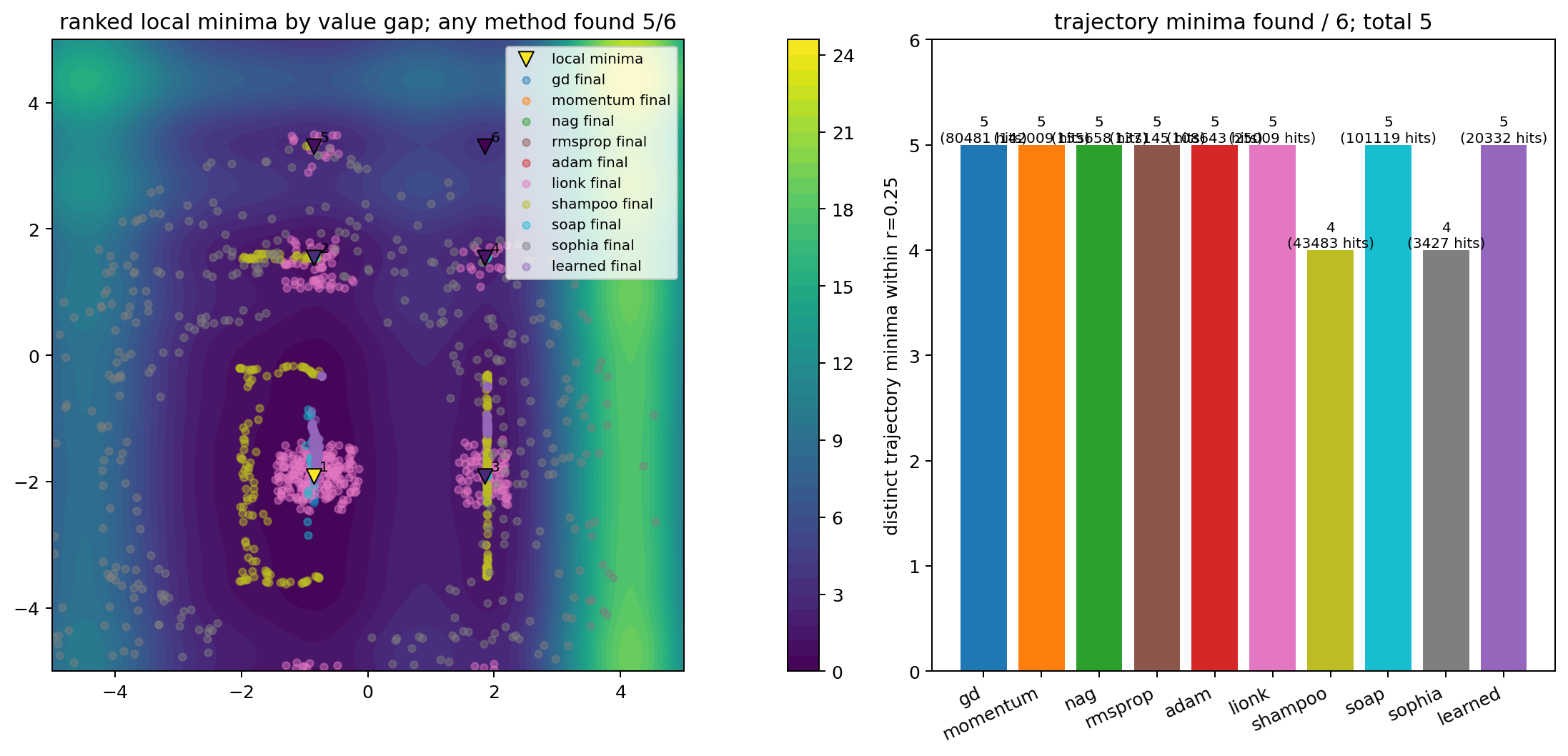}
    \caption{The minima finding performance over $512$ random initialization starting locations. These figures display one sampled task instance of Ackley(Left) and Levy(Right) functional. The mean performance and some sampled trajectories are displayed in Figure~\ref{fig:2d}. Together these figures show that SHAPE can traverse the landscape but each search trajectory is more informative compared with fixed gradient descent methods.}
    \label{fig:minima:counting}
\end{figure}
\vspace{-0.5em}
\begin{table}[t]
\centering
\caption{Ablation over task count, rollout particles, and per-particle descent budget for the Ackley first-order benchmark.  Each column is one evaluation setting \((N_{\rm task},N_{\rm part},B)\); all settings have the same total per-task rollout budget \(N_{\rm part}B=32768\).  Lower gaps are better; higher hit rates and minima counts are better.}
\label{tab:ablation_ackley_compact_transposed}
\scriptsize
\setlength{\tabcolsep}{3.0pt}
\begin{tabular}{lcccccc}
\toprule
Setting & S1 & S2 & S3 & S4 & S5 & S6 \\
\midrule
\((N_{\rm task},N_{\rm part},B)\) &
\((16,128,256)\) &
\((32,128,256)\) &
\((64,128,256)\) &
\((64,64,512)\) &
\((64,32,1024)\) &
\((64,16,2048)\) \\
\midrule
SHAPE final gap \(\downarrow\) & 2.997 & 2.321 & 1.645 & 2.105 & 1.861 & 1.651 \\
SHAPE best gap \(\downarrow\) & \(2.2{\times}10^{-4}\) & \(1.4{\times}10^{-4}\) & \(1.0{\times}10^{-3}\) & \(3.6{\times}10^{-5}\) & \(9.5{\times}10^{-6}\) & \(<10^{-10}\) \\
SHAPE hit-traj. \(\uparrow\) & 40.0\% & 45.3\% & 63.7\% & 52.1\% & 62.5\% & 72.7\% \\
SHAPE minima \(\uparrow\) & 57 & 70 & 105 & 71 & 78 & 70 \\
Best terminal baseline & LionK & LionK & NAG & LionK & NAG & NAG \\
Baseline final gap \(\downarrow\) & 5.910 & 5.901 & 5.808 & 5.704 & 4.927 & 4.759 \\
\bottomrule
\end{tabular}
\vspace{-0.4em}
\end{table}

\subsection{Results for different optimization problems}
\label{subsec:main_result_table}

\paragraph{Summary of results.}
\Cref{tab:main:result_placeholder} summarizes the appendix tables using a space-efficient aggregation. For a dimension sweep, each entry is the uniform average of the per-dimension means reported in Table~\ref{tab:summary:synthetic:benchmarks} and Table ~\ref{tab:real:world:example:summary}; the full mean$\pm$standard-deviation values for every dimension remain in the appendix. For each family, the baseline row is the strongest classical baseline selected by the average \emph{BestGap}; this avoids choosing a different baseline for each metric. 1D  multi-well problems are replicated directly from Table~\ref{tab:multiwell_fair_budget}, where only gap, success, and gradient-call metrics are reported.

\paragraph{Main observations.}
The dimension-averaged table supports the intended empirical claim on basin-escape and nonconvex navigation tasks: SHAPE improves BestGap and hit rate on the multi-well study, Ackley, low-to-moderate L\'evy summaries, Lennard--Jones, phase retrieval, and the control best-so-far metric. The same table also exposes two important limitations. First, coordinate-adaptive baselines remain very strong on high-dimensional separable analytic functions, most visibly Rastrigin, where RMSProp dominates the dimension-averaged gap metrics. Second, in control trajectory optimization, SHAPE attains a better best-so-far objective but not the best terminal objective, indicating that the current controller discovers good regions before fully stabilizing to them within the fixed budget. Figure~\ref{fig:minima:counting} displays how many local minima different optimizers visit.


\subsection{Ablation Summary}
\label{subsec:main_ablation_summary}

We justify the importance of the additional structure we introduce in our SHAPE optimizer. We use the Ackley first-order benchmark to isolate whether SHAPE's gains come from its learned closed-loop structure rather than from a larger evaluation budget.  Here \(N_{\rm task}\) denotes the number of distinct Ackley task instances, \(N_{\rm part}\) denotes the number of rollout particles or random initial starts evaluated per task, and \(B\) denotes the per-particle descent budget, i.e., the number of optimizer steps allocated to each rollout. Table~\ref{tab:ablation_ackley_compact_transposed} fix total budget at
\(N_{\rm part}B=32768\).  SHAPE remains competitive across all settings and improves as the evaluation covers more tasks or allocates longer per-particle rollouts: the best-seen gap decreases from \(2.2\times 10^{-4}\) in S1 to below \(10^{-10}\) in S6, while the trajectory hit rate increases from \(40.0\%\) to \(72.7\%\).  Table~\ref{tab:ackley_memory_controller_ablation_main}
then isolates the role of the local interconnected controller.  Removing the controller substantially worsens the best-seen gap in all reported dimensions, especially in \(d=2\) and \(d=100\), indicating that the learned controller is not merely adding parameters but contributes to the navigation policy. We also provide comparison with some searching based algorithm including basin-hopping\cite{wales1997global} and Cross-Entropy Methods\cite{de2005tutorial} and display the results in Figure~\ref{fig:ackley:with:search:algo}.

\paragraph{Relevance of the interconnected controller.}
The ablation supports the intended interpretation of SHAPE as a closed-loop optimizer for fixed-budget nonconvex navigation.  The controller-enabled model is better at finding low-value points along the rollout, which is the relevant criterion when the optimizer must decide whether to exploit a basin, escape a poor local region, or continue exploration from memory-informed states.  The comparison against the strongest classical baseline in Table~\ref{tab:ablation_ackley_compact_transposed} further shows that SHAPE's advantage is most visible in best-seen performance and hit rate, rather than only terminal convergence.  We therefore report terminal gaps and best-seen gaps separately: the former measures final-state stability, while the latter measures fixed-budget minima hunting, which is the main target of the proposed stagewise SHAPE design. In addition, Table~\ref{tab:ackley_oracle_ablation_compact_transposed} reports the empirical oracle-input ablation using the same checkpoint trained on first-order gradients to justify our method is robust against different test-time observed gradients.
\begin{table}[t]
\centering
\small
\caption{Main fixed-budget performance summary average over $64$ tasks. For performance across different dimensions, entries in this table are uniform averages of the per-dimension means whose per-dimension value are given in Appendix \ref{app:exp:task_full_details}. ``matched'' indicates the same oracle-query budget as the corresponding appendix experiment.}
\label{tab:main:result_placeholder}
\setlength{\tabcolsep}{3.0pt}
\resizebox{\linewidth}{!}{%
\begin{tabular}{lllccccccc}
\toprule
Family & Oracle & Dim. & Method & Final dist. $\downarrow$ & Final gap $\downarrow$ & Best gap $\downarrow$ & Hit rate $\uparrow$ & AUC gap $\downarrow$ & Calls $\downarrow$ \\
\midrule
\multicolumn{10}{l}{\textit{Illustrative basin-escape task}} \\
Multi-well & noisy 1st & $1$ & SHAPE / full & -- & \textbf{0.987} & \textbf{0.477} & \textbf{0.602} & -- & 1251.2 \\
Multi-well & noisy 1st & $1$ & SHAPE w/o memory & -- & 1.249 & 1.066 & 0.300 & -- & 707.3 \\
Multi-well & noisy 1st & $1$ & Momentum & -- & 1.290 & 1.115 & 0.300 & -- & 1251.2 \\
\addlinespace[2pt]
\midrule
\multicolumn{10}{l}{\textit{Analytic synthetic benchmarks}} \\

Ackley & exact 1st & $2,20,100,500$ & SHAPE / full & \textbf{0.226} & \textbf{0.679} & \textbf{0.323} & \textbf{0.486} & \textbf{2.27} & matched \\
Ackley & exact 1st & $2,20,100,500$ & NAG & 1.15 & 1.71 & 1.31 & 0.389 & 2.34 & matched \\
L\'evy & exact 1st & $2,20,100,500$ & SHAPE / full & \textbf{0.303} & \textbf{0.074} & \textbf{0.0427} & 0.227 & 1.10 & matched \\
L\'evy & exact 1st & $2,20,100,500$ & RMSProp & 0.728 & 0.202 & 0.202 & \textbf{0.533} & \textbf{0.420} & matched \\
Rastrigin & exact 1st & $2,20,100,500$ & SHAPE / full & 3.01 & $1.34\mathrm{e}{+3}$ & 613 & 0.00675 & $1.19\mathrm{e}{+3}$ & matched \\
Rastrigin & exact 1st & $2,20,100,500$ & RMSProp & \textbf{2.20} & \textbf{6.81} & \textbf{6.81} & \textbf{0.171} & \textbf{33.7} & matched \\
\addlinespace[2pt]
\midrule
\multicolumn{10}{l}{\textit{Differentiable scientific and engineering tasks}} \\
Lennard--Jones & autodiff 1st & $6,18$ & SHAPE / full & \textbf{0.416} & \textbf{0.254} & \textbf{0.113} & \textbf{0.190} & 0.837 & matched \\
Lennard--Jones & autodiff 1st & $6,18$ & RMSProp & 1.73 & 0.519 & 0.519 & 0.000 & \textbf{0.717} & matched \\
Phase retrieval & full/mini-batch 1st & $8,32$ & SHAPE / full & \textbf{0.168} & \textbf{0.00601} & \textbf{0.00351} & \textbf{0.631} & 0.420 & matched \\
Phase retrieval & full/mini-batch 1st & $8,32$ & NAG & 2.09 & 0.040 & 0.040 & 0.000 & \textbf{0.169} & matched \\
Control trajopt & adjoint/autodiff 1st & $8,32$ & SHAPE / full & \textbf{0.363} & 2.49 & \textbf{0.106} & \textbf{0.316} & 9.93 & matched \\
Control trajopt & adjoint/autodiff 1st & $8,32$ & RMSProp & 1.81 & \textbf{1.33} & 1.33 & 0.002 & \textbf{3.51} & matched \\
\bottomrule
\end{tabular}%
}
\end{table}
\begin{table}[t]
\centering
\caption{Ackley ablation by local-controller activation. Entries report best-seen gap with terminal gap in parentheses, averaged over the evaluation trials. Lower is better; the best best-seen gap in each dimension is bold.}
\label{tab:ackley_memory_controller_ablation_main}
\small
\begin{tabular}{lccc}
\toprule
Controller & $d=2$ & $d=20$ & $d=100$ \\
\midrule
\cmark & \textbf{0.794 (1.638)} & \textbf{2.406 (2.406)} & \textbf{0.001 (0.008)} \\
 \xmark & 2.865 (2.922) & 2.680 (2.681) & 0.685 (0.687) \\
\bottomrule
\end{tabular}
\vspace{-0.5em}
\end{table}
\begin{table}[t]
\centering
\caption{Gradient-oracle ablation on the Ackley benchmark using the same SHAPE checkpoint trained with first-order oracle inputs. Test-time oracle input is varied across first-order, stochastic, and zeroth-order gradient estimates. Lower gaps are better; higher trajectory hit rate and minima count are better. Parentheses identify the strongest non-SHAPE baseline for the corresponding metric.}
\label{tab:ackley_oracle_ablation_compact_transposed}
\scriptsize
\setlength{\tabcolsep}{4.0pt}
\begin{tabular}{lccc}
\toprule
Test-time oracle & First-order & Stochastic & Zeroth-order \\
\midrule
SHAPE final gap $\downarrow$ & 2.105 & 1.709 & 1.574 \\
SHAPE best gap $\downarrow$ & $3.6{\times}10^{-5}$ & $5.7{\times}10^{-6}$ & $2.5{\times}10^{-4}$ \\
SHAPE hit-traj. $\uparrow$ & 52.1\% & 66.6\% & 72.7\% \\
\midrule
Best baseline final $\downarrow$ & 5.704 (LionK) & 6.030 (LionK) & 3.800 (NAG) \\
Best baseline best $\downarrow$ & $4.2{\times}10^{-5}$ (NAG) & $3.2{\times}10^{-4}$ (NAG) & $1.1{\times}10^{-5}$ (NAG) \\
Best baseline hit-traj. $\uparrow$ & 3.1\% (Momentum) & 2.7\% (NAG) & 90.4\% (NAG) \\
\bottomrule
\end{tabular}
\end{table}

\section{Conclusion}
\label{sec:conclusion}
We propose SHAPE, an adaptive port Hamiltonian template that aims at learning to find minima of a task potential under finite budgets. SHAPE shows that fixed-budget optimization benefits from treating stagnation as a control event. Stable local descent is useful, but under a finite budget the optimizer must also decide when a basin has become uninformative. Port-Hamiltonian staging gives a structured way to separate local stabilization, active energy shaping, and memory updates. Across the tested task families, this separation improves best-so-far minima hunting under matched budgets, while high-dimensional separable landscapes remain challenging for the current implementation.
\paragraph{Limitation of our work}
In high-dimensional spaces, online gradient-descent variants are often preferred and have proved practically effective for training overparameterized or large-scale deep neural-network models \cite{li2018visualizing}. Although our method supports compressed memory independently of the ambient parameter dimension, information loss---especially the validation of vector-field topology---is intractable in the current implementation. How to decompose high-dimensional state spaces and perform efficient updates remains an open research question. Another relevant extension is to incorporate uncertainty quantification, or the variance of stochastic-gradient estimates, into the controller. Non-asymptotic analyses of SGHMC-like update schemes provide one possible route \cite{liang2024non, chau2022stochastic}. Nevertheless, worst-case dimensional dependence remains a limitation because some bounds contain terms that grow rapidly with dimension.
\paragraph{LLM usage disclaimer}
Large language models were used to help polish the manuscript and correct grammar. Some code, visualizations, and table formatting were prepared with AI assistance. The core technical ideas, theoretical claims, and experimental design are not claimed by LLM agents.
\paragraph{Social Impact Statement}
 Our work does not rely on private assets. Although there exists risk of the abuse of usage of our learning algorithm, but none of which is significant to our best knowledge.

\paragraph{Acknoledgement}
    This research was supported in part by a grant from the Peter O’Donnell Foundation, the Michael J Fox Foundation, Jim Holland- Backcountry Foundation and in part from a grant from the Army Research Office accomplished under Cooperative Agreement Number W911NF-19-2-0333

\newpage
\bibliographystyle{splncs04}
\bibliography{refs}

\newpage
\appendix
\section{Notation and nomenclature}
\label{app:notation}

We provide a notation list that list out used symbols throughout the paper and the following appendices.

\begin{table}[h]
\caption{Notation table.}
\footnotesize
\begin{tabular}{p{0.22\textwidth}p{0.70\textwidth}}
\toprule
\textbf{Symbol} & \textbf{Meaning and convention} \\
\toprule
$\mathcal Q\subset\mathbb R^d$ & Search domain or feasible set for the optimization variable. \\
$q\in\mathcal Q$ & Decision variable / primal optimizer state.  The paper uses $q$ for the optimizee variable. \\
$q^\star$ & Global or reference minimizer when it is used for analysis or supervised training diagnostics.  It is not assumed available at test time unless explicitly stated. \\
$\widehat q_s$ & Accepted best-so-far iterate after event stage $s$. \\
$f(q)$ & Base task objective, target potential, or loss. \\
$f_\tau(q)$ & Objective for a sampled task instance $\tau$.  In main-text formulas with a fixed task, the subscript is suppressed and $f_\tau$ is written as $f$. \\
$\varphi_\tau$ & Forward dynamics map for a simulator/control task, used only when the objective is induced by a rollout. \\
$g(q)$, $\nabla f(q)$ & First-order oracle gradient.  In stochastic or zeroth-order settings, the force estimator is denoted by $\widetilde g_{s,n}$. \\
$\mathcal O(q)$ & Oracle output, typically $\mathcal O(q)=(f(q),g(q))$ for first-order access. \\
$x=(q,p)$ & Phase-space optimizer state. \\
$p\in T_q^\ast\mathcal Q$ & Momentum / cotangent transport coordinate conjugate to $q$. \\
$H_{s,k}$, $H_{d,s}$ & Stagewise shaped Hamiltonian.  It contains the base objective $f$, kinetic energy, and stage-dependent shaping terms. \\
$M_s,M_k$ & Mass, metric, or preconditioning matrix; assumed positive definite when inverted. \\
$J$, $J_s$, $\Omega_s$ & Skew-symmetric interconnection or transport operators.  $\Omega_s^\top=-\Omega_s$. \\
$R_s,D_s$ & Dissipation or damping operators; positive semidefinite when used in passivity estimates. \\
$G,B_s$ & Port/input matrices that map controller input into plant dynamics. \\
$u_s$, $u_s^{\rm port}$, $u_s^{\rm shp}$ & Controller input, port input, and shaping component of the port input. \\
$y_s$ & Collocated port output, power-conjugate to the corresponding port input. \\
$m_s$ & Event-stage memory state summarizing previously visited regions and observed landscape statistics. \\
$U_\eta(q;m_s)$ & Memory-induced reshaping potential. \\
$U_{\mathrm{goal},s}$, $U_{\mathrm{mem},s}$, $V_{\mathrm{bar},s}$ & Goal, memory, and barrier/exclusion shaping terms in the shaped Hamiltonian. \\
$b_\mu(q)$ & Feasibility or barrier penalty used only when constraints are modeled; $b_\mu\equiv0$ in unconstrained experiments. \\
$s,k,n$ & Event-stage index, global/local discrete step index, and within-stage integration index, respectively. \\
$h_s$, $H_s$, $b_s$ & Stage step size, local rollout horizon, and allocated oracle budget. \\
$\pi_\phi^{\mathrm G}$, $\pi_\psi^{\mathrm L}$ & Slow global planner and fast local controller. \\
$\Pi_\eta$, $\mathcal U_\eta^{\rm mem}$ & Event update / memory-write map. \\
$\mu_s$ & Discrete stage mode, e.g., settle/refine/escape. \\
$F_s$ & Scalar accepted best-so-far objective value, $F_s=f(\widehat q_s)$. \\
$\mathcal J_s(a)$ & Probe-based planner score used for training the slow policy.  This notation avoids conflict with the objective $f$. \\
$\mathcal L$ & Training loss, composed of local, planner, best-so-far, port, memory, and budget terms. \\
\bottomrule
\end{tabular}
\label{tab:shape_notation}
\end{table}

\section{More Related Works}
\label{sec:more:related:works}
\paragraph{Hamiltonian dynamics and symplectic integration}

The advantage of constructing Hamiltonian dynamics and applying symplectic integration is twofold. On the one hand, one can introduce a pseudo-Hamiltonian system for descent dynamics with an associated kinetic energy.\cite{wibisono2016variational} propose to construct the Hamiltonian via a convex and smooth functional $\phi:\mathcal Q\rightarrow \mathbb{R}$ and with an associated \textit{Bregman divergence} term:
\begin{equation}
    D_{\phi}(y,x):= \phi(y)-\phi(x)-\langle\nabla \phi(x), y-x\rangle,
\end{equation}

which induces the following \textit{Bregman Lagrangian}:
\begin{equation}
    \mathcal{L}(q,v,t) =  e^{\alpha(t)+\gamma(t)}\left(D_{\phi}(q+e^{-\alpha(t)}v,q) - e^{\beta(t)}f(q)\right),
\end{equation}

and by Legendre transform, a \textit{Bregman Hamiltonian}:
\begin{equation}
     \mathcal{H}(q,p,t) =  e^{\alpha(t)+\gamma(t)}\left(D_{\phi^{*}}(e^{-\gamma(t)}p+\nabla \phi(q),\nabla \phi(q)) + e^{\beta(t)}f(q)\right),
     \label{eq:bregman:hamiltonian}
\end{equation}

where $\phi^{*}$ is the Fenchel conjugate:
\begin{equation}
   \phi^{*}(q):= \sup_{v\in T\mathcal Q} \langle v,q\rangle- \phi(q).
\end{equation}

Note that it can reduce to locally Riemannian flow if $\phi$ is strongly convex (i.e. Hessian is a valid Riemannian metric tensor and  $D_{\phi}(y,x) = \frac{1}{2}\|y-x\|_{\nabla^2\phi(x)}^2+o(\|y-x\|_2^2)$) and further reduce to Euclidean case if $\phi(q):=\frac{1}{2}\|q\|_2^2$. Moreover, it is a generalized time dependent Hamiltonian, where dissipation is incorporated in energy form.  To be specific, define the following \textit{ideal scaling conditions} stated in \cite{wibisono2016variational}:
\begin{equation}
    \dot{\beta}(t)\leq \exp(\alpha(t)),\quad \dot\gamma(t) = \exp(\alpha(t)).
\end{equation}
Then Thm 2.1 in \cite{wibisono2016variational} stated asymptotic convergence of the potential if dynamics follows the E-L equation, which is essentially a 2nd-order ODE of form:
\begin{equation}
\begin{aligned}
    \ddot{q}(t) &+ (e^{\alpha(t)}-\dot{\alpha}(t))\dot{q}(t) + e^{2\alpha(t)+\beta(t)}\left[\nabla^2\phi(q(t)+e^{-\alpha(t)}\dot{q}_t)\right]^{-1}\nabla f(q(t))=0.
\end{aligned}
\end{equation}

It is equivalent to the following first order ODE as well:
\begin{equation}
\begin{aligned}
    \frac{d}{dt}\nabla\phi(q(t)+e^{-\alpha(t)}\dot{q}(t))=-e^{\alpha(t)+\beta(t)}\nabla f(q(t)).
\end{aligned}
\end{equation}

In particular, if one set $C>0$, $r\in\mathbb{Z}_{+}$ and:
\begin{equation}
    \alpha(t)= \log r -\log t,\quad \beta(t)= r\log t +\log C,\quad \gamma(t)= r\log t.
\end{equation}

Then the 2nd-order ODE is:
\begin{equation}
    \ddot{q}(t) + \frac{r+1}{t}\dot{q}(t) + Cr^2t^{r-2}\left[\nabla^2\phi(q(t)+\frac{t}{r}\dot{q}(t))\right]^{-1}\nabla f(q(t))=0.
\end{equation}

The proposed discretization scheme in \cite{wibisono2016variational} follows an idea of Nesterov type of gradient descent dynamics, where a Crank-Nicolson
discretization of the position updates and a backward Euler discretization are coupled plus an auxiliary projection:
\begin{equation}
    \begin{aligned}
        q_{k+1}&=\frac{r}{k+r}z_k + \frac{k}{k+r} y_k,  \\
        y_{k+1}&=\underset{q\in \mathcal Q}{argmin} \left[f_{r-1}(q;q_k) + \frac{C_2}{\epsilon^r r}\|q-q_k\|_2^2\right], \\
         z_{k+1}&=\underset{z\in \mathcal Q}{argmin} \left[Cr k^{r-1}\langle \nabla f(y_k), z\rangle + \frac{1}{\epsilon^r}D_{\phi}(z,z_k)\right],
    \end{aligned}
    \label{eq:bregman:scheme}
\end{equation}
where $f_{r-1}(q;q_k)$ is Taylor's expansion $f_{r-1}$ of the potential $f$ up to order $r-1$ at $q_k$. It is heuristic, hard to compute, and non-autonomous.\cite{jordan2018dynamical, betancourt2018symplectic} bring another advantage by justifying a symplectic integration scheme, which approximates Hamiltonian dynamics and appears to provide an effective
and flexible way to obtain discrete-time approximations, compared with heuristic schemes proposed in \cite{wibisono2016variational} (and reiterated in \eqref{eq:bregman:scheme}). In Hamiltonian \eqref{eq:bregman:hamiltonian}, the effective scaling parameter of potential term $f(q)$ is $\exp(\alpha(t)+\beta(t)+\gamma(t))=Crt^{2r-1}$ and it blows up as $t\rightarrow \infty$. Therefore it is asymptotically stable if discrete dynamics is symplectic and $\phi$ is a strictly convex and smooth function.  

Another particular example to show the advantage of symplectic integration is the conforming GD scheme \cite{francca2020conformal, francca2020conformal2, ghirardelli2024optimization}. With appropriate \textit{conforming} numerical integration schemes adapted from symplectic schemes (especially Symplectic Euler and Symplectic Leapfrog), one can guarantee, in the discrete setting, that
$$dQ_k\wedge dP_k = e^{-\gamma\eta}dQ_{k-1}\wedge dP_{k-1}.$$

A conforming integration scheme can be formulated as composition of several individual flows. For instance, one defines a symplectic Euler step $\Phi_{C,\eta}:(q,p)\rightarrow (Q,P)$ as:
\begin{equation}
    Q = q + \eta\nabla_pH(q,P),\quad P=p- \eta\nabla_qH(q,P).
\end{equation}
On the other hand, the dissipation with a constant damping coefficient (denoted as $\gamma$) yields an ODE system $(\dot{q},\dot{p})=(0,-\gamma p),q(0)=q_0,p(0)=p_0$, whose one step integration along $(0,\eta)$, has an exact solution, which yields a dissipative step $\Phi_{D,\eta}:(q,p)\rightarrow (Q,P)$ as:
\begin{equation}
    Q = q,\quad P=e^{-\gamma \eta}p
\end{equation}
Therefore, a composition $\Phi_{D,\eta}\circ \Phi_{C,\eta}$ can be interpreted as:
\begin{equation}
\begin{cases}
    p_{k+1} = e^{-\gamma \eta} p_k - \eta \nabla_qH(q_k,p_{k+1}) \\
    q_{k+1} = q_k + \eta \nabla_pH(q_{k},p_{k+1})
\end{cases}
\end{equation}

This is a first-order conformal/symplectic Euler step over Euclidean phase space.

\paragraph{Casimir/IDA--PBC energy shaping as a constrained design program}
\label{subsec:prelim_casimir_idapbc}
We consider an input--state--output port-Hamiltonian (pH) system
\begin{align}
\dot x &= (J(x)-R(x))\nabla H(x) + G(x)u, 
\qquad
y = G(x)^\top \nabla H(x),
\label{eq:phs_plant}
\end{align}
with $J(x)=-J(x)^\top$ and $R(x)=R(x)^\top\succeq 0$. In this class, the Hamiltonian $H$ is a storage function and the system is passive (energy cannot increase faster than supplied power). IDA--PBC\cite{ortega2002interconnection} designs a static state feedback $u=\beta(x)+v$ such that the closed-loop is again pH:
\begin{align}
\dot x &= (J_d(x)-R_d(x))\nabla H_d(x) + G(x)v,
\qquad 
y' = G(x)^\top \nabla H_d(x),
\label{eq:ida_target}
\end{align}
where $J_d=-J_d^\top$, $R_d\succeq 0$, and $H_d$ has a strict minimum at the desired equilibrium $x^\star$.
A standard constructive route is to fix $(J_a,R_a)$ and solve the linear \emph{matching PDE} in $H_a$:
\begin{align}
H_d &= H + H_a,\quad J_d = J + J_a,\quad R_d = R + R_a,\\
G^\perp(x)\big(J_d(x)-R_d(x)\big)\nabla H_a(x)
&=
-\,G^\perp(x)\big(J_a(x)-R_a(x)\big)\nabla H(x),
\label{eq:matching_pde}
\end{align}
where $G^\perp(x)$ is a left annihilator of $G(x)$ (i.e., $G^\perp G=0$). This highlights that \emph{energy shaping} is a PDE-constrained problem. 
A closely related presentation \cite{hamroun2010control} writes the closed loop as
\begin{align}
\dot x &= (J_d(x)-R_d(x))\nabla H_d(x),
\\
J_d &= J + J_a,\;\; R_d = R + R_a,\;\; H_d = H + H_a,\;\; \nabla H_a = K(x),
\label{eq:hamroun_closedloop}
\end{align}
and obtains admissible $H_a$ by projecting a matching relation onto the orthogonal complement of the input map $G_u$:
\begin{align}
G_{u}^{\perp}(x)\big(J(x)+J_a(x)\big)\nabla H_a(x)
=
-\,G_u^{\perp}(x)\,J_a(x)\nabla H(x).
\label{eq:hamroun_projected}
\end{align}
This makes explicit how the \emph{choice} of $G_u^\perp$ and the \emph{parametrization} of $J_a$ simplify (or complicate) solvability.

While IDA--PBC is commonly presented as a \emph{static} state-feedback shaping program
\(u=\beta(x)+v\) that enforces a desired pH form \eqref{eq:ida_target} by solving a matching PDE
(e.g.\ \eqref{eq:matching_pde}), Casimir-based designs proceed by \emph{dynamic interconnection}
of the plant with an auxiliary pH controller whose internal state generates \emph{invariants}
(Casimirs) for the \emph{interconnected} closed-loop.

Concretely, consider a dynamic pH controller with state \(x_c\in\mathbb{R}^{n_c}\),
Hamiltonian \(H_c(x_c)\), and port variables \((u_c,y_c)\):
\begin{align}
\dot x_c &= (J_c(x_c)-R_c(x_c))\nabla H_c(x_c) + G_c(x_c)\,u_c, 
\qquad 
y_c = G_c(x_c)^\top \nabla H_c(x_c).
\label{eq:phs_controller}
\end{align}
A power-preserving interconnection (one canonical choice) is
\begin{align}
u = -y_c + v,\qquad u_c = y,
\label{eq:power_interconnect}
\end{align}
so that the total stored energy \(H_{\mathrm{tot}}(x,x_c)=H(x)+H_c(x_c)\) is dissipated by
\(R,R_c\) and driven only by the external port \(v\).
Casimir-based shaping searches for functions \(C(x,x_c)\) (often vector-valued) such that along the
\emph{unforced} interconnection (\(v\equiv 0\)) one has
\begin{align}
\dot C(x,x_c)=0,
\qquad
\text{i.e., } C(x(t),x_c(t)) \equiv \text{const.}
\label{eq:casimir_invariant}
\end{align}
These invariants define a constraint manifold linking plant and controller states, typically of the form
\begin{align}
C(x,x_c) = \xi(x) - x_c = 0 \quad\Rightarrow\quad x_c = \xi(x),
\label{eq:casimir_manifold}
\end{align}
so that on the invariant manifold the closed-loop energy reduces to an \emph{effective} shaped energy
\begin{align}
H_{\mathrm{eff}}(x) \;=\; H(x) + H_c(\xi(x)),
\label{eq:heff}
\end{align}
which can be shaped via the choice of \(H_c\) and \(\xi\).

The key structural distinction is therefore:
\begin{itemize}
\item \textbf{IDA--PBC:} directly enforces a \emph{target pH form} for the plant state \(x\) under static feedback,
leading to a matching condition (PDE) that constrains \((J_a,R_a,H_a)\).
\item \textbf{Casimir approach:} enforces \emph{closed-loop invariants} under plant--controller interconnection.
Energy shaping occurs \emph{indirectly} through the invariant manifold \(x_c=\xi(x)\), yielding
\(H_{\mathrm{eff}}(x)\) in \eqref{eq:heff}.
\end{itemize}
This difference has practical consequences. Casimir-based designs can simplify shaping in some architectures
(because one can select \(H_c\) and \(\xi\) to obtain a desired \(H_{\mathrm{eff}}\)), but the price is that
\emph{existence of Casimirs imposes strong structural constraints} on admissible interconnections and damping.
In particular, for many dissipative plants (nontrivial \(R\)) and/or for restricted controller ports,
nontrivial Casimirs may fail to exist unless additional structure is imposed (e.g.\ specific factorization of \(R\),
or using power-preserving interconnection with no dissipation injection in the invariant directions).

\paragraph{Remark on Casimir invariants and controller-side locks}
\label{app:casimir_remark}

Casimirs become useful when the plant is interconnected with a dynamic
controller.  Let \(x_p\) denote plant variables and \(x_c\) controller variables.
A closed-loop Casimir has the form
\[
    C(x_p,x_c)
\]
and satisfies
\begin{equation}
\label{eq:closed_loop_casimir}
    J_{\mathrm{cl}}(x_p,x_c)
    \nabla C(x_p,x_c)
    =
    0 ,
\end{equation}
where \(J_{\mathrm{cl}}\) is the closed-loop interconnection matrix.  In
control-by-interconnection, such invariants can impose relations between plant
and controller states, for example
\[
    x_c = \Phi(x_p),
\]
thereby allowing the controller energy \(H_c(x_c)\) to shape the effective
closed-loop energy
\[
    H_{\mathrm{cl}}(x_p,x_c)
    =
    H_p(x_p)+H_c(x_c).
\]

The present method does not require this exact mechanism.  Our stagewise
navigator instead uses the following weaker and more implementation-faithful
structure:
\[
    (g_s,m_s,\mu_s)
    \longrightarrow
    H_{d,s}(q,p)
    \longrightarrow
    \text{dissipative local motion}.
\]
The slow planner and event-triggered memory update produce the frozen stage
context \((g_s,m_s,\mu_s)\).  The local controller then shapes the Hamiltonian
through \(U_{\mathrm{goal},s}\), \(U_{\mathrm{mem},s}\), and \(V_{\mathrm{bar},s}\),
and injects damping through \(D_s\) and \(K_s^d\).  The stage equilibrium is
therefore induced by energy shaping and damping injection, not by an exact
Casimir relation.

Consequently, Casimir invariants should be viewed as an optional extension,
not as a required part of the proposed algorithm.  A future dynamic-controller
variant could introduce a controller state \(x_c\) and enforce an approximate
or exact relation
\[
    C(q,p,x_c)=\mathrm{const.}
\]
to obtain a controller-side lock.  However, the current implementation and
analysis only require the stagewise IDA-PBC structure.
\section{Discrete optimizer updates and their relationship with a PH system}
\label{app:disc_ph_optimizers}

\paragraph{Notation.}
We write $g_k:=\nabla f(q_k)$ (or $g_k:=\nabla\ell(q_k;\xi_k)$ in the stochastic/minibatch case),
$\odot$ for elementwise product, and $\epsilon_k\sim\mathcal N(0,I)$.
The canonical pH template on $(q,p)$ uses $H(q,p)=f(q)+\tfrac12 p^\top M^{-1}p$,
interconnection $\dot q=\partial_p H$, $\dot p=-\partial_q H-\gamma\,\partial_p H$, and (optionally) diffusion in $p$.

\paragraph{Heavy-Ball / Polyak momentum.}
Discrete update (explicit damped integrator):
\begin{equation}
\label{eq:app_hb_disc}
\boxed{
\begin{aligned}
p_{k+1} &= (1-\gamma\eta)p_k - \eta\,g_k,\\
q_{k+1} &= q_k + \eta\,p_{k+1}.
\end{aligned}}
\end{equation}
This is a discretization of the damped pH flow on $(q,p)$ with Hamiltonian
$H(q,p)=f(q)+\tfrac12 p^\top M^{-1}p$ (typically $M=I$), canonical interconnection, and dissipation in $p$.

\paragraph{Nesterov accelerated gradient.}
A standard NAG form can be written as
\begin{equation}
\label{eq:app_nag_disc}
\boxed{
\begin{aligned}
\tilde q_k &= q_k + \alpha p_k,\\
p_{k+1} &= \beta\,p_k - \eta\,\nabla f(\tilde q_k),\\
q_{k+1} &= q_k + p_{k+1},
\end{aligned}}
\end{equation}
where $(\alpha,\beta)$ encode the look-ahead and momentum choices.
pH lens: same underlying interconnection as Heavy-Ball, but the internal force $\nabla_q H=\nabla f$
is \emph{measured} at the predicted configuration $\tilde q_k$.

\paragraph{AdaGrad.}
AdaGrad maintains an accumulated second moment and scales gradients:
\begin{equation}
\label{eq:app_adagrad_disc}
\boxed{
\begin{aligned}
v_{k+1} &= v_k + g_k\odot g_k,\\
q_{k+1} &= q_k - \eta\,\frac{g_k}{\sqrt{v_{k+1}}+\epsilon}.
\end{aligned}}
\end{equation}
pH lens (overdamped): $q$ follows a gradient flow $\dot q=-P(v)\nabla f(q)$ with
$P(v)=\mathrm{diag}\!\big((\sqrt v+\epsilon)^{-1}\big)\succ0$; $v$ is an auxiliary controller state driven by $g\odot g$.

\paragraph{RMSProp.}
RMSProp is the EMA version of AdaGrad:
\begin{equation}
\label{eq:app_rmsprop_disc}
\boxed{
\begin{aligned}
v_{k+1} &= \beta_2 v_k + (1-\beta_2)\,g_k\odot g_k,\\
q_{k+1} &= q_k - \eta\,\frac{g_k}{\sqrt{v_{k+1}}+\epsilon}.
\end{aligned}}
\end{equation}
pH lens: $v$ is a stable filter/controller state defining a time-varying diagonal metric (or dissipation gain).

\paragraph{AdaDelta \cite{zeiler2012adadelta}}
A consistent AdaDelta update is
\begin{equation}
\label{eq:app_adadelta_disc}
\boxed{
\begin{aligned}
v_{k+1} &= \rho v_k + (1-\rho)\,g_k\odot g_k,\\
\Delta q_k &= -\frac{\sqrt{m_k+\epsilon}}{\sqrt{v_{k+1}}+\epsilon}\odot g_k,\\
m_{k+1} &= \rho m_k + (1-\rho)\,\Delta q_k\odot \Delta q_k,\\
q_{k+1} &= q_k + \Delta q_k.
\end{aligned}}
\end{equation}
pH lens: $(m,v)$ are auxiliary dissipative states shaping the effective diagonal gain multiplying $\nabla f$.

\paragraph{Adam / AdamW.}
A bias-corrected Adam update can be written as
\begin{equation}
\label{eq:app_adam_disc}
\boxed{
\begin{aligned}
m_{k+1} &= \beta_1 m_k + (1-\beta_1)\,g_k,\\
v_{k+1} &= \beta_2 v_k + (1-\beta_2)\,g_k\odot g_k,\\
\hat m_{k+1} &= \frac{m_{k+1}}{1-\beta_1^{k+1}},\qquad
\hat v_{k+1} = \frac{v_{k+1}}{1-\beta_2^{k+1}},\\
q_{k+1} &= q_k - \eta\,\frac{\hat m_{k+1}}{\sqrt{\hat v_{k+1}}+\epsilon}.
\end{aligned}}
\end{equation}
AdamW adds decoupled weight decay: $q_{k+1}\leftarrow q_{k+1}-\eta\lambda q_k$.
pH lens: $z=(m,v)$ are controller states driven by $(g,g\odot g)$ and shape geometry/dissipation.

\paragraph{Lookahead.}
Let $\mathrm{OptStep}(\cdot)$ denote \emph{any} base optimizer (e.g., Adam) applied for $K$ inner steps.
Define the slow weights $\bar q$ and coupling parameter $\alpha\in(0,1]$.
A standard Lookahead cycle is
\begin{equation}
\label{eq:app_lookahead_disc}
\boxed{
\begin{aligned}
q^{(0)} &= \bar q_{r},\\
q^{(j+1)} &= \mathrm{OptStep}\big(q^{(j)}\big),\qquad j=0,1,\dots,K-1,\\
q_{r+1} &= q^{(K)},\\
\bar q_{r+1} &= \bar q_r + \alpha\,(q_{r+1}-\bar q_r),
\qquad q_{r+1}\leftarrow \bar q_{r+1}.
\end{aligned}}
\end{equation}
pH lens: augment state $(q,\bar q)$ with coupling energy $\tfrac{\lambda}{2}\|q-\bar q\|^2$;
the last line is a dissipative synchronization step that decreases this coupling storage.

\paragraph{SAM.}
A commonly used SAM step (first-order inner maximization) is
\begin{equation}
\label{eq:app_sam_disc}
\boxed{
\begin{aligned}
g_k &= \nabla f(q_k),\\
\delta_k &= \rho\,\frac{g_k}{\|g_k\|_2+\epsilon},\qquad \tilde q_k = q_k + \delta_k,\\
q_{k+1} &= q_k - \eta\,\nabla f(\tilde q_k).
\end{aligned}}
\end{equation}
pH lens: replace the base potential $f$ by a robustified energy
$H_\rho(q)\approx f(q+\delta^\star(q))$; operationally this is \emph{potential shaping}
using one extra gradient evaluation at $\tilde q_k$.

\paragraph{SGHMC.}
A basic SGHMC Euler update is
\begin{equation}
\label{eq:app_sghmc_disc}
\boxed{
\begin{aligned}
p_{k+1} &= (1-\gamma\eta)p_k - \eta\,g_k + \sqrt{2\gamma\eta}\,M^{1/2}\epsilon_k,\\
q_{k+1} &= q_k + \eta\,M^{-1}p_{k+1}.
\end{aligned}}
\end{equation}
(Equivalent parameterizations appear in the literature depending on whether $M$ multiplies the noise or is absorbed into $p$.)
pH lens: stochastic pH on $(q,p)$ with Hamiltonian $H=f+\tfrac12 p^\top M^{-1}p$ and diffusion in $p$.

\paragraph{Shampoo.}
For a matrix-shaped parameter $q$ and gradient $g_k$ of matching shape, Shampoo maintains
\begin{equation}
\label{eq:app_shampoo_disc}
\boxed{
\begin{aligned}
L_{k+1} &= L_k + g_k g_k^\top,\qquad R_{k+1} = R_k + g_k^\top g_k,\\
\hat g_k &= L_{k+1}^{-1/4}\, g_k\, R_{k+1}^{-1/4},\\
q_{k+1} &= q_k - \eta\,\hat g_k,
\end{aligned}}
\end{equation}
(optionally with momentum/EMA variants on $L,R$).
pH lens: $(L,R)$ are controller states driven by $g g^\top$ and $g^\top g$; they define a structured geometry operator.

\paragraph{SOAP.}
SOAP applies diagonal adaptation in a rotating eigen-basis (schematically):
\begin{equation}
\label{eq:app_soap_disc}
\boxed{
\begin{aligned}
L_{k+1} &= \beta_2 L_k + (1-\beta_2)\,g_k g_k^\top,\qquad
R_{k+1} = \beta_2 R_k + (1-\beta_2)\,g_k^\top g_k,\\
(Q_L,\Lambda_L) &= \mathrm{eig}(L_{k+1}),\qquad (Q_R,\Lambda_R)=\mathrm{eig}(R_{k+1}),\\
\hat g_k &= Q_L^\top g_k Q_R,\\
\hat v_{k+1} &= \beta_2 \hat v_k + (1-\beta_2)\,\hat g_k\odot \hat g_k,\\
q_{k+1} &= q_k - \eta\,Q_L\Big(\frac{\hat g_k}{\sqrt{\hat v_{k+1}}+\epsilon}\Big)Q_R^\top.
\end{aligned}}
\end{equation}
pH lens: $z=(Q_L,Q_R,\hat v)$ forms an auxiliary chart/metric state; the primal update is a charted diagonal descent.

\paragraph{Sophia.}
A representative Sophia-style step can be expressed as
\begin{equation}
\label{eq:app_sophia_disc}
\boxed{
\begin{aligned}
m_{k+1} &= \beta_1 m_k + (1-\beta_1)\,g_k,\\
h_{k+1} &= \beta_2 h_k + (1-\beta_2)\,\hat h_k,\qquad
\hat h_k \approx \mathrm{diag}\big(\nabla^2f(q_k)\big)\;\;(\text{proxy}),\\
d_{k+1} &= \mathrm{clip}\!\Big(\frac{m_{k+1}}{h_{k+1}+\epsilon},\,\rho\Big),\\
q_{k+1} &= q_k - \eta\,d_{k+1}.
\end{aligned}}
\end{equation}
pH lens: $h$ is an auxiliary curvature/geometry state; clipping enforces bounded power/step (a dissipative bound).


\paragraph{Lion/Lion-K.}
A representative Lion update can be written with one stored EMA and one look-ahead EMA direction:
\begin{equation}
\label{eq:app_lion_disc}
\boxed{
\begin{aligned}
c_{k+1} &= \beta_1 m_k + (1-\beta_1)\,g_k,\\
q_{k+1} &= q_k - \eta\big(\mathrm{sign}(c_{k+1}) + \lambda\,q_k\big),\\
m_{k+1} &= \beta_2 m_k + (1-\beta_2)\,g_k .
\end{aligned}}
\end{equation}
Here $g_k=\nabla f(q_k)$, $\beta_1,\beta_2\in(0,1)$ are EMA coefficients, and $\lambda\ge 0$ denotes optional decoupled weight decay.  The variable $c_{k+1}$ is the interpolated momentum used for the step, while the stored EMA state is $m_{k+1}$. \emph{Lion-K} refers the general construction scheme that is replacing the normalized/variance-scaled direction
$\frac{m}{\|m\|}$ or $\frac{m}{\sqrt v+\epsilon}$ by the much cheaper and more aggressive \emph{coordinate-wise sign} $\mathrm{sign}(m)$,
together with carefully tuned momentum/weight-decay, can match or improve empirical performance (especially in large-scale training)
while removing the need for second-moment estimation and its associated memory/compute overhead.
From a pH lens, Lion fits as an overdamped descent on $f$ with a \emph{bounded, nonsmooth constitutive law} for the velocity:
the effective “mobility” maps gradients to a bounded direction, which can be seen as enforcing an implicit port/step bound.

\paragraph{Muon}
Muon (MomentUm Orthogonalized by Newton--Schulz) is defined for \emph{matrix-shaped} parameters
(e.g.\ $W\in\mathbb{R}^{n\times m}$ in hidden layers).
At iteration $t$:
\begin{equation}
\label{eq:app_muon_disc}
\boxed{
\begin{aligned}
G_t &:= \nabla_W \mathcal{L}_t(W_{t-1}),\\
B_t &:= \mu B_{t-1} + G_t,\\
O_t &:= \mathrm{NewtonSchulz5}(B_t),\\
W_t &:= W_{t-1} - \eta_t\, O_t.
\end{aligned}
}
\end{equation}
Here $\mathrm{NewtonSchulz5}(\cdot)$ is a fixed-step Newton--Schulz matrix iteration that approximately
computes the \emph{semi-orthogonal projection} (polar factor) of its input:
\[
\mathrm{Ortho}(G)=\arg\min_{O}\{\|O-G\|_F:\ O^\top O=I\ \text{or}\ OO^\top=I\},
\]
equivalently $\mathrm{Ortho}(G)=UV^\top$ if $USV^\top$ is the SVD of $G$ (up to the transpose convention).

\section{SHAPE implementation details}
\label{app:shape_implementation}

This appendix section records the concrete optimizer interface used in the experiments:
the two-timescale algorithm, oracle features, event interface, implemented local
port-Hamiltonian update, memory representation, and training losses.

\paragraph{Notation synchronization with the main text.}
The appendix uses $f_\tau$ only to denote a sampled task objective from a task family; for a fixed test problem this is the same object denoted $f$ in the main text.  The symbol $\bar q_s$ denotes the stage anchor or local target, while $g(q)$ or $\widetilde g_{s,n}$ denotes a gradient/force observation.  The frozen shaped potential is written
\[
    U_{\tau,s}(q) = f_\tau(q)+U_s^{\rm shp}(q),
    \qquad
    U_s^{\rm shp}(q)
    =U_{\rm goal,s}(q;\bar q_s)+U_{\rm mem,s}(q;m_s^0)+V_{\rm bar,s}(q;m_s^0,\ell_s).
\]
Here $U_{\rm goal,s}$ is the local anchoring term, $U_{\rm mem,s}$ is the memory-induced reshaping term, and $V_{\rm bar,s}$ is a mode-dependent exclusion/barrier term.  The actual port input is $u_s^{\rm port}=u_s^{\rm shp}-K_s^dy_s$; in proofs where only one input appears, $u_s$ is shorthand for $u_s^{\rm port}$.

\subsection{Algorithm sketch and two-timescale viewpoint}
\label{subsec:dp_algorithm_sketch}

The proposed optimizer is a hybrid two-timescale closed-loop system with three components:
\begin{enumerate}
    \item a slow \emph{global planning policy} $\pi^{\mathrm G}_\phi$ that proposes a stage anchor $\bar q_s$ at event times,
    \item a fast \emph{local port-Hamiltonian policy} $\pi^{\mathrm L}_\psi$ that moves the state $(q,p)$ under the current shaped potential during one local stage,
    \item an \emph{event-triggered update map} $\Pi_\eta$ that decides when to stop the current local stage, summarize the observed trajectory, update memory, and replan.
\end{enumerate}
The global policy is responsible for large-scale navigation and basin selection, whereas the local policy is responsible for short-horizon reachability and stable integration. This separation makes the exploration--exploitation split explicit.

\begin{algorithm}[ht]
\caption{SHAPE}
\label{alg:dual_policy_shape}
\begin{algorithmic}[1]
\State \textbf{Input:} initial plant state $(q_0,p_0)$, initial slow memory $m_0$, task oracle $\mathcal O_\tau$
\State \textbf{Output:} accepted iterate sequence $\{\widehat q_s\}_{s\ge 0}$ and trajectory diagnostics
\State Set stage index $s\gets 0$ and accepted iterate $\widehat q_0\gets q_0$ 
\State Build the initial planner context $\omega_0$ and compute $(\bar q_0,\mu_0,\bar\alpha_0^J,\bar\alpha_0^R,\bar\kappa_0^{\mathrm{goal}})=\pi^{\mathrm G}_\phi(\omega_0)$ \Comment{see \S\ref{subsec:dp_oracle_events}, Eqs.~\eqref{eq:dp_context}--\eqref{eq:dp_subgoal}}
\While{stopping criterion not met}
    \State Initialize stage $s$ from $(q_s^0,p_s^0,m_s^0)$ \Comment{see \S\ref{subsec:dp_local_ivp}, Eqs.~\eqref{eq:dp_stage_initial_state}--\eqref{eq:dp_cross_stage_continuity}}
    \State Call Algorithm~\ref{alg:dp_local_stage_rollout} with frozen stage variables $(\bar q_s,\mu_s,\bar\alpha_s^J,\bar\alpha_s^R,\bar\kappa_s^{\mathrm{goal}},m_s^0)$
    \State Receive the stage terminal state $(q_s^+,p_s^+)$ and summary $\mathcal S_s$ \Comment{the local one-step map is detailed in \S\ref{subsec:dp_discrete_pc_update}, Eq.~\eqref{eq:dp_local_one_step_map}}
    \State Let $\widetilde q_s\in\operatorname*{arg\,min}_{0\le n\le N_s}f_\tau(q_{s,n})$ and update $\widehat q_{s+1}\in\operatorname*{arg\,min}\{f_\tau(\widehat q_s),f_\tau(\widetilde q_s)\}$. \Comment{best-so-far acceptance over the executed stage}
    \State Call Algorithm~\ref{alg:dp_event_interface} to obtain $(m_{s+1}^0,\bar q_{s+1},\mu_{s+1},\bar\alpha_{s+1}^J,\bar\alpha_{s+1}^R,\bar\kappa_{s+1}^{\mathrm{goal}})$ \Comment{see \S\ref{subsec:dp_oracle_events}, Eqs.~\eqref{eq:dp_event_update}--\eqref{eq:dp_memory_write_rule}}
    \State Set $s\gets s+1$
\EndWhile
\State \textbf{return} the accepted iterate sequence $\{\widehat q_s\}$ and recorded diagnostics
\end{algorithmic}
\end{algorithm}

\begin{algorithm}[ht]
\caption{SHAPE-inner}
\label{alg:dp_local_stage_rollout}
\begin{algorithmic}[1]
\State \textbf{Input:} $(q_s^0,p_s^0,m_s^0,\bar q_s,\mu_s,\bar\alpha_s^J,\bar\alpha_s^R,\bar\kappa_s^{\mathrm{goal}})$
\State Freeze the stage variables and initialize $(q_{s,0},p_{s,0})\gets(q_s^0,p_s^0)$ \Comment{the frozen-stage IVP is defined in \S\ref{subsec:dp_local_ivp}, Eqs.~\eqref{eq:dp_local_ivp_compact}--\eqref{eq:dp_stage_terminal_state}}
\State Initialize the stage summary accumulator $\mathcal S_s$
\For{$n=0,1,\dots,N_s-1$}
    \State Query the oracle and memory readout to form $z_{s,n}$ \Comment{see \S\ref{subsec:dp_oracle_events}, Eq.~\eqref{eq:dp_assimilated_signal}}
    \State Evaluate the local structural heads and assemble $J_{s,n},R_{s,n}$ \Comment{see \S\ref{subsec:dp_local_ivp}, Eqs.~\eqref{eq:dp_local_heads}--\eqref{eq:dp_local_ph_form}}
    \State Advance one fast step with $x_{s,n+1}=\Phi^{h_s}_{\tau,\psi}(x_{s,n};\bar q_s,m_s^0,\mu_s)$ \Comment{see \S\ref{subsec:dp_discrete_pc_update}, Eqs.~\eqref{eq:dp_dissipative_substep}--\eqref{eq:dp_local_one_step_map}}
    \State Update $\mathcal S_s$ with progress, stagnation, and memory-write statistics \Comment{the summary variables are specified in \S\ref{subsec:dp_oracle_events}, Eq.~\eqref{eq:dp_event_summary}}
\EndFor
\State Set $(q_s^+,p_s^+)\gets(q_{s,N_s},p_{s,N_s})$ \Comment{see Eq.~\eqref{eq:dp_numerical_stage_terminal_state}}
\State \textbf{return} $(q_s^+,p_s^+,\mathcal S_s)$
\end{algorithmic}
\end{algorithm}

\begin{algorithm}[ht]
\caption{SHAPE-outer( $\Pi_\eta$ update)}
\label{alg:dp_event_interface}
\begin{algorithmic}[1]
\State \textbf{Input:} terminal state $(q_s^+,p_s^+)$, frozen memory $m_s^0$, stage summary $\mathcal S_s$
\State Detect the event mode (settle / refine / escape) from $\mathcal S_s$ \Comment{see \S\ref{subsec:dp_oracle_events}, Eq.~\eqref{eq:dp_event_indicator}}
\State Update memory by the nonparametric write rule $m_s^0\mapsto m_{s+1}^0$ when the trigger is active \Comment{see Eqs.~\eqref{eq:dp_event_update}--\eqref{eq:dp_memory_write_rule} and \S\ref{subsec:dp_memory_repr}}
\State Build the next planner context $\omega_{s+1}=\Xi(q_s^+,p_s^+,m_{s+1}^0,\mathcal S_s)$ \Comment{see Eq.~\eqref{eq:dp_context} with the next-stage state substituted in}
\State Compute $(\bar q_{s+1},\mu_{s+1},\bar\alpha_{s+1}^J,\bar\alpha_{s+1}^R,\bar\kappa_{s+1}^{\mathrm{goal}})=\pi^{\mathrm G}_\phi(\omega_{s+1})$ \Comment{same planner map as in Eq.~\eqref{eq:dp_subgoal}}
\State \textbf{return} $(m_{s+1}^0,\bar q_{s+1},\mu_{s+1},\bar\alpha_{s+1}^J,\bar\alpha_{s+1}^R,\bar\kappa_{s+1}^{\mathrm{goal}})$
\end{algorithmic}
\end{algorithm}

Algorithm~\ref{alg:dual_policy_shape} makes the cross-stage transition explicit: the next local stage starts from the previous terminal plant state $(q_s^+,p_s^+)$, while the slow controller variables are updated only at the event interface. This is the key design feature behind the dual-policy formulation.

\subsection{Oracle, planner context, and event times}
\label{subsec:dp_oracle_events}

The current implementation assumes direct access to a local oracle
\begin{equation}
\label{eq:dp_oracle}
\mathcal O_\tau(q)
=
\bigl(f_\tau(q),\nabla f_\tau(q)\bigr).
\end{equation}
In the code, the retained observation at local step $n$ of stage $s$ is the concatenated feature vector
\begin{equation}
\label{eq:dp_assimilated_signal}
z_{s,n}
=
\Gamma\bigl(\mathcal O_\tau(q_{s,n}),m_s^0\bigr)
=
\bigl[q_{s,n},\,p_{s,n},\,\nabla f_\tau(q_{s,n}),\,f_\tau(q_{s,n}),\,\|\nabla f_\tau(q_{s,n})\|,\,d_\tau,\,\mathrm{MemRead}(m_s^0,q_{s,n})\bigr],
\end{equation}
where $d_\tau$ is the task descriptor and $m_s^0$ is the frozen stage memory. The present implementation does not maintain a separate intra-stage filter state. Equivalently, one may regard the information state as frozen within each stage,
\begin{equation}
\label{eq:dp_xi_predict}
\xi_{s,n+1}^- = \xi_{s,n+1} = \xi_s^0,
\end{equation}
so that the local controller is conditioned directly on the current oracle query and the memory readout. This is a deliberate simplification relative to the predictor--corrector variant discussed earlier.

In the current implementation, event times are placed on a fixed slow clock. If $H_{\mathrm{evt}}\in\mathbb N$ denotes the event horizon and $h$ the fast-step size, then
\begin{equation}
\label{eq:dp_next_event_time}
\tau_{s+1}=\tau_s + H_{\mathrm{evt}}h,
\qquad H_{\mathrm{evt}}=\texttt{event\_horizon}.
\end{equation}
This corresponds to using a fixed stage budget $b_s=H_{\mathrm{evt}}$ in the implementation.  The more general main-text notation with a planner-selected $b_s$ is recovered by replacing $H_{\mathrm{evt}}$ with a stage-dependent horizon $H_s=b_s$.
At the beginning of stage $s$, we form the planner context
\begin{equation}
\label{eq:dp_context}
\omega_s
=
\Xi\bigl(z_{s,0}\bigr)
=
\Xi\bigl(q_s^0,p_s^0,\nabla f_\tau(q_s^0),f_\tau(q_s^0),m_s^0,d_\tau\bigr),
\end{equation}
and the global planner outputs the stage variables
\begin{equation}
\label{eq:dp_subgoal}
(\bar q_s,\mu_s,\bar\alpha_s^J,\bar\alpha_s^R,\bar\kappa_s^{\mathrm{goal}},b_s)
=
\pi^{\mathrm G}_\phi(\omega_s),
\end{equation}
where $\bar q_s\in\mathbb R^d$ is the anchor, $\mu_s\in\{\mathrm{settle},\mathrm{refine},\mathrm{escape}\}$ is the stage mode, $\bar\alpha_s^J,\bar\alpha_s^R\in(0,1)$ are stagewise modulation coefficients for the antisymmetric and dissipative channels, $\bar\kappa_s^{\mathrm{goal}}>0$ is a stagewise goal-attraction gain, and $b_s$ is the stage oracle budget. The planner is therefore responsible only for slow-scale navigation decisions; it does not directly update the plant state.

The choice of $\bar q_s$ is implementation-dependent but always local.  During
training we supervise $\bar q_s$ with the probe rule in
\S\ref{subsec:dp_training}: generate candidates in a trust region around
$q_s^0$, score them by local improvement, memory novelty, and risk, and set the
teacher anchor to the highest-scoring candidate.  At test time no probe teacher
or known optimum is used; $\bar q_s$ is predicted directly by
$\pi^{\mathrm G}_\phi(\omega_s)$ and is clipped when required to the feasible box or
trust region.  In all theoretical statements below, $\bar q_s$ is treated as a
frozen stage parameter produced by this event map, so the proofs do not require
that $\bar q_s$ equal a local or global minimizer of $f_\tau$.

In the implementation the event-triggered update map $\Pi_\eta$ is hybrid: its learned part is the planner $\pi^{\mathrm G}_\phi$, whereas its trigger and memory-write logic are nonparametric. Let
\begin{equation}
\label{eq:dp_event_indicator}
\chi_s^{\mathrm{write}}
=
\mathbf 1\bigl\{\mu_s=\mathrm{escape}\ \text{or}\ \mathrm{stall}_s=1\bigr\},
\end{equation}
where $\mathrm{stall}_s$ is detected from low gradient norm, low momentum norm, and lack of short-horizon progress. Then
\begin{equation}
\label{eq:dp_event_update}
\Pi_\eta:
(\omega_s,m_s^0)
\mapsto
\bigl(\bar q_s,\mu_s,\bar\alpha_s^J,\bar\alpha_s^R,\bar\kappa_s^{\mathrm{goal}},m_{s+1}^0\bigr),
\end{equation}
with
\begin{equation}
\label{eq:dp_memory_write_rule}
m_{s+1}^0
=
\begin{cases}
\mathcal U_{\eta}^{\mathrm{mem}}\bigl(m_s^0,\mathcal S_s\bigr), & \chi_s^{\mathrm{write}}=1,\\
m_s^0, & \chi_s^{\mathrm{write}}=0,
\end{cases}
\end{equation}
and event summary
\begin{equation}
\label{eq:dp_event_summary}
\mathcal S_s
=
\bigl(q_s^{\mathrm{evt}},p_s^{\mathrm{evt}},f_\tau(q_s^{\mathrm{evt}}),\nabla f_\tau(q_s^{\mathrm{evt}}),\mu_s\bigr).
\end{equation}
Here $\eta$ denotes the fixed event hyperparameters---event horizon, stall thresholds, probe radius, and memory-update weights---rather than a separate trainable neural module.

\subsection{Stagewise local initial value problem and the implemented choice of $J$ and $R$}
\label{subsec:dp_local_ivp}

The local state of stage $s$ is the second-order plant state
\[
x_s=(q_s,p_s),
\qquad q_s,p_s\in\mathbb R^d,
\]
with initialization
\begin{equation}
\label{eq:dp_stage_initial_state}
x_s(0)=x_s^0=(q_s^0,p_s^0),
\qquad m_s(0)=m_s^0.
\end{equation}
For the first stage we use the user-specified initialization,
\begin{equation}
\label{eq:dp_stage0_init}
(q_0^0,p_0^0,m_0^0)=(q_0,p_0,m_0),
\end{equation}
while every later stage inherits the previous terminal plant state,
\begin{equation}
\label{eq:dp_cross_stage_continuity}
q_s^0=q_{s-1}^+,
\qquad
p_s^0=p_{s-1}^+,
\qquad s\ge 1.
\end{equation}
Thus the stage transition is continuous in the physical state, and only the slow planning variables are refreshed at event times.

The local dynamics are trained and implemented inside a fixed canonical mechanical interconnection. Specifically, the Hamiltonian used by the fast controller is
\begin{equation}
\label{eq:dp_stage_hamiltonian}
\mathcal H^{\tau}_{\psi,s}(q,p;\bar q_s,m_s^0,\mu_s)
=
\frac12 p^\top M_{s,n}^{-1}p
+
U_{\tau,s,n}(q),
\end{equation}
with
\begin{align}
\label{eq:dp_stage_potential_discrete}
U_{\tau,s,n}(q)
&:=
f_\tau(q)+U^{\rm shp}_{s,n}(q),
\\
U^{\rm shp}_{s,n}(q)
&:=
\frac12\kappa_{s,n}^{\mathrm{goal}}\|q-\bar q_s\|^2
+
U_{\rm mem,s}(q;m_s^0)
+
V_{\rm bar,s}(q;m_s^0,\ell_s).
\end{align}
When memory or barrier shaping is disabled, the corresponding term is set to zero. Here $M_{s,n}=\operatorname{diag}(m_{s,n})\succ0$ and
\begin{equation}
\label{eq:dp_goal_gain_split}
\kappa_{s,n}^{\mathrm{goal}} = \bar\kappa_s^{\mathrm{goal}} + \kappa_{s,n}^{\mathrm{loc}},
\end{equation}
with a stagewise contribution from the planner and a local correction from the controller. In the present implementation the task energy $f_\tau$ is queried directly from the oracle; there is no hidden force-to-goal term involving an unknown optimizer.

The crucial design choice is that the antisymmetric and dissipative parts are explicit but structured. The local controller, evaluated on the observation $z_{s,n}$, outputs
\begin{equation}
\label{eq:dp_local_heads}
\bigl(m_{s,n},\kappa_{s,n}^{\mathrm{loc}},u_{s,n}^{\mathrm{port}},U_{s,n}^{\Omega},V_{s,n}^{\Omega},B_{s,n}^{D},d_{s,n}^{D}\bigr)
=
\pi^{\mathrm L}_\psi(z_{s,n},\bar q_s,\mu_s,\bar\alpha_s^J,\bar\alpha_s^R,\bar\kappa_s^{\mathrm{goal}}),
\end{equation}
where $U_{s,n}^{\Omega},V_{s,n}^{\Omega},B_{s,n}^{D}\in\mathbb R^{d\times r}$ and $d_{s,n}^{D}\in\mathbb R_+^d$. These heads define the low-rank skew operator
\begin{equation}
\label{eq:dp_skew_operator}
\Omega_{s,n}(v)
=
U_{s,n}^{\Omega}\bigl((V_{s,n}^{\Omega})^\top v\bigr)
-
V_{s,n}^{\Omega}\bigl((U_{s,n}^{\Omega})^\top v\bigr),
\end{equation}
and the low-rank-plus-diagonal positive-semidefinite operator
\begin{equation}
\label{eq:dp_psd_operator}
D_{s,n}(v)
=
B_{s,n}^{D}\bigl((B_{s,n}^{D})^\top v\bigr)
+
d_{s,n}^{D}\odot v.
\end{equation}
Equation~\eqref{eq:dp_skew_operator} is skew by construction, while \eqref{eq:dp_psd_operator} is positive semidefinite by construction. This is the main reason the implementation can meaningfully test whether stagewise $J$/$R$ structure is learnable.

To expose the corresponding port-Hamiltonian structure, define the velocity variable $v_{s,n}=M_{s,n}^{-1}p_{s,n}$ and the mode-dependent effective gains
\begin{equation}
\label{eq:dp_mode_scaled_alphas}
\alpha_{s,n}^J = \bar\alpha_s^J\,\beta_J(\mu_s),
\qquad
\alpha_{s,n}^R = \bar\alpha_s^R\,\beta_R(\mu_s),
\end{equation}
where in code the weights $\beta_J,\beta_R$ are fixed scalars for settle/refine/escape. Then the lower-right block of the stagewise interconnection is the skew operator $\alpha_{s,n}^J\Omega_{s,n}$ and the lower-right block of the dissipation is $\alpha_{s,n}^R D_{s,n}$. Equivalently, the implemented local operators are
\begin{equation}
\label{eq:dp_local_ph_form}
J_{s,n}
=
\begin{bmatrix}
0 & I\\
-I & \alpha_{s,n}^J\Omega_{s,n}
\end{bmatrix},
\qquad
R_{s,n}
=
\begin{bmatrix}
0 & 0\\
0 & \alpha_{s,n}^R D_{s,n}
\end{bmatrix},
\end{equation}
with $J_{s,n}^\top=-J_{s,n}$ and $R_{s,n}\succeq0$ on the momentum channel.

Therefore one local stage solves the initial value problem
\begin{equation}
\label{eq:dp_local_ivp_compact}
\dot x_s(t)
=
F^{\tau,\mathrm L}_{\psi}\bigl(x_s(t);\bar q_s,m_s^0,\mu_s\bigr),
\qquad
x_s(0)=x_s^0,
\qquad
t\in[0,T_s],
\end{equation}
with explicit coordinates
\begin{align}
\dot q_s &= v_s = M_s^{-1}p_s,
\label{eq:dp_local_qdot}
\\
\dot p_s &= -\nabla_q U_{\tau,s}(q_s)
+ \alpha_s^J\Omega_s(v_s) - \alpha_s^R D_s(v_s) + u_s^{\mathrm{port}},
\label{eq:dp_local_pdot}
\end{align}
where all stage and step indices are suppressed for readability. In the present implementation there is no separate fast controller state update,
\begin{equation}
\label{eq:dp_local_xidot}
\dot \xi_s = 0,
\end{equation}
and oracle information enters directly through reevaluation of $f_\tau(q)$ and $\nabla f_\tau(q)$ at every fast step.

The terminal state of stage $s$ is
\begin{equation}
\label{eq:dp_stage_terminal_state}
x_s^+ := x_s(T_s)=(q_s^+,p_s^+),
\end{equation}
which is passed to the next stage via \eqref{eq:dp_cross_stage_continuity}.

\subsection{Stagewise structured discrete local update}
\label{subsec:dp_discrete_pc_update}

The present implementation uses a single structured explicit step rather than a St\"ormer--Verlet predictor--corrector solver. Let $h_s$ be the fast step size and write
\[
x_{s,n}=(q_{s,n},p_{s,n}),
\qquad n=0,\dots,N_s-1.
\]
Given the frozen stage variables $(\bar q_s,\mu_s,\bar\alpha_s^J,\bar\alpha_s^R,\bar\kappa_s^{\mathrm{goal}},m_s^0)$, one step proceeds as follows. First the current oracle is queried and the local heads \eqref{eq:dp_local_heads} are evaluated. We retain the notation
\begin{equation}
\label{eq:dp_observation_model}
\widehat z_{s,n+1}=z_{s,n+1}=\Gamma\bigl(\mathcal O_\tau(q_{s,n}),m_s^0\bigr),
\end{equation}
so that there is no separate prediction-correction mismatch in the implemented code. Accordingly,
\begin{equation}
\label{eq:dp_innovation}
r_{s,n+1}=0.
\end{equation}
\begin{equation}
\label{eq:dp_xi_correct}
\xi_{s,n+1}=\xi_s^0.
\end{equation}
Second, using \eqref{eq:dp_local_qdot}--\eqref{eq:dp_local_pdot}, the momentum update is
\begin{equation}
\label{eq:dp_dissipative_substep}
p_{s,n+1}
=
\Pi_{[-p_{\max},p_{\max}]}
\Bigl(
 p_{s,n}
 + h_s\bigl[-\nabla_q U_{\tau,s,n}(q_{s,n})
 + \alpha_{s,n}^J\Omega_{s,n}(v_{s,n})
 - \alpha_{s,n}^R D_{s,n}(v_{s,n})
 + u_{s,n}^{\mathrm{port}}\bigr]
\Bigr),
\end{equation}
where $\Pi_{[-p_{\max},p_{\max}]}$ denotes componentwise clipping. The configuration update is then
\begin{equation}
\label{eq:dp_half_kick_plus}
q_{s,n+1}
=
\Pi_{[-q_{\max},q_{\max}]}
\bigl(q_{s,n} + h_s M_{s,n}^{-1}p_{s,n+1}\bigr),
\end{equation}
with componentwise clipping $\Pi_{[-q_{\max},q_{\max}]}$ and
\begin{equation}
\label{eq:dp_q_update_final}
q_{s,n+1}^- = q_{s,n+1},
\qquad
p_{s,n+\frac12}^- = p_{s,n+1}.
\end{equation}
for notational compatibility with the earlier split notation. Hence the one-step map implemented in the code is
\begin{equation}
\label{eq:dp_local_one_step_map}
x_{s,n+1}=\Phi_{\tau,\psi}^{h_s}\bigl(x_{s,n};\bar q_s,m_s^0,\mu_s\bigr),
\end{equation}
with $\Phi_{\tau,\psi}^{h_s}$ defined by \eqref{eq:dp_local_heads}, \eqref{eq:dp_skew_operator}, \eqref{eq:dp_psd_operator}, \eqref{eq:dp_dissipative_substep}, and \eqref{eq:dp_half_kick_plus}. The numerical terminal state is therefore
\begin{equation}
\label{eq:dp_numerical_stage_terminal_state}
x_s^+ = x_{s,N_s},
\qquad
(q_s^+,p_s^+)=(q_{s,N_s},p_{s,N_s}).
\end{equation}

\subsection{Planner memory and dimension-dependent representation}
\label{subsec:dp_memory_repr}

The planner operates on a slow memory state $m_s$ that summarizes large-scale geometry. In the two-dimensional case we use an explicit multigrid spatial memory,
\begin{equation}
\label{eq:dp_2d_multigrid_memory}
m_s = \bigl(m_s^{(0)},m_s^{(1)},\dots,m_s^{(L-1)}\bigr),
\qquad
m_s^{(\ell)}\in\mathbb R^{C_\ell\times H_\ell\times W_\ell},
\end{equation}
with readout by cell lookup at the current configuration. For $d>2$, we replace the explicit grid by a blockwise latent hierarchy,
\begin{equation}
\label{eq:dp_highd_latent_memory}
m_s = \bigl(m_s^{(0)},m_s^{(1)},\dots,m_s^{(L-1)}\bigr),
\qquad
m_s^{(\ell)}\in\mathbb R^{B_\ell\times C},
\end{equation}
where blocks group coordinates into fixed-size subsets. In both cases memory is read only on the slow stage clock and appears in the planner/local networks through the observation \eqref{eq:dp_assimilated_signal}.

In the implementation, the memory update backend $\mathcal U_{\eta}^{\mathrm{mem}}$ is nonparametric. In 2D it increments coarse-to-fine occupancy, running potential statistics, gradient norms, and best-so-far values on visited cells; in higher dimension it updates analogous blockwise summaries. A write occurs only when the stage is classified as escape or stalled via \eqref{eq:dp_event_indicator}. Consequently, the map $\Pi_\eta$ is only partially learned: planning variables are predicted by $\pi_\phi^{\mathrm G}$, whereas memory writes are governed by fixed event thresholds and summary accumulators.

\subsection{Training the local-global navigation pipeline and the event update map}
\label{subsec:dp_training}

The implementation trains the planner and local $J/R$ controller in three phases. Importantly, no training loss assumes that a global optimizer $q^\star$ is known. When synthetic tasks provide a known optimizer, it is used only for offline evaluation after training.

\paragraph{Probe-based stage teacher.}
At every event time the code builds a finite candidate set
\begin{equation}
\label{eq:dp_probe_candidates}
\mathcal C_s = \{c_{s,j}=q_s^0 + \rho_j d_j\}_{j=1}^{K_{\mathrm{probe}}},
\end{equation}
where $d_j$ are normalized gradient, momentum, and random orthogonal probe directions and $\rho_j$ are fixed radii. Each candidate is scored by
\begin{equation}
\label{eq:dp_probe_score}
S_s(c)
=
\widehat{\mathrm{Improve}}_s(c)
+ \lambda_{\mathrm{nov}}\,\mathrm{Novelty}(c;m_s^0)
- \lambda_{\mathrm{risk}}\,\mathrm{Risk}(c),
\end{equation}
with predicted improvement given by a one-step objective decrease, novelty computed from memory occupancy, and risk penalizing large excursions. The teacher anchor is
\begin{equation}
\label{eq:dp_teacher_anchor}
\bar q_s^{\mathrm T}\in\arg\max_{c\in\mathcal C_s} S_s(c),
\end{equation}
and the teacher mode label is
\begin{equation}
\label{eq:dp_probe_labels}
\mu_s^{\mathrm T}
=
\begin{cases}
\mathrm{refine}, & S_s(\bar q_s^{\mathrm T}) \text{ indicates clear improvement},\\
\mathrm{escape}, & S_s(\bar q_s^{\mathrm T}) \text{ is small and the stage is stalled or novel},\\
\mathrm{settle}, & \text{otherwise}.
\end{cases}
\end{equation}
This teacher uses only local oracle information and memory summaries.

\paragraph{Phase I: local $J/R$ pretraining.}
With planner outputs forced to the probe teacher $(\bar q_s^{\mathrm T},\mu_s^{\mathrm T})$, the local controller is trained to match a simple teacher force
\begin{equation}
\label{eq:dp_local_teacher_force}
F_s^{\mathrm T}
=
-\nabla f_\tau(q_s^0) - c_g(q_s^0-\bar q_s^{\mathrm T}) - c_p p_s^0 + F_{\mathrm{escape}}^{\mathrm T}(\mu_s^{\mathrm T}),
\end{equation}
by minimizing
\begin{equation}
\label{eq:dp_local_pretrain_loss}
\mathcal L_{\mathrm{loc}}^{\mathrm{pre}}
=
\sum_s \Bigl\|\frac{p_{s,1}-p_{s,0}}{h_s} - F_s^{\mathrm T}\Bigr\|^2
+ \lambda_{JR}\bigl(\|\Omega_{s,0}(v_{s,0})\| + \|D_{s,0}(v_{s,0})\|\bigr).
\end{equation}
This phase identifies the low-rank skew and PSD damping heads before the planner is trained jointly.

\paragraph{Phase II: supervised planner training.}
With the current local controller fixed, the planner is trained from probe labels by
\begin{equation}
\label{eq:dp_planner_supervised_loss}
\mathcal L_{\mathrm{plan}}^{\mathrm{sup}}
=
\lambda_{\mathrm{ce}}\,\mathrm{CE}(\hat\mu_s,\mu_s^{\mathrm T})
+
\lambda_{\mathrm{sg}}\,\mathrm{Huber}(\bar q_s,\bar q_s^{\mathrm T}),
\end{equation}
where $\hat\mu_s$ are the planner logits. This phase trains only the slow navigation decision variables inside $\Pi_\eta$.

\paragraph{Phase III: joint rollout training.}
The end-to-end rollout objective used in the code is
\begin{equation}
\label{eq:dp_training_objective}
\mathcal L_{roll}
=
\lambda_{\mathrm{term}}\mathcal L_{\mathrm{term}}
+
\lambda_{\mathrm{best}}\mathcal L_{\mathrm{best}}
+
\lambda_{\mathrm{prog}}\mathcal L_{\mathrm{prog}}
+
\mathcal L_{\mathrm{plan}}^{\mathrm{sup}}
+
\lambda_{\mathrm{ctrl}}\mathcal L_{\mathrm{ctrl}}
+
\lambda_{JR}\mathcal L_{JR} + \lambda_{port}\mathcal L_{port},
\end{equation}
where
\begin{align}
\mathcal L_{\mathrm{term}} &= \frac{1}{B}\sum_{i=1}^B \frac{f_\tau(q_{i,T})}{|f_\tau(q_{i,0})|+1},
\label{eq:dp_terminal_loss}
\\
\mathcal L_{\mathrm{best}} &= \frac{1}{B}\sum_{i=1}^B \frac{\min_n f_\tau(q_{i,n})}{|f_\tau(q_{i,0})|+1},
\label{eq:dp_best_loss}
\\
\mathcal L_{\mathrm{prog}} &= \frac{1}{B}\sum_{i=1}^B \frac1T\sum_{n=1}^{T}\|q_{i,n}-g_{s(i,n)}\|,
\label{eq:dp_progress_loss}
\\
\mathcal L_{\mathrm{ctrl}} &= \frac{1}{BT}\sum_{i,n}\|u_{i,n}^{\mathrm{port}}\|^2,
\label{eq:dp_control_loss}
\\
\mathcal L_{JR} &= \frac{1}{BT}\sum_{i,n}\Bigl(\|\Omega_{i,n}(v_{i,n})\| + \|D_{i,n}(v_{i,n})\| + \tfrac14\|u_{i,n}^{\mathrm{port}}\|\Bigr).
\label{eq:dp_jr_reg_loss}\\
\mathcal L_{\rm port}
    &=
    \frac{1}{BN}
    \sum_{i\in\mathcal B}
    \sum_{n=0}^{N-1}
    \bigl[
        \langle y_{s,n},u^{\rm port}_{s,n}\rangle 
    \bigr]_+^2,
    \quad
    [a]_+ := \max\{a,0\}.\label{eq:dp_port_power_penalty}
\end{align}
The terminal and best-so-far terms measure task performance, $\mathcal L_{\mathrm{prog}}$ ensures that the local controller actually realizes the planner's anchor, and $\mathcal L_{JR}$ regularizes the size of the learned $J/R$ corrections so that improvements are attributable to structured modulation rather than uncontrolled residual forcing.

\paragraph{What is and is not learned inside $\Pi_\eta$.}
In the current implementation, $\phi$ and $\psi$ are optimized by gradient descent, but the event thresholds and the memory backend inside $\Pi_\eta$ are fixed. Thus $\Pi_\eta$ is only partially learned: the planner outputs $(\bar q_s,\mu_s,\bar\alpha_s^J,\bar\alpha_s^R,\bar\kappa_s^{\mathrm{goal}})$ are learned, while event timing, write masks, and summary accumulation remain algorithmic. This distinction is important for interpreting the experiments: the code tests whether stagewise structured $J$ and $R$ are useful inside a learned local-global navigation loop, not whether a fully neural event map can be learned end-to-end.

\paragraph{How to balance the loss term} 
Throughout the paper, hyperparameters of SHAPE optimizer training are either fixed in the protocol or selected by the grid search. These losses train how the local motion is guided and how $\pi_{\phi}^G$ is updated, while the potential and pH template are balanced within each local rollout trajectories. Therefore, for each trained optimizers, there is no conventional hyper-tuned version compared with other conventional gradient descent system.

\subsection{Implemented stagewise structure of $J$ and $R$}
\label{app:dp_structure_JR}

The stagewise local model uses
\[
J_s=
\begin{bmatrix}
0 & I\\
-I & \Omega_s
\end{bmatrix},
\qquad
R_s=
\begin{bmatrix}
0 & 0\\
0 & D_s
\end{bmatrix},
\]
where $\Omega_s^\top=-\Omega_s$ and $D_s\succeq 0$. In the implementation,
\[
\Omega_s(v)=U_s^{\Omega}(V_s^{\Omega})^\top v-V_s^{\Omega}(U_s^{\Omega})^\top v,
\]
so the matrix representative
\[
\mathbf \Omega_s=U_s^{\Omega}(V_s^{\Omega})^\top-V_s^{\Omega}(U_s^{\Omega})^\top
\]
satisfies $\mathbf \Omega_s^\top=-\mathbf \Omega_s$. Likewise,
\[
D_s(v)=B_s^D(B_s^D)^\top v+d_s^D\odot v,
\qquad d_s^D\in\mathbb R_+^d,
\]
so the matrix representative
\[
\mathbf D_s=B_s^D(B_s^D)^\top+\operatorname{Diag}(d_s^D)
\]
satisfies $v^\top \mathbf D_s v\ge 0$ for all $v$. Therefore the lower-right block of $J_s$ is skew by construction and the lower-right block of $R_s$ is positive semidefinite by construction.

\section{Experimental Details}
\label{app:exp:details}

This appendix contains the details supporting the main experiment section.

\subsection{Gradient-Oracle Variants}
\label{app:exp:gradient_oracles}

The experiment section is organized so that task families can be combined with different local-information models. We write the oracle at iterate $q_k$ as $\mathcal O_\tau(q_k)$ and use the following variants.

\paragraph{Exact first-order oracle.}
The oracle returns the objective and its gradient,
\begin{equation}
  \mathcal O_\tau^{\mathrm{FO}}(q_k)=\bigl(f_\tau(q_k),\nabla f_\tau(q_k)\bigr).
  \label{eq:app:fo_oracle}
\end{equation}
This is the default for analytic synthetic functions and smooth differentiable physical objectives.

\paragraph{Noisy or mini-batch first-order oracle.}
The oracle returns an unbiased or approximately unbiased gradient estimate
\begin{equation}
  \tilde g_k = \nabla f_\tau(q_k)+\epsilon_k,
  \qquad
  \mathbb E[\epsilon_k\mid \mathcal F_k]=0,
  \qquad
  \mathbb E[\|\epsilon_k\|^2\mid \mathcal F_k]\le \sigma_g^2.
  \label{eq:app:noisy_fo_oracle}
\end{equation}
For finite-sum objectives, such as phase retrieval, this includes mini-batch gradients. For simulation objectives, this also covers stochastic rollout perturbations when the adjoint or automatic-differentiation gradient is estimated from a random rollout.

\paragraph{Adjoint or automatic-differentiation oracle.}
For differentiable simulator objectives of the form
\begin{equation}
  x_{t+1}=\varphi_\tau(x_t,u_t),
  \qquad
  q=(u_0,\ldots,u_{T-1}),
  \qquad
  f_\tau(q)=\Phi_\tau(x_T)+\sum_{t=0}^{T-1}\ell_\tau(x_t,u_t),
\end{equation}
we obtain $\nabla f_\tau(q)$ either by reverse-mode automatic differentiation through the rollout or by a discrete adjoint. We count one full adjoint gradient as one first-order oracle call, while reporting wall-clock time separately when relevant.

\paragraph{Zeroth-order or black-box oracle.}
When gradients are unavailable, the oracle returns only function values,
\begin{equation}
  \mathcal O_\tau^{\mathrm{ZO}}(q_k)=f_\tau(q_k)+\eta_k.
\end{equation}
A two-point random-direction estimator may be used:
\begin{equation}
  \widehat g_k
  =
  \frac{f_\tau(q_k+\mu u_k)-f_\tau(q_k-\mu u_k)}{2\mu}\,u_k,
  \qquad
  u_k\sim\mathrm{Unif}(\mathbb S^{d-1}).
  \label{eq:app:zo_oracle}
\end{equation}
For fair accounting, one such estimate consumes two function evaluations per sampled direction. If $m$ directions are averaged, the cost is $2m$ zeroth-order queries. This convention is important when comparing black-box variants to first-order baselines.

\subsection{Training Specification and Reproducibility of SHAPE optimizer}
\label{app:exp:shape_training_reproducibility}
All experiments regarding SHAPE optimizer in the functional benchmark suite were implemented in
PyTorch\cite{paszke2019pytorch} and trained on a single NVIDIA A100 GPU. Each training run uses a
three-stage update schedule: first, \texttt{local\_pretrain\_steps} calls to
\texttt{local\_pretrain\_step} are executed before the epoch loop; second, each
epoch performs \texttt{n\_controller\_updates} additional local-controller
updates; third, each epoch performs \texttt{n\_planner\_updates}
rollout/planner updates through \texttt{planner\_update\_step}.  The
configuration used for each dimensional setting is summarized in
\cref{tab:app:shape_training_specs}. 

\begin{table}[t]
\centering
\caption{Training and evaluation specification for SHAPE on functional benchmarks.}
\label{tab:app:shape_training_specs}
\small
\setlength{\tabcolsep}{3pt}
\resizebox{\textwidth}{!}{%
\begin{tabular}{rlrrrrcll}
\toprule
Dim.  & Epochs & Stage-0 pretrain & Ctrl. upd./epoch & Planner upd./epoch
& Batch/eval batch & Hidden dim. & Train rollout & Eval rollout \\
\midrule
2
& 500 & 100 & 2 & 2 & 64/64 & 32
& \texttt{n\_steps=128}
& \texttt{n\_steps=500,n\_eval=128} \\
20
& 800 & 100 & 2 & 2 & 64/64 & 64
& \texttt{n\_steps=128}
& \texttt{n\_steps=500,n\_eval=64} \\
100
& 1000 & 100 & 2 & 2 & 64/64 & 128
& \texttt{n\_steps=128}
& \texttt{n\_steps=500,n\_eval=32} \\
500
& 1000 & 100 & 2 & 2 & 64/64 & 128
& \texttt{n\_steps=128}
& \texttt{n\_steps=500,n\_eval=32}
\\
\bottomrule
\end{tabular}%
}
\end{table}

The no-local-controller ablation removes the first two stages by setting
\texttt{local\_pretrain\_steps=0} and \texttt{n\_controller\_updates=0}.  The
no-memory ablation sets \texttt{use\_memory=false}, so the controller receives
\texttt{mem\_dim=0}.  Otherwise, memory is enabled with a two-dimensional
spatial memory for \(d=2\) and a blockwise multi-scale memory for \(d>2\).
The ablation configurations use \texttt{event\_horizon=4}; when this value is
not overridden, the code default is \texttt{event\_horizon=6}.

\begin{table}[t]
\centering
\caption{
Loss weights used for SHAPE training.  The method section reports the same
objective in Eq.~\eqref{eq:main_training_objective}.  The configuration key is
listed to make the implementation mapping explicit.
}
\label{tab:shape_loss_weights}
\small
\begin{tabular}{llll}
\toprule
Paper symbol & Value & Role \\
\midrule
\(\lambda_{\rm term}\)  
& \(1.0\) 
& terminal normalized distance \\
\(\lambda_{\rm best}\)  
& \(0.5\) 
& best-seen normalized distance \\
\(\lambda_{\rm prog}\) 
& \(0.10\) 
& stage waypoint/progress distance \\
\(\lambda_{\rm ce}\) 
& \(0.40\) 
& planner mode cross-entropy \\
\(\lambda_{\rm sg}\) 
& \(0.25\) 
& planner penalty \\
\(\lambda_{\rm ctrl}\) 
& \(10^{-3}\) 
& control magnitude penalty \\
\(\lambda_{JR}\) 
& \(5\times 10^{-4}\) 
& learned \(J\), \(R\), and port-size regularization \\
\(\lambda_{\rm port}\) 
& \(5\times 10^{-4}\) 
& positive port-power penalty \\
\bottomrule
\end{tabular}
\end{table}

For all experiments pH time integration uses $dt=0.05$ as the default step size and is implemented via a semi-implicit Euler scheme. It is possible to implement implicit methods such as St\"{o}rmer Verlet but the cost of Newton solver is intractable when dimension becomes higher.

\subsection{Implementation and Reproducibility of Baseline Methods}
\label{app:exp:baseline_reproducibility}

Baseline comparisons use deterministic hyperparameters from \cref{tab:app:baseline_hparams_momentum,tab:app:baseline_hparams_adaptive,tab:app:baseline_hparams_statistical}
are used.  All baseline optimizers are evaluated on the same task instances, starting points, oracle streams, projection or clipping rules, and total oracle budget as
SHAPE.  For exact first-order experiments, one baseline iteration consumes one gradient-oracle call.  For stochastic or mini-batch first-order experiments, the
same random seed is used to generate paired oracle noise whenever a paired comparison is reported.  For zeroth-order variants, query cost follows the accounting in \cref{eq:app:zo_oracle}: a two-sided random-direction estimate with one direction consumes two function evaluations, and averaging over \(m\) directions consumes \(2m\) function evaluations.

To keep the tables compact, all methods that use a numerical stabilizer take \(\epsilon=10^{-8}\).  The tuple orders are stated in the column headers.  For LionK, \(k\) is the kick period and \(\gamma_{\rm kick}\) is the kick scale. The row labeled \texttt{fallback} is used when a task type has no specialized preassigned value of hyperparameter (for phase retrieval and control objective).

\begin{table}[ht]
\centering
\caption{Default hyperparameters for step-size and momentum-type baselines.}
\label{tab:app:baseline_hparams_momentum}
\small
\setlength{\tabcolsep}{5pt}
\begin{tabular}{lccc}
\toprule
Task type & GD \((\mathrm{lr})\) & Momentum \((\mathrm{lr},\beta)\) & NAG \((\mathrm{lr},\beta)\) \\
\midrule
\texttt{ackley}             & 0.030  & (0.028, 0.72)  & (0.026, 0.82) \\
\texttt{rastrigin}          & 0.012  & (0.011, 0.65)  & (0.010, 0.78) \\
\texttt{levy}               & 0.018  & (0.016, 0.68)  & (0.015, 0.80) \\
\texttt{multi\_well\_barrier} & 0.014  & (0.012, 0.72)  & (0.011, 0.80) \\
\texttt{lj\_cluster}        & 0.0025 & (0.0020, 0.65) & (0.0018, 0.75) \\
\texttt{fallback}           & 0.020  & (0.018, 0.75)  & (0.016, 0.82) \\
\bottomrule
\end{tabular}
\end{table}

\begin{table}[ht]
\centering
\caption{Default hyperparameters for adaptive first-order baselines.  All methods use \(\epsilon=10^{-8}\).}
\label{tab:app:baseline_hparams_adaptive}
\small
\setlength{\tabcolsep}{4pt}
\resizebox{\textwidth}{!}{%
\begin{tabular}{lccc}
\toprule
Task type
& RMSProp \((\mathrm{lr},\alpha)\)
& Adam \((\mathrm{lr},\beta_1,\beta_2)\)
& LionK \((\mathrm{lr},\beta_1,\beta_2,k,\gamma_{\rm kick})\) \\
\midrule
\texttt{ackley}             & (0.020, 0.99)   & (0.025, 0.9, 0.999)  & (0.010, 0.9, 0.99, 5, 1.15) \\
\texttt{rastrigin}          & (0.010, 0.99)   & (0.012, 0.9, 0.999)  & (0.006, 0.9, 0.99, 5, 1.10) \\
\texttt{levy}               & (0.013, 0.99)   & (0.016, 0.9, 0.999)  & (0.008, 0.9, 0.99, 5, 1.10) \\
\texttt{multi\_well\_barrier} & (0.010, 0.99)   & (0.012, 0.9, 0.999)  & (0.006, 0.9, 0.99, 5, 1.10) \\
\texttt{lj\_cluster}        & (0.0015, 0.99)  & (0.0020, 0.9, 0.999) & (0.0015, 0.9, 0.99, 4, 1.05) \\
\texttt{fallback}           & (0.012, 0.99)   & (0.015, 0.9, 0.999)  & (0.008, 0.9, 0.99, 5, 1.10) \\
\bottomrule
\end{tabular}%
}
\end{table}

\begin{table}[ht]
\centering
\caption{Default hyperparameters for matrix/statistic adaptive baselines.  All methods use \(\epsilon=10^{-8}\).}
\label{tab:app:baseline_hparams_statistical}
\small
\setlength{\tabcolsep}{4pt}
\resizebox{\textwidth}{!}{%
\begin{tabular}{lccc}
\toprule
Task type
& Shampoo \((\mathrm{lr})\)
& SOAP \((\mathrm{lr},\beta,\nu)\)
& Sophia \((\mathrm{lr},\beta_1,\beta_2,\rho)\) \\
\midrule
\texttt{ackley}             & 0.020  & (0.012, 0.95, 0.99)  & (0.030, 0.965, 0.99, 0.030) \\
\texttt{rastrigin}          & 0.010  & (0.008, 0.95, 0.99)  & (0.012, 0.965, 0.99, 0.025) \\
\texttt{levy}               & 0.013  & (0.010, 0.95, 0.99)  & (0.016, 0.965, 0.99, 0.030) \\
\texttt{multi\_well\_barrier} & 0.010  & (0.008, 0.95, 0.99)  & (0.012, 0.965, 0.99, 0.025) \\
\texttt{lj\_cluster}        & 0.0015 & (0.0018, 0.95, 0.99) & (0.0020, 0.965, 0.99, 0.020) \\
\texttt{fallback}           & 0.012  & (0.010, 0.95, 0.99)  & (0.016, 0.965, 0.99, 0.030) \\
\bottomrule
\end{tabular}%
}
\end{table}
\subsection{Additional Result For Different Tasks}
\label{app:exp:task_full_details}

The synthetic examples serve two purposes. First, they expose the distinct roles of the slow global planner and the fast local pH controller in a setting where the landscape geometry is easy to visualize. Second, they provide a controlled bridge from low-dimensional illustration to the higher-dimensional benchmark tables reported later in the section.

\label{subsec:first_order_online_gd}
\paragraph{Multi-Well Potentials.}
We begin with a one-dimensional family whose geometry can be prescribed explicitly. This experiment is designed to isolate the exploration mechanism: the local controller must remain stable inside each well, while the slow event-triggered planner must decide when to leave an already visited basin and scan for better regions. It is therefore the clearest diagnostic for the ``refine / explore / escape'' behavior discussed in the method section.

\emph{Task family.}
Fix a domain $\mathcal{X}=[-K,K]\subset\mathbb{R}$ and an integer $L\ge 2$. Each task $\tau$ specifies
$2L-1$ strictly ordered interior extrema locations
\[
  -K < x_1 < x_2 < \cdots < x_{2L-1} < K,
\]
where $\{x_{2\ell-1}\}_{\ell=1}^{L}$ are designated as local minima and $\{x_{2\ell}\}_{\ell=1}^{L-1}$ as local maxima.
We sample $\{x_i\}$ at random subject to a minimum separation constraint (to avoid degenerate wells), and then sample
function values $\{v_i\}_{i=1}^{2L-1}$ alternating between \emph{valleys} and \emph{barriers}.
To enforce a \emph{unique global minimum}, we select one valley index $i^\star\in\{1,3,\dots,2L-1\}$ and assign
$v_{i^\star}<\min_{\ell\neq i^\star,\ \ell\ \mathrm{odd}} v_{\ell}$ by a fixed margin.

We construct a $C^1$ potential $f_\tau:\mathcal{X}\to\mathbb{R}$ by a \emph{piecewise cubic Hermite spline}
interpolating the knots $(x_i,v_i)$ with \emph{zero slope at every knot}:
\[
  f_\tau(x_i)=v_i,\qquad f_\tau'(x_i)=0,\qquad i=1,\dots,2L-1.
\]
On each interval $[x_i,x_{i+1}]$, let $t=(x-x_i)/h_i$ with $h_i=x_{i+1}-x_i$ and define
\begin{equation}
  f_\tau(x)
  \;=\;
  v_i\,h_{00}(t) + v_{i+1}\,h_{01}(t),
  \qquad
  h_{00}(t)=2t^3-3t^2+1,\;\;
  h_{01}(t)=-2t^3+3t^2,
  \label{eq:hermite_zero_slope}
\end{equation}
which is the Hermite interpolant with endpoint derivatives set to zero. The derivative is available in closed form:
\begin{equation}
  f_\tau'(x)
  \;=\;
  \frac{1}{h_i}\Big(v_i\,h_{00}'(t)+v_{i+1}\,h_{01}'(t)\Big),
  \qquad
  h_{00}'(t)=6t^2-6t,\;\;
  h_{01}'(t)=-6t^2+6t.
\end{equation}
Because the endpoint slopes are zero and $h_{00}'(t)$ and $h_{01}'(t)$ have no additional roots in $(0,1)$ beyond
those induced by the endpoints, the interior stationary points of $f_\tau$ are exactly $\{x_i\}$, hence $f_\tau$
has precisely $L$ local minima and $L-1$ local maxima on $\mathcal{X}$ by construction.
To prevent boundary effects (e.g.\ spurious boundary minima), we impose high boundary walls by fixing boundary values
$f_\tau(\pm K)$ to exceed all interior barriers (or equivalently appending quadratic walls outside $\mathcal{X}$).

\emph{Noisy oracle.}
We provide noisy first-order information via
\[
  \tilde g(x_k;\xi_k)=f_\tau'(x_k)+\sigma_g\,\xi_k,\qquad \xi_k\sim\mathcal{N}(0,1).
\]

\emph{Train$\rightarrow$test protocol and relation to the curriculum.}
This experiment uses the simplest curriculum instantiation. We first train the controller on the easier double-well distribution
$\tau\sim\mathcal{D}^{\mathrm{DW}}_{\mathrm{train}}$, where Phase~I emphasizes local reachability and Phase~II introduces event-triggered escape decisions. We then freeze the learned parameters and evaluate \emph{zero-shot} on the harder multi-well spline family
$\tau\sim\mathcal{D}^{\mathrm{MW}}_{\mathrm{test}}$ with unseen extrema patterns and barrier heights. In other words, the multi-well study is not only a test of generalization across tasks, but also a direct test of whether the curriculum transfers from a simpler bimodal training family to a richer multi-basin deployment family.

\emph{Metrics.}
The table below reports exactly the quantities used in the comparison: \emph{success rate}, \emph{final gap}, \emph{best gap}, \emph{median best gap}, and the realized number of \emph{gradient calls}. Here the final gap measures the objective gap at the terminal rollout state, while the best gap records the best-so-far value over the full budget. Success rate is the fraction of runs whose best-so-far value reaches the prescribed near-global tolerance. The median best gap is included because this family has a heavy-tailed distribution of outcomes: some runs remain trapped in early wells while successful runs may cross several barriers. Reporting both mean and median therefore separates typical performance from rare but very difficult failures. Gradient calls are listed explicitly because all methods are evaluated under the same paired-budget protocol, and early termination of dissipative Hamiltonian trajectories can otherwise make raw rollout length misleading.

\begin{figure}
    \centering
    \includegraphics[width=0.8\linewidth]{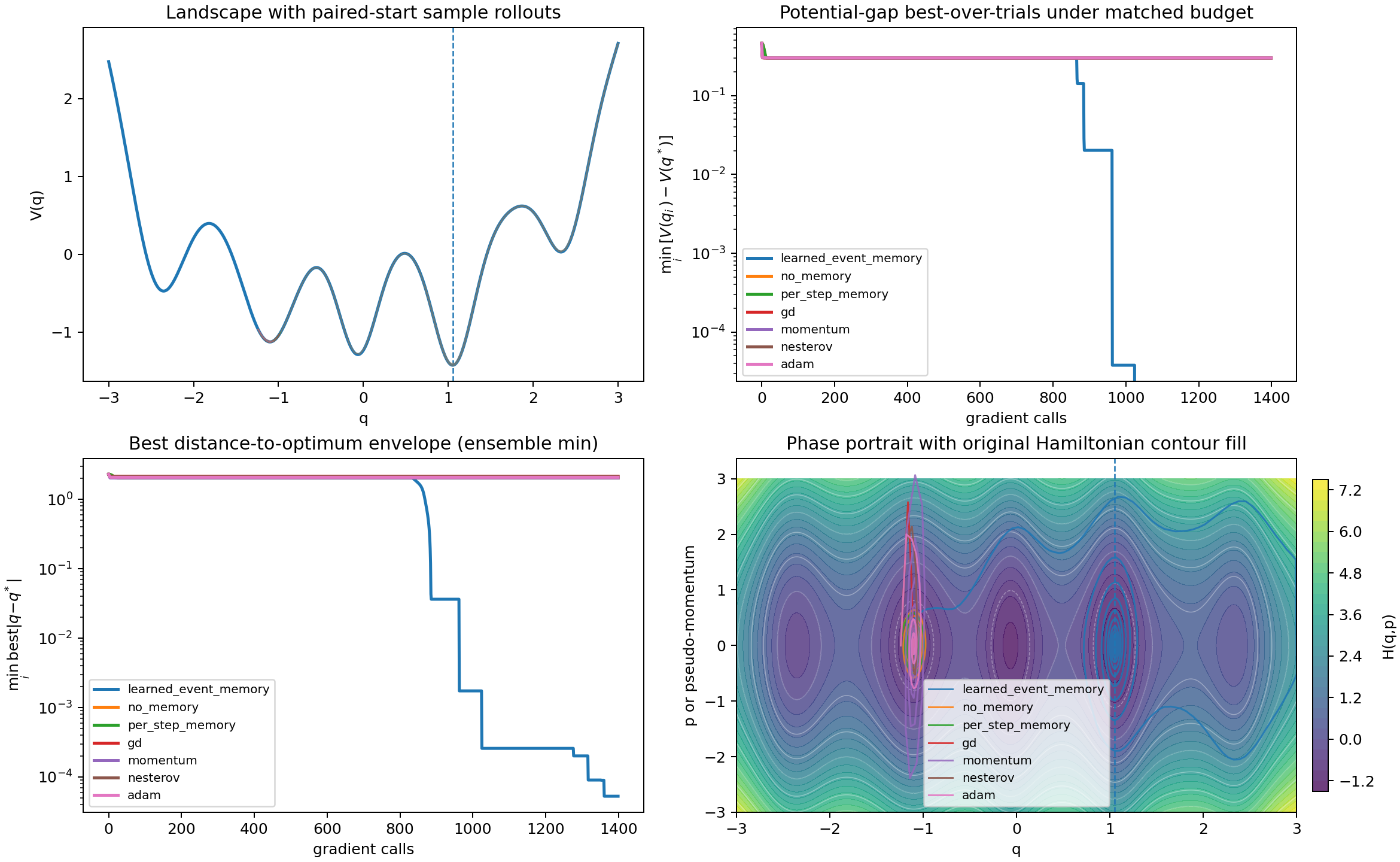}
    \caption{Multiwell potential visualization. The plot shows the best-so-far function value and how the dynamics learn to escape and probe the landscape compared with other methods. With a fixed initial point and configuration, baseline methods can converge to local minima, whereas effective escape usually requires a multistart policy or carefully tuned hyperparameters. Our method automates exploration through a traversal policy on the augmented phase space.}
    \label{fig:multiwell:viz:panel}
\end{figure}

\begin{table}[t]
\centering
\caption{Multi-well comparison over 500 tasks. All methods are evaluated under the same paired oracle-budget protocol: the initial starting point is randomly sampled within the search domain and each task only start at one initial point. Therefore, every method without prior guidance has an identical chance of falling into a local minimum, whereas our learned method can attempt to explore under the given budget constraints (cannot consume more online gradients compared with other methods). Our methods without pretrained memory shows fewer actual gradient calls on average because it terminates earlier than other methods as a Hamiltonian dynamics with dissipation. All numbers are reported as mean performance over all 500 random multi-well tasks.}
\small
\label{tab:multiwell_fair_budget}
\setlength{\tabcolsep}{4pt}
\begin{tabular}{lcccccc}
\toprule
Method & Success rate $\uparrow$ & Final gap $\downarrow$ & Best gap $\downarrow$ & Best gap median $\downarrow$ & Gradient Calls \\
\midrule
SHAPE & \textbf{0.602} & \textbf{0.987} & \textbf{0.477} & \textbf{0.007}  & 1251.2 \\
SHAPE(w.o. memory) & 0.300 & 1.249 & 1.066 & 0.895  & 707.3 \\
GD                   & 0.300 & 1.290 & 1.180 & 1.115  & 1251.2 \\
Momentum             & 0.300 & 1.290 & 1.115 & 0.950  & 1251.2 \\
Nesterov             & 0.300 & 1.290 & 1.162 & 1.066 & 1251.2 \\
Adam                 & 0.300 & 1.290 & 1.147 & 1.000  & 1251.2 \\
\bottomrule
\end{tabular}
\end{table}

\paragraph{Standard nonconvex benchmarks (Ackley, L\'evy, Rastrigin)}
We next consider the standard analytic benchmarks Ackley, L\'evy, and Rastrigin, which interpolate between smooth funnel-like geometry and highly multimodal oscillatory landscapes. These tasks provide the main dimension-scaling study reported in \cref{tab:summary:synthetic:benchmarks}. Each task is
\(
\tau=(n,\mathrm{fun},s,R)
\)
where \(n\) is the ambient dimension, \(\mathrm{fun}\in\{\mathrm{Ackley},\mathrm{Rastrigin},\mathrm{Levy}\}\), \(s\in\mathcal{X}\) is a random shift, and \(R\in\mathbb{R}^{n\times n}\) is an orthogonal rotation (used when we wish to destroy coordinate separability). The search domain is \(\mathcal{X}=[-5,5]^n\), and we define
\begin{equation}
  f_\tau(q) := f_{\mathrm{fun}}\big(R(q-s)\big), \qquad q\in\mathcal{X},
  \label{eq:ackley_levy_rastrigin_task}
\end{equation}
with the convention that for the L\'evy function we use the shifted form \(f_{\mathrm{Levy}}(x):=f_{\mathrm{Levy,orig}}(x+\mathbf{1})\), so that the global minimum is zero at \(x=0\) (hence the minimizer of \(f_\tau\) is \(q^\star=s\) when \(R=I\)). Explicit formulas for the three benchmark families are given below:
\begin{align}
  f_{\mathrm{Ackley}}(x)
  &:= -a\exp\!\Big(-b\sqrt{\tfrac{1}{n}\sum_{i=1}^n x_i^2}\Big)
      -\exp\!\Big(\tfrac{1}{n}\sum_{i=1}^n \cos(c x_i)\Big) + a + e,
  \label{eq:ackley_def}\\
  f_{\mathrm{Rastrigin}}(x)
  &:= 10n + \sum_{i=1}^n\big(x_i^2 - 10\cos(2\pi x_i)\big),
  \label{eq:rastrigin_def}\\
  f_{\mathrm{Levy}}(x)
  &:= \sin^2(\pi w_1)
    + \sum_{i=1}^{n-1}(w_i-1)^2\big(1+10\sin^2(\pi w_i+1)\big)
    \\&+ (w_n-1)^2\big(1+\sin^2(2\pi w_n)\big),
  ~ w_i := 1+\tfrac{x_i}{4}. \nonumber
  \label{eq:levy_def}
\end{align}
We use the standard parameters \(a=20\), \(b=0.2\), and \(c=2\pi\) in~\eqref{eq:ackley_def}. All three functions admit analytic gradients; when we study noisy online feedback we perturb the oracle as
\(
\tilde g(q_k;\xi_k) = \nabla f_\tau(q_k) + \sigma_g\,\xi_k,\;\xi_k\sim\mathcal{N}(0,I).
\)

\begin{table}[!htbp]
\centering
\small
\setlength{\tabcolsep}{4pt}
\caption{Benchmark summary for the new Ackley, Levy, and Rastrigin experiments in mean$\pm$std format. Each block header indicates the ambient dimension, with $d\in\{2,20,100,500\}$. Entries in bold indicate the best mean in each metric column within each task-and-dimension block.}
\label{tab:summary:synthetic:benchmarks}
\resizebox{\textwidth}{!}{%
\begin{tabular}{llccccccc}
\toprule
Task & Method & Final dist. $\downarrow$ & Final gap $\downarrow$ & Best gap $\downarrow$ & Hit rate $\uparrow$ & AUC dist. $\downarrow$ & AUC gap $\downarrow$ & AUC best gap $\downarrow$ \\
\midrule
\multicolumn{9}{l}{\textit{Ackley ($d=2$)}} \\
 & Adam & $2.398 \pm 0.217$ & $5.973 \pm 0.982$ & $5.962 \pm 0.982$ & $0.000 \pm 0.000$ & $2.443 \pm 0.200$ & $6.516 \pm 0.951$ & $6.474 \pm 0.950$ \\
 & GD & $2.337 \pm 0.279$ & $6.493 \pm 1.035$ & $5.920 \pm 1.039$ & $0.012 \pm 0.033$ & $2.392 \pm 0.219$ & $6.701 \pm 1.005$ & $6.237 \pm 0.950$ \\
 & Learned & {\boldmath $0.273 \pm 0.041$} & {\boldmath $2.179 \pm 0.349$} & {\boldmath $1.004 \pm 0.401$} & {\boldmath $0.318 \pm 0.115$} & {\boldmath $1.324 \pm 0.089$} & {\boldmath $5.143 \pm 0.421$} & $4.795 \pm 0.382$ \\
 & Momentum & $1.868 \pm 0.354$ & $5.026 \pm 0.844$ & $4.429 \pm 1.145$ & $0.105 \pm 0.062$ & $2.002 \pm 0.277$ & $5.615 \pm 0.783$ & $5.168 \pm 0.859$ \\
 & NAG & $1.539 \pm 0.579$ & $4.660 \pm 1.006$ & $3.075 \pm 1.614$ & $0.084 \pm 0.064$ & $1.787 \pm 0.347$ & $5.360 \pm 0.672$ & {\boldmath $4.312 \pm 1.009$} \\
 & RMSProp & $2.395 \pm 0.214$ & $5.960 \pm 0.978$ & $5.955 \pm 0.977$ & $0.000 \pm 0.000$ & $2.404 \pm 0.212$ & $6.077 \pm 0.976$ & $6.067 \pm 0.974$ \\
\addlinespace[2pt]
\multicolumn{9}{l}{\textit{Ackley ($d=20$)}} \\
 & Adam & $2.043 \pm 0.244$ & $1.691 \pm 0.185$ & $1.689 \pm 0.185$ & $0.051 \pm 0.059$ & $2.145 \pm 0.185$ & $2.203 \pm 0.231$ & $2.201 \pm 0.231$ \\
 & GD & $2.254 \pm 0.147$ & $2.004 \pm 0.262$ & $2.004 \pm 0.262$ & $0.006 \pm 0.012$ & $2.307 \pm 0.111$ & $2.765 \pm 0.287$ & $2.765 \pm 0.287$ \\
 & Learned & {\boldmath $0.228 \pm 0.060$} & {\boldmath $0.387 \pm 0.171$} & {\boldmath $0.175 \pm 0.038$} & {\boldmath $0.436 \pm 0.369$} & {\boldmath $1.178 \pm 0.058$} & $2.065 \pm 0.157$ & $2.047 \pm 0.151$ \\
 & Momentum & $2.221 \pm 0.184$ & $1.827 \pm 0.224$ & $1.826 \pm 0.224$ & $0.033 \pm 0.037$ & $2.258 \pm 0.147$ & $2.311 \pm 0.247$ & $2.310 \pm 0.247$ \\
 & NAG & $2.205 \pm 0.205$ & $1.806 \pm 0.213$ & $1.805 \pm 0.213$ & $0.045 \pm 0.061$ & $2.245 \pm 0.163$ & $2.247 \pm 0.234$ & $2.246 \pm 0.233$ \\
 & RMSProp & $2.015 \pm 0.242$ & $1.653 \pm 0.179$ & $1.652 \pm 0.179$ & $0.059 \pm 0.065$ & $2.039 \pm 0.228$ & {\boldmath $1.766 \pm 0.182$} & {\boldmath $1.765 \pm 0.182$} \\
\addlinespace[2pt]
\multicolumn{9}{l}{\textit{Ackley ($d=100$)}} \\
 & Adam & $0.384 \pm 0.353$ & $0.150 \pm 0.120$ & $0.148 \pm 0.120$ & $0.794 \pm 0.204$ & $0.985 \pm 0.258$ & $0.731 \pm 0.093$ & $0.731 \pm 0.093$ \\
 & GD & $1.925 \pm 0.253$ & $1.595 \pm 0.192$ & $1.595 \pm 0.192$ & $0.003 \pm 0.010$ & $2.493 \pm 0.178$ & $2.341 \pm 0.184$ & $2.341 \pm 0.184$ \\
 & Learned & {\boldmath $0.178 \pm 0.039$} & {\boldmath $0.102 \pm 0.038$} & {\boldmath $0.078 \pm 0.019$} & $0.701 \pm 0.366$ & $1.277 \pm 0.061$ & $1.291 \pm 0.106$ & $1.261 \pm 0.103$ \\
 & Momentum & $0.943 \pm 0.367$ & $0.472 \pm 0.151$ & $0.472 \pm 0.152$ & $0.406 \pm 0.171$ & $1.766 \pm 0.262$ & $1.429 \pm 0.146$ & $1.429 \pm 0.146$ \\
 & NAG & $0.809 \pm 0.381$ & $0.367 \pm 0.146$ & $0.366 \pm 0.146$ & $0.505 \pm 0.202$ & $1.609 \pm 0.279$ & $1.257 \pm 0.137$ & $1.257 \pm 0.137$ \\
 & RMSProp & $0.376 \pm 0.298$ & $0.156 \pm 0.096$ & $0.154 \pm 0.095$ & {\boldmath $0.852 \pm 0.179$} & {\boldmath $0.537 \pm 0.275$} & {\boldmath $0.310 \pm 0.086$} & {\boldmath $0.309 \pm 0.086$} \\
\addlinespace[2pt]
\multicolumn{9}{l}{\textit{Ackley ($d=500$)}} \\
 & Adam & {\boldmath $0.035 \pm 0.002$} & {\boldmath $0.004\!\pm\!6.40e\!-\!4$} & {\boldmath $0.004\!\pm\!6.28e\!-\!4$} & {\boldmath $1.000 \pm 0.000$} & {\boldmath $0.459 \pm 0.026$} & $0.157 \pm 0.020$ & $0.157 \pm 0.020$ \\
 & GD & $1.951 \pm 0.199$ & $0.829 \pm 0.088$ & $0.829 \pm 0.088$ & $0.000 \pm 0.000$ & $2.510 \pm 0.167$ & $1.181 \pm 0.123$ & $1.181 \pm 0.123$ \\
 & Learned & $0.224 \pm 0.068$ & $0.049 \pm 0.026$ & $0.033 \pm 0.008$ & $0.490 \pm 0.423$ & $1.358 \pm 0.067$ & $0.576 \pm 0.064$ & $0.558 \pm 0.060$ \\
 & Momentum & $0.293 \pm 0.158$ & $0.068 \pm 0.037$ & $0.068 \pm 0.037$ & $0.557 \pm 0.231$ & $1.530 \pm 0.188$ & $0.644 \pm 0.062$ & $0.644 \pm 0.062$ \\
 & NAG & $0.046 \pm 0.036$ & $0.007 \pm 0.006$ & $0.007 \pm 0.006$ & $0.922 \pm 0.081$ & $1.223 \pm 0.155$ & $0.515 \pm 0.049$ & $0.515 \pm 0.049$ \\
 & RMSProp & $0.281 \pm 0.008$ & $0.062 \pm 0.010$ & $0.062 \pm 0.010$ & $0.000 \pm 0.000$ & $0.492 \pm 0.011$ & {\boldmath $0.152 \pm 0.021$} & {\boldmath $0.140 \pm 0.019$} \\
\addlinespace[2pt]
\midrule
\multicolumn{9}{l}{\textit{Levy ($d=2$)}} \\
 & Adam & $1.990 \pm 0.231$ & $0.974 \pm 0.200$ & $0.973 \pm 0.200$ & $0.000 \pm 0.000$ & $2.265 \pm 0.213$ & $1.568 \pm 0.383$ & $1.568 \pm 0.383$ \\
 & GD & $2.023 \pm 0.188$ & $0.739 \pm 0.090$ & $0.739 \pm 0.090$ & $0.000 \pm 0.000$ & $2.226 \pm 0.174$ & $1.143 \pm 0.124$ & $1.143 \pm 0.124$ \\
 & Learned & {\boldmath $0.357 \pm 0.106$} & {\boldmath $0.139 \pm 0.072$} & {\boldmath $0.044 \pm 0.023$} & {\boldmath $0.227 \pm 0.189$} & {\boldmath $1.507 \pm 0.084$} & $1.061 \pm 0.225$ & $1.014 \pm 0.213$ \\
 & Momentum & $1.724 \pm 0.231$ & $0.513 \pm 0.136$ & $0.513 \pm 0.136$ & $0.039 \pm 0.045$ & $1.989 \pm 0.191$ & $0.836 \pm 0.122$ & $0.835 \pm 0.120$ \\
 & NAG & $1.608 \pm 0.240$ & $0.489 \pm 0.140$ & $0.489 \pm 0.140$ & $0.066 \pm 0.069$ & $1.905 \pm 0.201$ & {\boldmath $0.781 \pm 0.133$} & {\boldmath $0.780 \pm 0.132$} \\
 & RMSProp & $1.577 \pm 0.293$ & $0.614 \pm 0.179$ & $0.614 \pm 0.179$ & $0.049 \pm 0.049$ & $1.885 \pm 0.251$ & $0.932 \pm 0.222$ & $0.932 \pm 0.222$ \\
\addlinespace[2pt]
\multicolumn{9}{l}{\textit{Levy ($d=20$)}} \\
 & Adam & $1.210 \pm 0.327$ & $0.298 \pm 0.120$ & $0.298 \pm 0.120$ & $0.033 \pm 0.026$ & $1.723 \pm 0.185$ & $1.114 \pm 0.344$ & $1.114 \pm 0.344$ \\
 & GD & $1.769 \pm 0.098$ & $0.605 \pm 0.067$ & $0.605 \pm 0.067$ & $0.000 \pm 0.000$ & $2.077 \pm 0.090$ & $1.303 \pm 0.215$ & $1.303 \pm 0.215$ \\
 & Learned & {\boldmath $0.171 \pm 0.055$} & {\boldmath $0.020 \pm 0.018$} & {\boldmath $0.010 \pm 0.007$} & {\boldmath $0.680 \pm 0.345$} & {\boldmath $1.085 \pm 0.061$} & $0.779 \pm 0.226$ & $0.765 \pm 0.220$ \\
 & Momentum & $1.429 \pm 0.150$ & $0.279 \pm 0.077$ & $0.279 \pm 0.077$ & $0.004 \pm 0.010$ & $1.780 \pm 0.094$ & $0.797 \pm 0.141$ & $0.797 \pm 0.141$ \\
 & NAG & $1.321 \pm 0.192$ & $0.230 \pm 0.087$ & $0.230 \pm 0.087$ & $0.018 \pm 0.031$ & $1.689 \pm 0.107$ & $0.707 \pm 0.144$ & $0.706 \pm 0.144$ \\
 & RMSProp & $0.995 \pm 0.429$ & $0.154 \pm 0.114$ & $0.154 \pm 0.114$ & $0.340 \pm 0.205$ & $1.227 \pm 0.336$ & {\boldmath $0.396 \pm 0.156$} & {\boldmath $0.396 \pm 0.156$} \\
\addlinespace[2pt]
\multicolumn{9}{l}{\textit{Levy ($d=100$)}} \\
 & Adam & $0.463 \pm 0.227$ & $0.062 \pm 0.045$ & $0.062 \pm 0.045$ & $0.534 \pm 0.133$ & $1.210 \pm 0.137$ & $0.887 \pm 0.315$ & $0.887 \pm 0.315$ \\
 & GD & $1.694 \pm 0.181$ & $0.798 \pm 0.056$ & $0.798 \pm 0.056$ & $0.000 \pm 0.000$ & $2.246 \pm 0.158$ & $1.835 \pm 0.291$ & $1.835 \pm 0.291$ \\
 & Learned & $0.402 \pm 0.042$ & $0.091 \pm 0.052$ & $0.080 \pm 0.039$ & $0.000 \pm 0.000$ & $1.327 \pm 0.056$ & $1.150 \pm 0.366$ & $1.121 \pm 0.352$ \\
 & Momentum & $0.934 \pm 0.102$ & $0.199 \pm 0.036$ & $0.199 \pm 0.036$ & $0.029 \pm 0.050$ & $1.654 \pm 0.134$ & $1.015 \pm 0.139$ & $1.015 \pm 0.139$ \\
 & NAG & $0.705 \pm 0.126$ & $0.111 \pm 0.041$ & $0.111 \pm 0.041$ & $0.167 \pm 0.118$ & $1.450 \pm 0.105$ & $0.848 \pm 0.140$ & $0.848 \pm 0.140$ \\
 & RMSProp & {\boldmath $0.339 \pm 0.265$} & {\boldmath $0.041 \pm 0.040$} & {\boldmath $0.041 \pm 0.040$} & {\boldmath $0.742 \pm 0.208$} & {\boldmath $0.571 \pm 0.217$} & {\boldmath $0.232 \pm 0.098$} & {\boldmath $0.232 \pm 0.098$} \\
\addlinespace[2pt]
\multicolumn{9}{l}{\textit{Levy ($d=500$)}} \\
 & Adam & $0.069 \pm 0.002$ & $0.003 \pm 7.97e\!-\!4$ & $0.003 \pm 7.97e\!-\!4$ & {\boldmath $1.000 \pm 0.000$} & $0.608 \pm 0.024$ & $0.456 \pm 0.133$ & $0.456 \pm 0.133$ \\
 & GD & $1.062 \pm 0.354$ & $0.481 \pm 0.161$ & $0.481 \pm 0.161$ & $0.000 \pm 0.000$ & $1.852 \pm 0.285$ & $1.778 \pm 0.169$ & $1.778 \pm 0.169$ \\
 & Learned & $0.283 \pm 0.045$ & $0.046 \pm 0.023$ & $0.037 \pm 0.016$ & $0.000 \pm 0.000$ & $1.421 \pm 0.046$ & $1.416 \pm 0.400$ & $1.358 \pm 0.377$ \\
 & Momentum & $0.217 \pm 0.175$ & $0.021 \pm 0.026$ & $0.021 \pm 0.026$ & $0.609 \pm 0.441$ & $1.038 \pm 0.277$ & $0.829 \pm 0.051$ & $0.829 \pm 0.051$ \\
 & NAG & $0.088 \pm 0.072$ & $0.002 \pm 0.004$ & $0.002 \pm 0.004$ & $0.875 \pm 0.264$ & $0.808 \pm 0.218$ & $0.680 \pm 0.060$ & $0.680 \pm 0.060$ \\
 & RMSProp & {\boldmath $1.72e\!-\!5\!\pm\!2.05e\!-\!5$} & {\boldmath $2.05e\!-\!9\!\pm\!5.14e\!-9\!$} & {\boldmath $2.05e\!-\!9\!\pm\!5.14e\!-\!9$} & {\boldmath $1.000 \pm 0.000$} & {\boldmath $0.118 \pm 0.006$} & {\boldmath $0.118 \pm 0.035$} & {\boldmath $0.118 \pm 0.035$} \\
\addlinespace[2pt]
\midrule
\multicolumn{9}{l}{\textit{Rastrigin ($d=2$)}} \\
 & Adam & $2.592 \pm 0.157$ & $7.648 \pm 1.122$ & $7.482 \pm 1.116$ & $0.000 \pm 0.000$ & $2.598 \pm 0.157$ & $14.050 \pm 1.836$ & $13.949 \pm 1.810$ \\
 & GD & $4.406 \pm 1.951$ & $53.778 \pm 20.002$ & $11.615 \pm 3.247$ & $0.000 \pm 0.000$ & $3.745 \pm 1.364$ & $46.169 \pm 14.126$ & $15.164 \pm 2.962$ \\
 & Learned & {\boldmath $1.157 \pm 0.120$} & $22.553 \pm 3.983$ & {\boldmath $4.973 \pm 0.823$} & {\boldmath $0.027 \pm 0.062$} & {\boldmath $2.039 \pm 0.106$} & $21.720 \pm 3.603$ & $9.763 \pm 1.263$ \\
 & Momentum & $5.569 \pm 1.816$ & $64.593 \pm 28.991$ & $12.203 \pm 3.817$ & $0.002 \pm 0.008$ & $4.679 \pm 1.274$ & $53.658 \pm 21.374$ & $15.397 \pm 4.231$ \\
 & NAG & $5.875 \pm 0.857$ & $72.123 \pm 17.224$ & $13.970 \pm 3.874$ & $0.000 \pm 0.000$ & $5.196 \pm 0.821$ & $63.207 \pm 14.933$ & $17.460 \pm 3.944$ \\
 & RMSProp & $2.593 \pm 0.159$ & {\boldmath $7.404 \pm 1.119$} & $7.404 \pm 1.119$ & $0.000 \pm 0.000$ & $2.594 \pm 0.158$ & {\boldmath $8.545 \pm 1.150$} & {\boldmath $8.544 \pm 1.149$} \\
\addlinespace[2pt]
\multicolumn{9}{l}{\textit{Rastrigin ($d=20$)}} \\
 & Adam & $2.821 \pm 0.150$ & $10.689 \pm 0.962$ & $10.614 \pm 0.978$ & {\boldmath $0.000 \pm 0.000$} & $2.713 \pm 0.125$ & $75.108 \pm 10.712$ & $75.044 \pm 10.708$ \\
 & GD & $11.511 \pm 3.849$ & $423.482 \pm 126.094$ & $226.015 \pm 39.074$ & {\boldmath $0.000 \pm 0.000$} & $9.493 \pm 2.898$ & $375.206 \pm 94.820$ & $228.092 \pm 37.746$ \\
 & Learned & {\boldmath $1.468 \pm 0.069$} & $192.520 \pm 34.445$ & $102.301 \pm 17.594$ & {\boldmath $0.000 \pm 0.000$} & {\boldmath $2.203 \pm 0.071$} & $172.436 \pm 29.983$ & $115.500 \pm 18.448$ \\
 & Momentum & $16.992 \pm 4.154$ & $546.558 \pm 193.823$ & $225.813 \pm 43.235$ & {\boldmath $0.000 \pm 0.000$} & $14.356 \pm 3.198$ & $472.217 \pm 151.929$ & $227.131 \pm 41.940$ \\
 & NAG & $19.324 \pm 1.795$ & $646.385 \pm 109.594$ & $234.128 \pm 38.804$ & {\boldmath $0.000 \pm 0.000$} & $16.945 \pm 1.715$ & $578.123 \pm 107.969$ & $234.412 \pm 38.778$ \\
 & RMSProp & $2.796 \pm 0.145$ & {\boldmath $8.299 \pm 0.824$} & {\boldmath $8.299 \pm 0.824$} & {\boldmath $0.000 \pm 0.000$} & $2.778 \pm 0.141$ & {\boldmath $19.530 \pm 2.078$} & {\boldmath $19.530 \pm 2.078$} \\
\addlinespace[2pt]
\multicolumn{9}{l}{\textit{Rastrigin ($d=100$)}} \\
 & Adam & $3.045 \pm 0.405$ & $14.003 \pm 2.529$ & $14.003 \pm 2.529$ & {\boldmath $0.005 \pm 0.014$} & {\boldmath $2.980 \pm 0.340$} & $240.776 \pm 37.823$ & $240.755 \pm 37.820$ \\
 & GD & $30.736 \pm 7.475$ & $2406.321 \pm 543.681$ & $1058.369 \pm 159.683$ & $0.000 \pm 0.000$ & $25.015 \pm 5.707$ & $2074.005 \pm 408.555$ & $1058.386 \pm 159.695$ \\
 & Learned & {\boldmath $2.437 \pm 0.101$} & $673.918 \pm 117.713$ & $443.705 \pm 73.797$ & $0.000 \pm 0.000$ & $3.869 \pm 0.203$ & $711.503 \pm 120.807$ & $489.722 \pm 79.549$ \\
 & Momentum & $39.406 \pm 6.219$ & $2826.351 \pm 685.981$ & $1057.645 \pm 160.794$ & $0.000 \pm 0.000$ & $33.055 \pm 5.304$ & $2436.860 \pm 577.485$ & $1057.690 \pm 160.768$ \\
 & NAG & $39.003 \pm 2.674$ & $2924.122 \pm 387.284$ & $1058.635 \pm 160.268$ & $0.000 \pm 0.000$ & $35.244 \pm 2.554$ & $2711.828 \pm 374.436$ & $1058.647 \pm 160.269$ \\
 & RMSProp & $3.035 \pm 0.402$ & {\boldmath $10.947 \pm 2.303$} & {\boldmath $10.947 \pm 2.303$} & {\boldmath $0.005 \pm 0.014$} & $3.020 \pm 0.391$ & {\boldmath $50.345 \pm 7.440$} & {\boldmath $50.345 \pm 7.440$} \\
\addlinespace[2pt]
\multicolumn{9}{l}{\textit{Rastrigin ($d=500$)}} \\
 & Adam & $0.425 \pm 0.214$ & $2.289 \pm 0.631$ & $2.289 \pm 0.631$ & {\boldmath $0.677 \pm 0.164$} & $0.880 \pm 0.177$ & $269.633 \pm 50.044$ & $269.633 \pm 50.044$ \\
 & GD & $68.899 \pm 17.474$ & $1.21e\!4\! \pm 2963.870$ & $2281.595 \pm 381.549$ & $0.000 \pm 0.000$ & $56.180 \pm 13.478$ & $1.04e4 \pm 2264.387$ & $2281.595 \pm 381.549$ \\
 & Learned & $6.992 \pm 0.078$ & $4471.672 \pm 698.520$ & $1901.137 \pm 312.086$ & $0.000 \pm 0.000$ & $10.218 \pm 0.040$ & $3838.967 \pm 576.628$ & $1916.049 \pm 314.890$ \\
 & Momentum & $90.777 \pm 16.515$ & $1.47e4 \pm 3961.116$ & $2281.595 \pm 381.549$ & $0.000 \pm 0.000$ & $76.336 \pm 13.997$ & $1.26e4 \pm 3311.684$ & $2281.595 \pm 381.549$ \\
 & NAG & $87.333 \pm 4.673$ & $1.47e4 \pm 1767.227$ & $2281.595 \pm 381.549$ & $0.000 \pm 0.000$ & $79.442 \pm 5.006$ & $1.37e4 \pm 1857.393$ & $2281.595 \pm 381.549$ \\
 & RMSProp & {\boldmath $0.387 \pm 0.226$} & {\boldmath $0.590 \pm 0.407$} & {\boldmath $0.590 \pm 0.407$} & {\boldmath $0.677 \pm 0.164$} & {\boldmath $0.466 \pm 0.220$} & {\boldmath $56.395 \pm 9.620$} & {\boldmath $56.395 \pm 9.620$} \\
\bottomrule
\end{tabular}
}
\end{table}

\begin{figure}[ht]
    \centering
\includegraphics[width=0.32\linewidth]{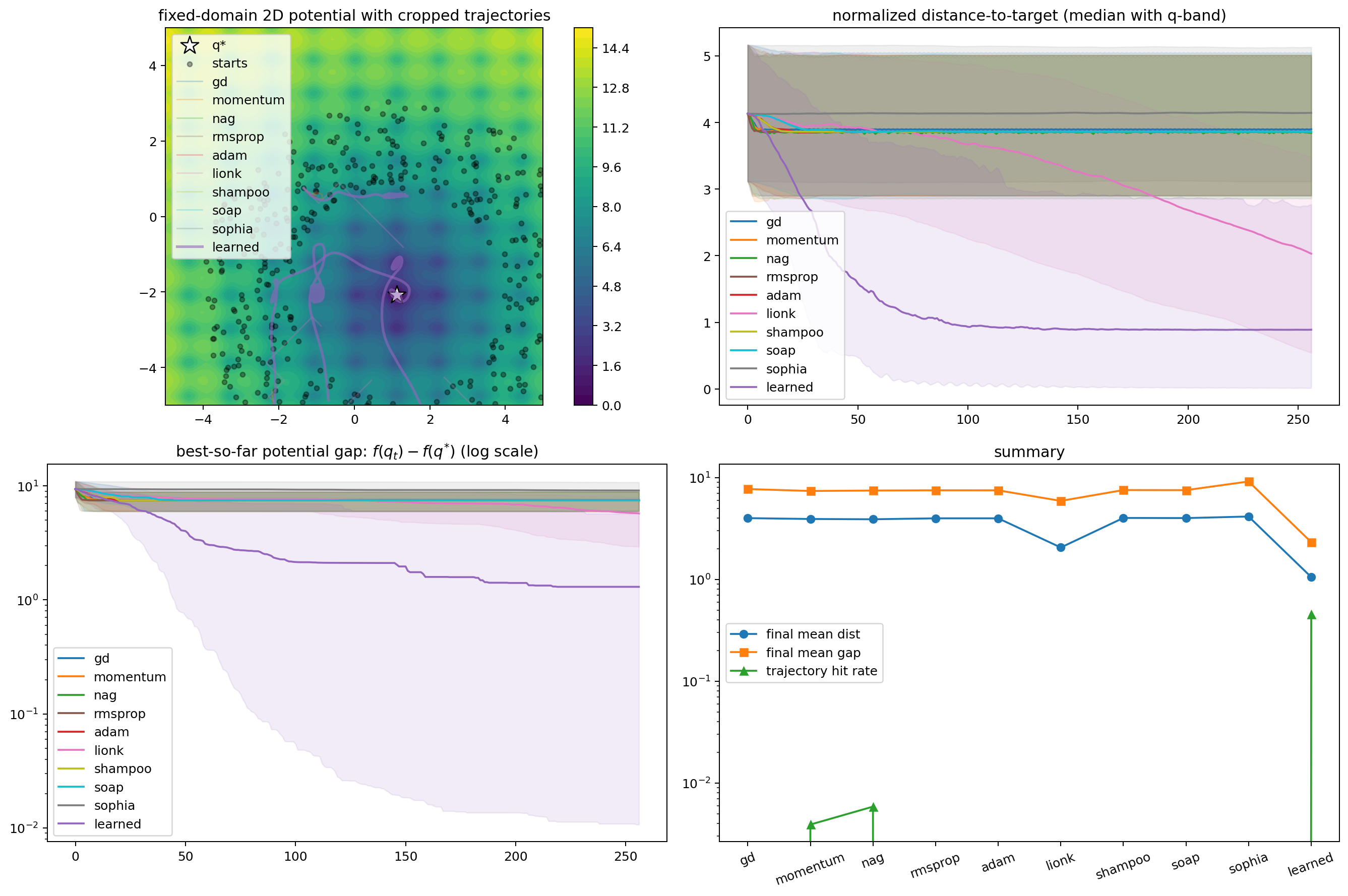}
\includegraphics[width=0.32\linewidth]{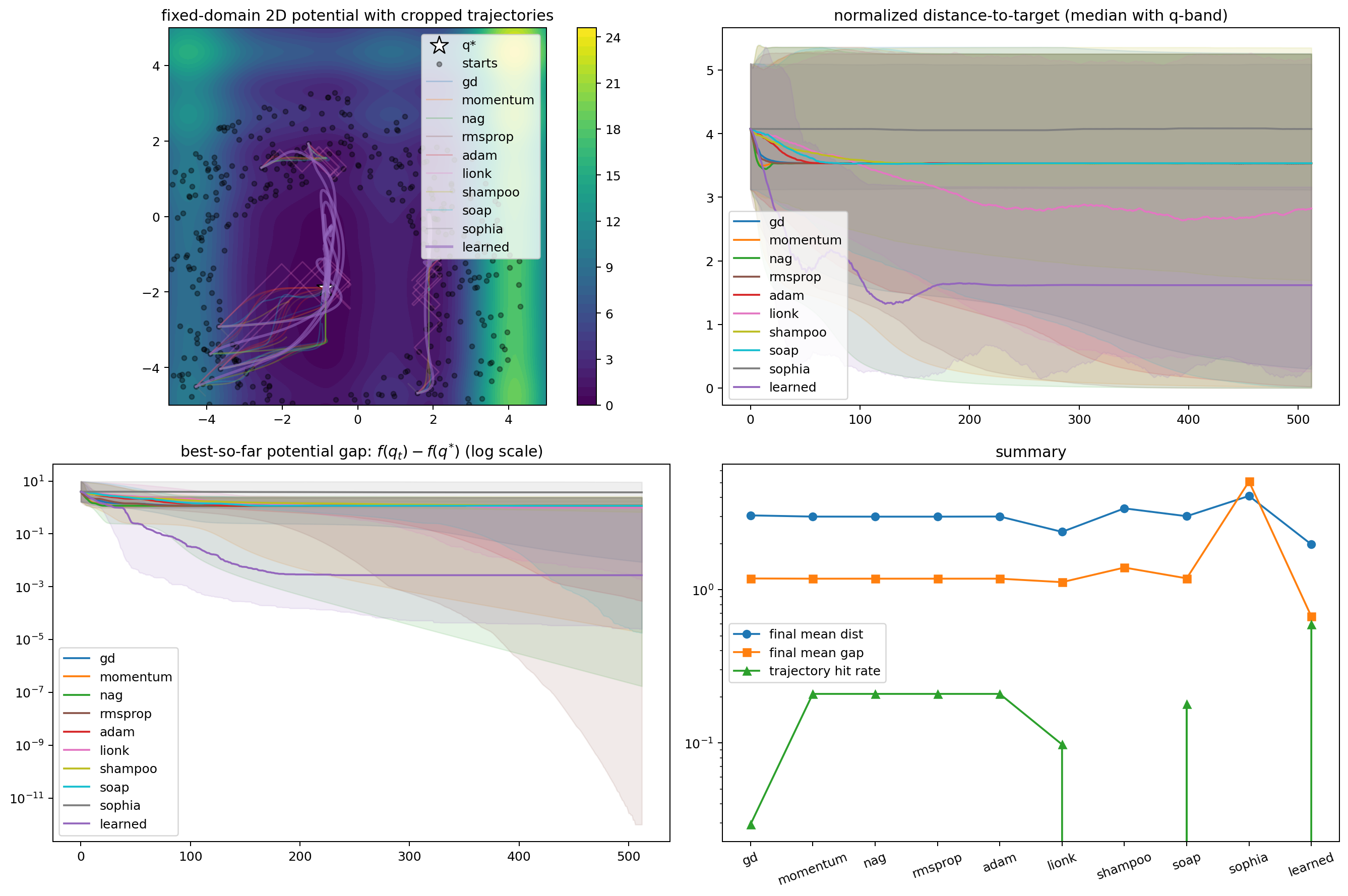}
\includegraphics[width=0.32\linewidth]{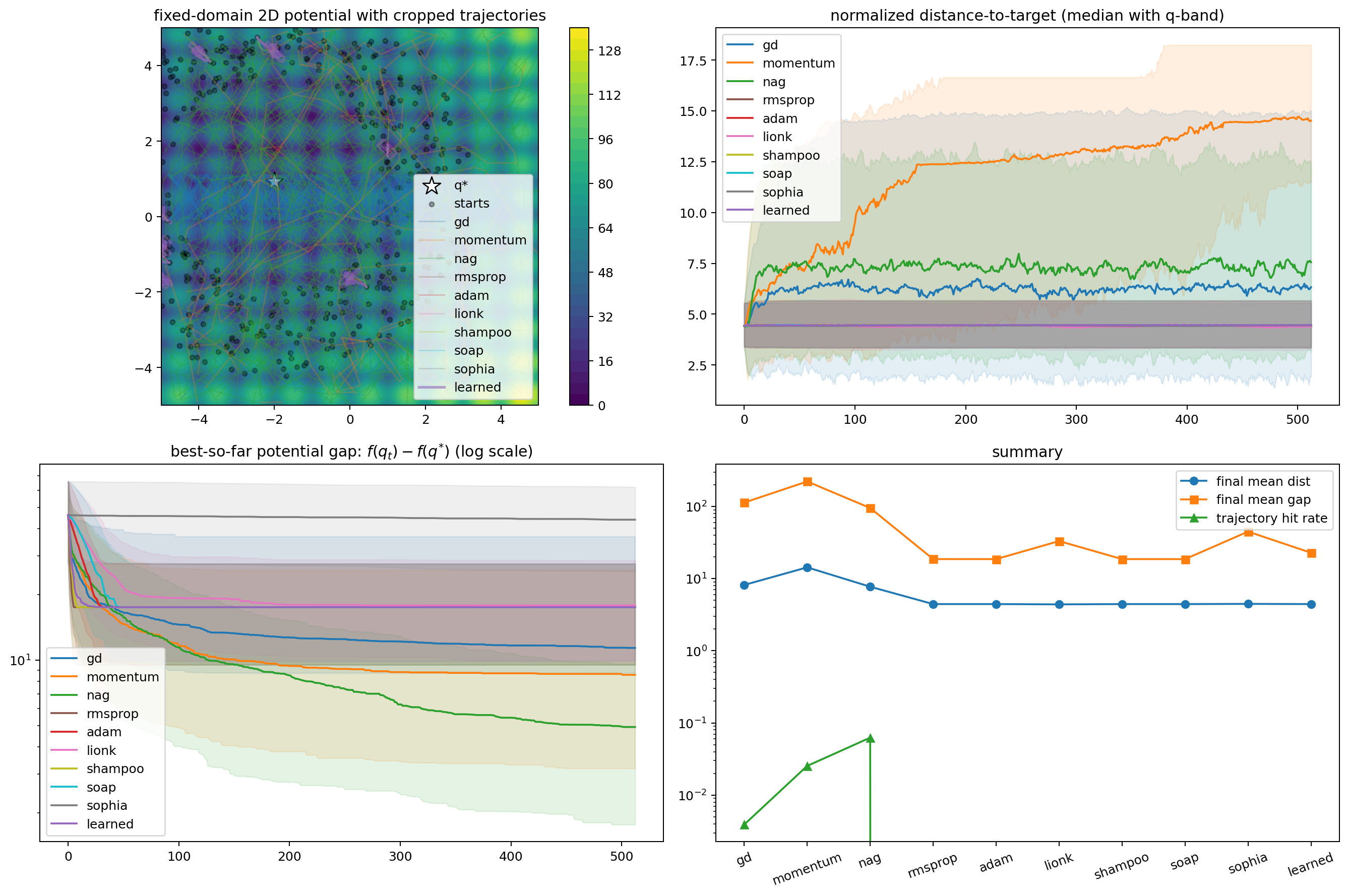}    
    \caption{Illustrative Result in 2D example. Left: Ackley; Middle: Levy; Right: Rastrigin.}
    \label{fig:2d}
\end{figure}

\begin{figure}[ht]
    \centering
\includegraphics[width=0.3\linewidth]{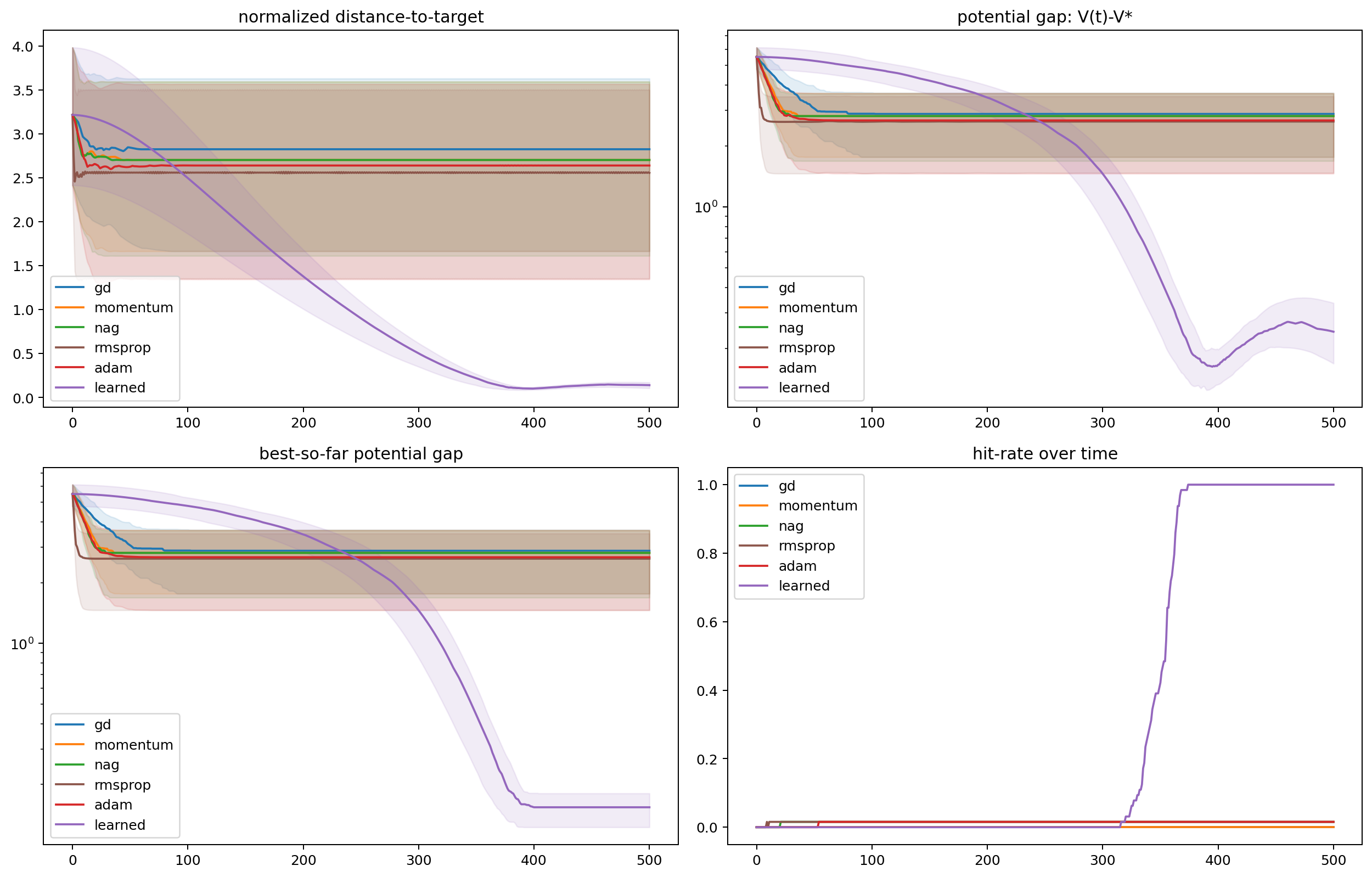}
\includegraphics[width=0.3\linewidth]{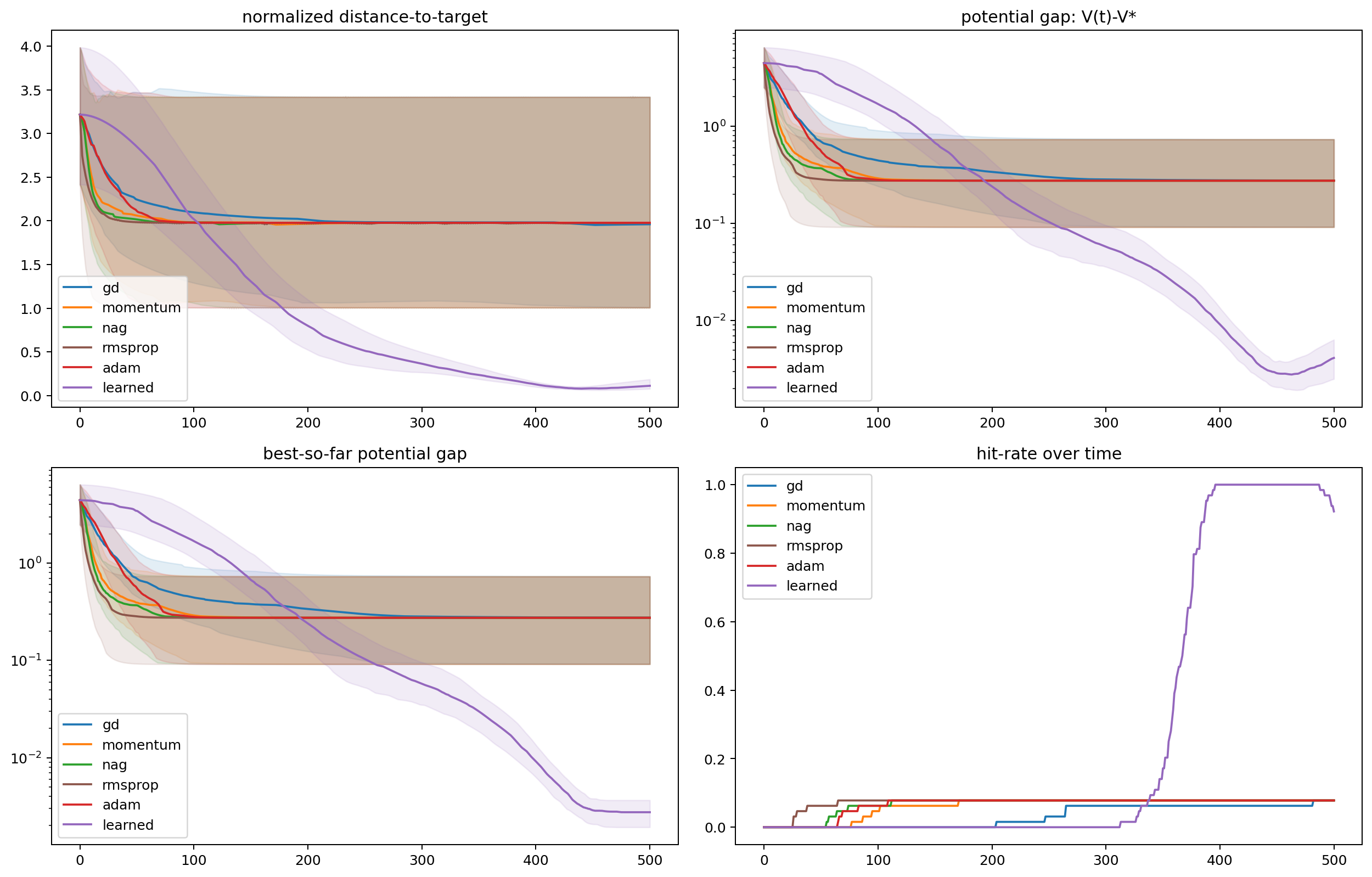}
\includegraphics[width=0.3\linewidth]{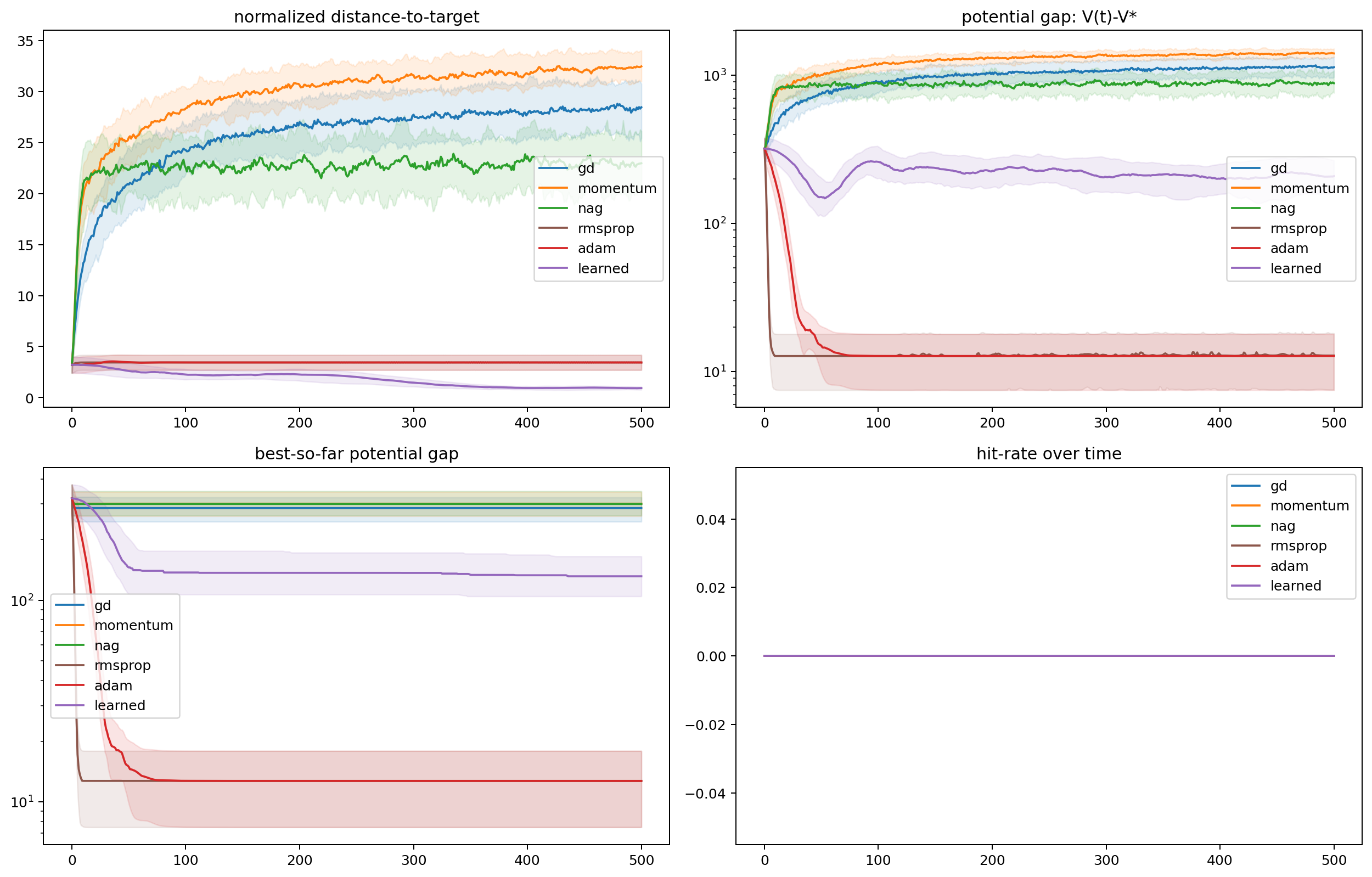}    
    \caption{Illustrative Result in 20D example. Left: Ackley; Middle: Levy; Right: Rastrigin.}
    \label{fig:20d}
\end{figure}


\begin{figure}[ht]
    \centering
    \includegraphics[width=0.48\linewidth]{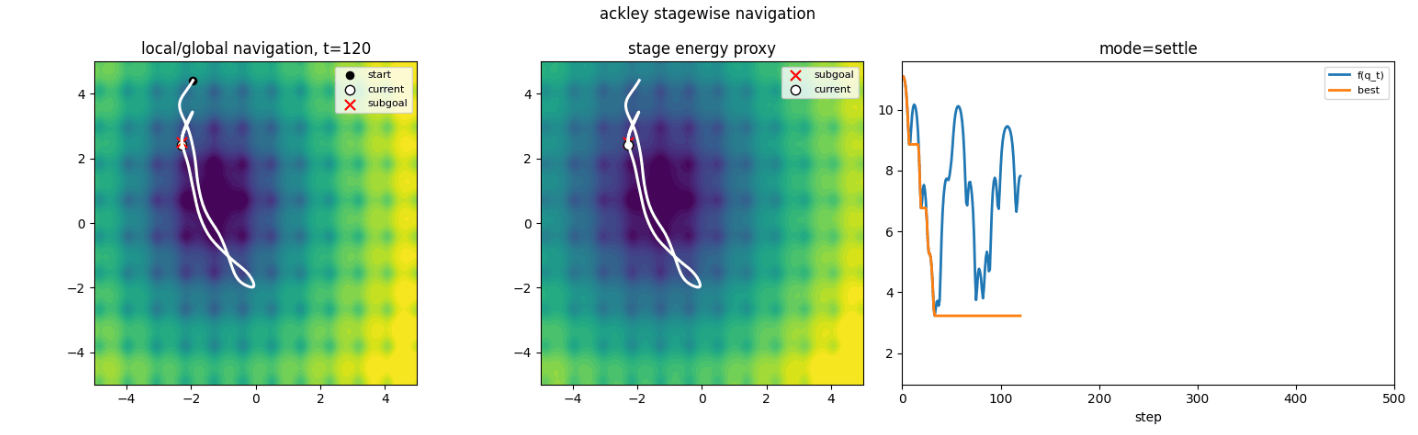}
    \includegraphics[width=0.48\linewidth]{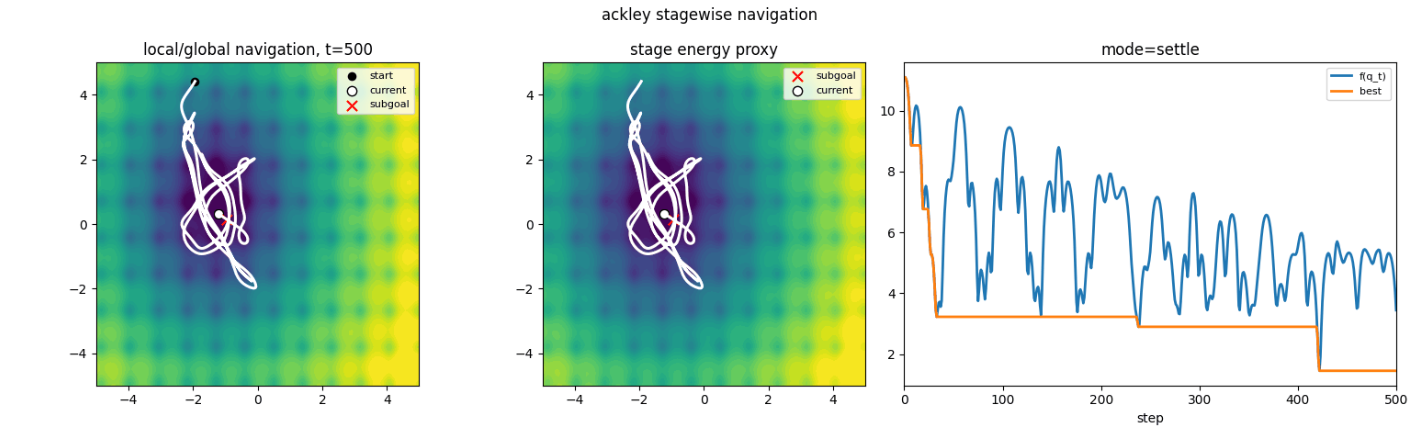}
    \caption{We display a single trajectory traversion based on reshaped energy landscape from trained SHAPE optimizer. Left: When running $120$ total steps, there exists a local information drive force that reshape the second column of the stage energy and the optimizer is taking descent steps based on ``memorized" energy. When it stabilizes (as $480$ steps passed on Right panel), the energy proxy also converges to the true local energy landscape and either the optimizer halts or learn to propose new descent dynamics.}
    \label{fig:navigation:display}
\end{figure}

\paragraph{Differentiable Scientific and Engineering Task Details}
\label{subsec:real_world_tasks}

We now turn to task families that are scientifically meaningful and remain differentiable end-to-end. These experiments use exactly the same dual-policy training template as the synthetic studies, but the objectives are no longer hand-designed benchmark landscapes. Instead, each family contributes its own oracle geometry, initialization law, and notion of task variation. The role of this subsection is therefore twofold: it tests whether the planner/local-controller decomposition transfers beyond toy functions, and it checks whether the common metric bundle used in the tables remains informative once the objective has genuine modeling content.

\paragraph{Lennard--Jones cluster energy.}
The Lennard--Jones (LJ) potential is a classical pairwise interaction model for
neutral atoms and molecules, and has long served as a standard testbed for
cluster geometry optimization and rugged energy-landscape exploration
\cite{Jones1924a,Jones1924b,Northby1987,WalesDoye1997,MullerSbalzarini2012}.
Let $N_a$ denote the number of particles and let
\[
q=(x_1,\dots,x_{N_a})\in\mathbb{R}^{3N_a},\qquad x_i\in\mathbb{R}^3.
\]
For each task $\tau$, the objective is the smooth pairwise interaction energy
\begin{equation}
  f_\tau(q)
  =
  4\varepsilon_\tau \sum_{1\le i<j\le N_a}
  \left[
  \left(\frac{\sigma_\tau}{\|x_i-x_j\|}\right)^{12}
  -
  \left(\frac{\sigma_\tau}{\|x_i-x_j\|}\right)^6
  \right].
  \label{eq:lj_energy}
\end{equation}

The corresponding pair potential
\(
v_\tau(r)=4\varepsilon_\tau\Big[\big(\sigma_\tau/r\big)^{12}
-\big(\sigma_\tau/r\big)^6\Big]
\) has a unique minima at
\(
r_\tau^\star = 2^{1/6}\sigma_\tau,
\)
with value $-\varepsilon_\tau$. Hence low-energy configurations favor compact
packings whose typical nearest-neighbor distances concentrate near
$r_\tau^\star$, while the steep $r^{-12}$ core prevents collapse. Since
$f_\tau$ depends only on pairwise distances, it is invariant under global
translations, global rotations, and particle relabelings; the relevant geometry
is therefore the internal cluster shape rather than absolute pose. In the size
regime considered here, one typically expects compact shell-forming motifs and,
at magic sizes such as $N_a=13$, highly symmetric icosahedral arrangements.
More generally, the landscape contains many metastable isomers and may exhibit
funnel or multi-funnel organization, which makes LJ clusters a stringent test of
nonconvex search and basin-to-basin rearrangement
\cite{Northby1987,WalesDoye1997,DoyeMillerWales1999}.


In the summary table we report the same primary metrics used elsewhere: final distance to a reference optimum or terminal proxy, final gap, best-so-far gap, hit rate, and the corresponding AUC statistics. For LJ these primary metrics are supplemented in our qualitative analysis by the physical interpretation of the visited configurations: compactness, radius of gyration, and qualitative basin mobility. This family is particularly suitable for the proposed method because it combines smooth local pair forces with a rugged global isomer landscape, exactly the regime in which a slow coarse planner and a fast local stabilizer should be complementary.

\paragraph{Phase retrieval.}
Let $A_\tau\in\mathbb{R}^{m\times n}$ denote a sensing matrix and let $x_\tau^\star\in\mathbb{R}^n$ denote a latent signal.
The observations are
\[
  b_i = |\langle a_i,x_\tau^\star\rangle|^2 + \eta_i,
  \qquad i=1,\dots,m,
\]
where $a_i^\top$ is the $i$-th row of $A_\tau$.
Each task minimizes the quartic least-squares objective
\begin{equation}
  f_\tau(x)
  =
  \frac{1}{m}\sum_{i=1}^m
  \left(
  |\langle a_i,x\rangle|^2-b_i
  \right)^2
  + \frac{\lambda_x}{2}\|x\|^2,
  \qquad x\in\mathbb{R}^n.
  \label{eq:phase_retrieval_obj}
\end{equation}
The task distribution varies the sensing design \(A_\tau\), the signal class, and the observation noise level. A typical meta-train/test split samples Gaussian or coded-diffraction-type matrices during training and holds out different sensing realizations and signal draws during evaluation. As in the synthetic section, training follows the staged curriculum: the local policy first learns short-horizon improvement on easier low-noise instances, then the planner and memory are exposed to harder sensing patterns and broader initializations, and finally the full system is trained end-to-end.

The online oracle returns either the full gradient of \eqref{eq:phase_retrieval_obj} or a mini-batch gradient formed from a subset of measurements. Because the objective is invariant to the global sign of \(x_\tau^\star\), we use the sign-invariant reconstruction distance only as an auxiliary diagnostic. The primary numbers reported in the table remain the common metric bundle (final distance, final gap, best gap, hit rate, and the three AUC statistics), so that phase retrieval can be compared directly with the other task families without changing the reporting format. This task probes whether memory and slow replanning remain useful when the landscape is induced indirectly by measurements rather than by an explicit low-dimensional potential.

\paragraph{Control-based objective via direct trajectory optimization.}
To bridge optimization and control, we consider finite-horizon nonlinear trajectory optimization problems. Let the state and
control satisfy
\begin{equation}
  x_{t+1} = \varphi_\tau(x_t,u_t),
  \qquad t=0,\dots,T-1,
  \label{eq:control_dynamics}
\end{equation}
with initial condition $x_0\sim \rho_{0,\tau}$. The decision variable is the stacked control sequence
\[
q=(u_0,\dots,u_{T-1})\in\mathbb{R}^{m_u T}.
\]
The objective is
\begin{equation}
  f_\tau(q)
  =
  \Phi_\tau(x_T)
  +
  \sum_{t=0}^{T-1}
  \ell_\tau(x_t,u_t)
  +
  \lambda_u \sum_{t=0}^{T-1}\|u_t\|^2,
  \label{eq:control_obj}
\end{equation}
where the trajectory $(x_t)$ is induced by \eqref{eq:control_dynamics}.
The task distribution varies system parameters, initial states, target sets, obstacle layouts, and penalty weights. In the reported experiments we instantiate this family on low-dimensional systems such as a double integrator, pendulum swing-up, and moderately parameterized navigation problems, so that the total decision dimension remains comparable to the earlier benchmarks.

Gradients are obtained by automatic differentiation through the rollout or by a discrete adjoint. Again, the table reports the same common metric bundle as in the synthetic section. In this family ``distance'' is measured relative to the task-specific terminal target proxy, while the gap quantities are computed from the trajectory cost \eqref{eq:control_obj}. This makes the control study the most direct test of the dual-policy interpretation: the slow planner proposes globally useful control-sequence directions, whereas the fast local controller must still stabilize short-horizon movement in the decision space under the trajectory dynamics.

\begin{table}[!htbp]
\centering
\small
\setlength{\tabcolsep}{4pt}
\caption{Benchmark summary in mean$\pm$std format. The columns \texttt{scale}, \texttt{dim}, and \texttt{n\_instances} are omitted from the table body. Dimensions are encoded by the low/high blocks: for LJ cluster, low $d=6$ and high $d=18$; for phase retrieval and control trajectory optimization, low $d=8$ and high $d=32$. Entries in bold indicate the best mean in each metric column within each task-and-dimension block.}
\label{tab:real:world:example:summary}
\resizebox{\textwidth}{!}{%
\begin{tabular}{llccccccc}
\toprule
Task & Method & Final dist. $\downarrow$ & Final gap $\downarrow$ & Best gap $\downarrow$ & Hit rate $\uparrow$ & AUC dist. $\downarrow$ & AUC gap $\downarrow$ & AUC best gap $\downarrow$ \\
\midrule
\multicolumn{9}{l}{\textit{Low-dimensional ($d=6$)}} \\
LJ cluster & Adam & $2.812 \pm 0.182$ & $1.167 \pm 0.445$ & $1.167 \pm 0.445$ & $0.000 \pm 0.000$ & $2.928 \pm 0.189$ & $1.266 \pm 0.472$ & $1.266 \pm 0.472$ \\
 & GD & $2.953 \pm 0.194$ & $1.249 \pm 0.448$ & $1.249 \pm 0.448$ & $0.000 \pm 0.000$ & $3.000 \pm 0.194$ & $1.295 \pm 0.462$ & $1.295 \pm 0.462$ \\
 & Learned & {\boldmath$0.324 \pm 0.138$} & {\boldmath$0.213 \pm 0.222$} & {\boldmath$0.020 \pm 0.023$} & {\boldmath$0.353 \pm 0.371$} & {\boldmath$1.745 \pm 0.062$} & {\boldmath$0.740 \pm 0.345$} & {\boldmath$0.692 \pm 0.307$} \\
 & Momentum & $2.847 \pm 0.194$ & $1.172 \pm 0.430$ & $1.170 \pm 0.428$ & $0.000 \pm 0.000$ & $2.948 \pm 0.194$ & $1.255 \pm 0.453$ & $1.253 \pm 0.451$ \\
 & NAG & $2.802 \pm 0.196$ & $1.140 \pm 0.422$ & $1.137 \pm 0.420$ & $0.000 \pm 0.000$ & $2.928 \pm 0.194$ & $1.240 \pm 0.449$ & $1.237 \pm 0.447$ \\
 & RMSProp & $2.640 \pm 0.174$ & $1.018 \pm 0.413$ & $1.018 \pm 0.413$ & $0.000 \pm 0.000$ & $2.788 \pm 0.181$ & $1.143 \pm 0.442$ & $1.143 \pm 0.442$ \\
\addlinespace[2pt]
\multicolumn{9}{l}{\textit{High-dimensional ($d=18$)}} \\
 & Adam & $0.666 \pm 0.073$ & $0.130 \pm 0.118$ & $0.130 \pm 0.118$ & $0.000 \pm 0.000$ & $0.632 \pm 0.062$ & $0.912 \pm 0.421$ & $0.912 \pm 0.421$ \\
 & GD & $0.686 \pm 0.076$ & $0.032 \pm 0.036$ & $0.032 \pm 0.036$ & $0.004 \pm 0.014$ & $0.644 \pm 0.066$ & $0.338 \pm 0.137$ & $0.333 \pm 0.136$ \\
 & Learned & {\boldmath$0.509 \pm 0.041$} & $0.295 \pm 0.201$ & $0.207 \pm 0.151$ & {\boldmath$0.026 \pm 0.040$} & {\boldmath$0.585 \pm 0.045$} & $0.933 \pm 0.405$ & $0.796 \pm 0.344$ \\
 & Momentum & $0.867 \pm 0.107$ & $0.074 \pm 0.080$ & $0.073 \pm 0.079$ & $0.000 \pm 0.000$ & $0.768 \pm 0.073$ & $0.434 \pm 0.239$ & $0.388 \pm 0.215$ \\
 & NAG & $0.964 \pm 0.122$ & $0.153 \pm 0.141$ & $0.135 \pm 0.125$ & $0.000 \pm 0.000$ & $0.849 \pm 0.088$ & $0.533 \pm 0.309$ & $0.426 \pm 0.242$ \\
 & RMSProp & $0.820 \pm 0.107$ & {\boldmath$0.021 \pm 0.035$} & {\boldmath$0.021 \pm 0.035$} & $0.000 \pm 0.000$ & $0.705 \pm 0.079$ & {\boldmath$0.291 \pm 0.165$} & {\boldmath$0.291 \pm 0.165$} \\
\midrule
\multicolumn{9}{l}{\textit{Low-dimensional ($d=8$)}} \\
Phase retrieval & Adam & $2.110 \pm 0.116$ & $0.250 \pm 0.082$ & $0.250 \pm 0.082$ & $0.000 \pm 0.000$ & $2.444 \pm 0.097$ & $0.964 \pm 0.269$ & $0.964 \pm 0.269$ \\
 & GD & $2.135 \pm 0.110$ & $0.165 \pm 0.039$ & $0.165 \pm 0.039$ & $0.000 \pm 0.000$ & $2.388 \pm 0.096$ & $0.529 \pm 0.077$ & $0.529 \pm 0.077$ \\
 & Learned & {\boldmath$0.268 \pm 0.038$} & {\boldmath$0.012 \pm 0.012$} & {\boldmath$0.007 \pm 0.007$} & {\boldmath$0.262 \pm 0.157$} & {\boldmath$1.731 \pm 0.102$} & $0.790 \pm 0.150$ & $0.789 \pm 0.149$ \\
 & Momentum & $1.788 \pm 0.140$ & $0.052 \pm 0.026$ & $0.052 \pm 0.026$ & $0.000 \pm 0.000$ & $2.056 \pm 0.117$ & $0.258 \pm 0.051$ & $0.258 \pm 0.050$ \\
 & NAG & $1.737 \pm 0.147$ & $0.045 \pm 0.024$ & $0.045 \pm 0.024$ & $0.000 \pm 0.000$ & $2.019 \pm 0.120$ & {\boldmath$0.258 \pm 0.051$} & {\boldmath$0.258 \pm 0.051$} \\
 & RMSProp & $1.850 \pm 0.153$ & $0.087 \pm 0.030$ & $0.087 \pm 0.030$ & $0.000 \pm 0.000$ & $2.113 \pm 0.115$ & $0.379 \pm 0.110$ & $0.379 \pm 0.110$ \\
\addlinespace[2pt]
\multicolumn{9}{l}{\textit{High-dimensional ($d=32$)}} \\
 & Adam & $1.843 \pm 0.065$ & $0.005 \pm 0.001$ & $0.005 \pm 0.001$ & $0.000 \pm 0.000$ & $2.197 \pm 0.060$ & $0.032 \pm 0.004$ & $0.032 \pm 0.004$ \\
 & GD & $2.761 \pm 0.068$ & $0.098 \pm 0.007$ & $0.098 \pm 0.007$ & $0.000 \pm 0.000$ & $2.848 \pm 0.070$ & $0.131 \pm 0.012$ & $0.131 \pm 0.012$ \\
 & Learned & {\boldmath$0.068 \pm 0.017$} & {\boldmath$2.08\mathrm{e}{-5} \pm 4.43\mathrm{e}{-6}$} & {\boldmath$2.08\mathrm{e}{-5} \pm 4.43\mathrm{e}{-6}$} & {\boldmath$1.000 \pm 0.000$} & {\boldmath$1.465 \pm 0.050$} & $0.049 \pm 0.006$ & $0.049 \pm 0.006$ \\
 & Momentum & $2.500 \pm 0.062$ & $0.042 \pm 0.002$ & $0.042 \pm 0.002$ & $0.000 \pm 0.000$ & $2.690 \pm 0.066$ & $0.086 \pm 0.006$ & $0.086 \pm 0.006$ \\
 & NAG & $2.448 \pm 0.061$ & $0.035 \pm 0.002$ & $0.035 \pm 0.002$ & $0.000 \pm 0.000$ & $2.660 \pm 0.065$ & $0.080 \pm 0.006$ & $0.080 \pm 0.006$ \\
 & RMSProp & $1.611 \pm 0.072$ & $0.002 \pm 6.04\mathrm{e}{-4}$ & $0.002 \pm 6.04\mathrm{e}{-4}$ & $0.000 \pm 0.000$ & $1.870 \pm 0.060$ & {\boldmath$0.011 \pm 0.002$} & {\boldmath$0.011 \pm 0.002$} \\
\midrule
\multicolumn{9}{l}{\textit{Low-dimensional ($d=8$)}} \\
Control trajopt & Adam & $2.270 \pm 0.264$ & $2.252 \pm 0.594$ & $2.248 \pm 0.592$ & $0.000 \pm 0.000$ & $2.599 \pm 0.171$ & $9.279 \pm 4.167$ & $9.260 \pm 4.152$ \\
 & GD & $2.969 \pm 5.933$ & $2718.1 \pm 9095.6$ & $6.902 \pm 21.552$ & $0.000 \pm 0.000$ & $3.324 \pm 4.975$ & $2315.7 \pm 7862.3$ & $8.193 \pm 21.166$ \\
 & Learned & $0.493 \pm 0.187$ & $1.440 \pm 1.374$ & $0.115 \pm 0.123$ & $0.083 \pm 0.205$ & $1.490 \pm 0.377$ & $7.181 \pm 3.896$ & $3.927 \pm 1.319$ \\
 & Momentum & {\boldmath$0.233 \pm 0.149$} & {\boldmath$0.014 \pm 0.013$} & {\boldmath$0.014 \pm 0.013$} & $0.533 \pm 0.350$ & {\boldmath$0.965 \pm 0.204$} & {\boldmath$1.993 \pm 1.602$} & {\boldmath$1.377 \pm 0.975$} \\
 & NAG & $3.842 \pm 8.293$ & $4589.6 \pm 1.05\mathrm{e}{+4}$ & $5.976 \pm 17.580$ & {\boldmath$0.634 \pm 0.381$} & $3.585 \pm 6.370$ & $3504.9 \pm 8785.3$ & $6.796 \pm 17.379$ \\
 & RMSProp & $1.753 \pm 0.404$ & $1.217 \pm 0.436$ & $1.217 \pm 0.436$ & $0.001 \pm 0.006$ & $2.198 \pm 0.301$ & $3.610 \pm 1.346$ & $3.609 \pm 1.345$ \\
\addlinespace[2pt]
\multicolumn{9}{l}{\textit{High-dimensional ($d=32$)}} \\
 & Adam & $2.140 \pm 0.509$ & $1.865 \pm 0.968$ & $1.852 \pm 0.957$ & $0.000 \pm 0.000$ & $2.542 \pm 0.282$ & $6.480 \pm 4.647$ & $5.882 \pm 4.041$ \\
 & GD & $26.817 \pm 21.580$ & $3.20\mathrm{e}{+5} \pm 3.69\mathrm{e}{+5}$ & $47.326 \pm 54.954$ & $0.000 \pm 0.000$ & $25.486 \pm 20.063$ & $3.06\mathrm{e}{+5} \pm 3.59\mathrm{e}{+5}$ & $47.840 \pm 54.519$ \\
 & Learned & {\boldmath$0.233 \pm 0.105$} & $3.549 \pm 6.163$ & {\boldmath$0.097 \pm 0.113$} & {\boldmath$0.548 \pm 0.403$} & {\boldmath$1.288 \pm 0.371$} & $12.671 \pm 13.876$ & $3.566 \pm 2.099$ \\
 & Momentum & $20.869 \pm 22.342$ & $3.03\mathrm{e}{+5} \pm 3.80\mathrm{e}{+5}$ & $44.297 \pm 56.830$ & $0.197 \pm 0.373$ & $20.385 \pm 21.014$ & $2.91\mathrm{e}{+5} \pm 3.68\mathrm{e}{+5}$ & $44.810 \pm 56.431$ \\
 & NAG & -- & -- & -- & $0.203 \pm 0.371$ & -- & -- & -- \\
 & RMSProp & $1.866 \pm 0.729$ & {\boldmath$1.451 \pm 1.082$} & $1.451 \pm 1.082$ & $0.003 \pm 0.012$ & $2.282 \pm 0.520$ & {\boldmath$3.411 \pm 2.141$} & {\boldmath$3.115 \pm 1.841$} \\
\bottomrule
\end{tabular}
}
\end{table}

The evaluation protocol is identical across the three real-world families. A task \(\tau\) is sampled from the family-specific distribution, an initial iterate \(q_0\sim\rho_{0,\tau}\) is drawn, and the learned controller is rolled out for a fixed number of oracle queries using the same event-triggered update rule as in the synthetic experiments. The model is always trained with the three-phase curriculum from the method section and evaluated zero-shot on held-out tasks. We compare against first-order baselines under matched query budgets and report mean\(\pm\)standard deviation over tasks and random seeds, exactly as shown in \cref{tab:real:world:example:summary}. When a task family admits symmetries---for example translation/rotation/permutation invariance in Lennard--Jones clusters, sign symmetry in phase retrieval, or task-equivalent terminal solutions in control---the distance metric is computed on the corresponding quotient whenever this is well defined.

\begin{figure}
    \centering
    \includegraphics[width=0.45\linewidth]{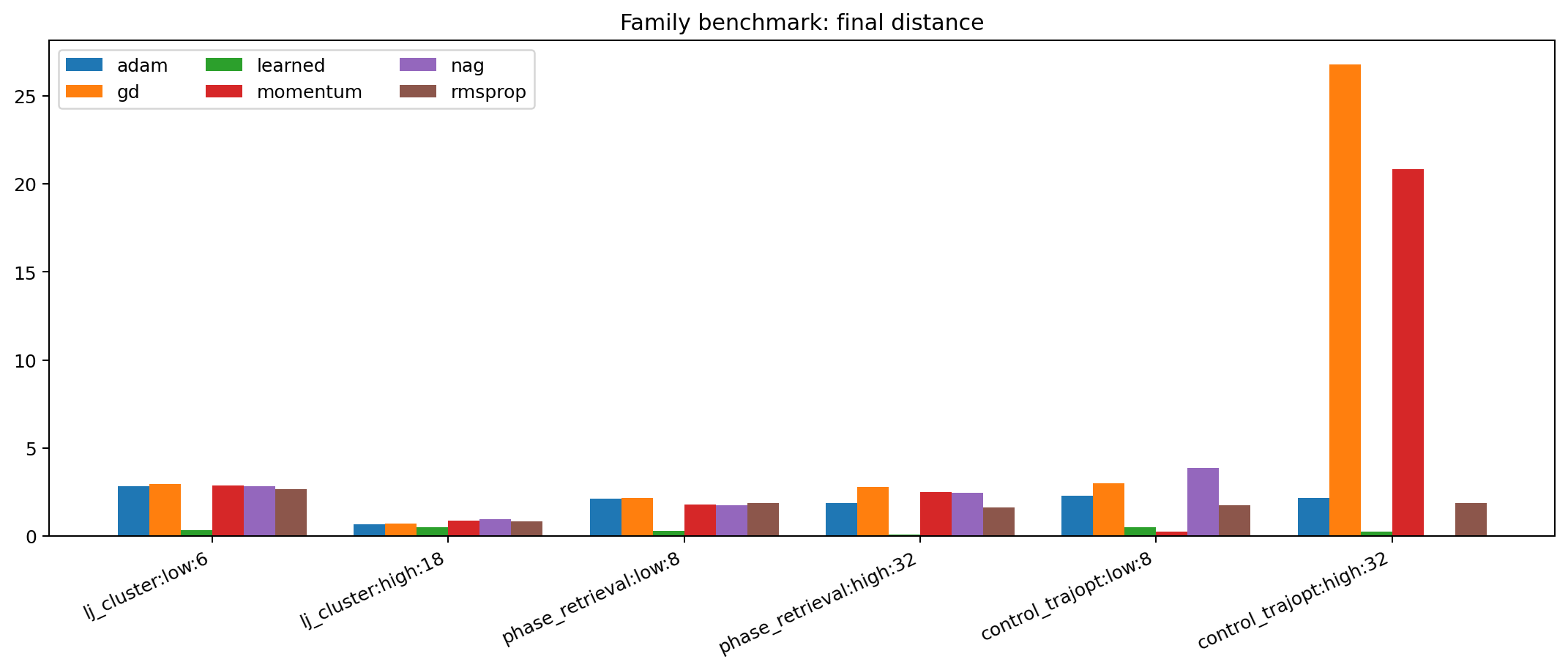}
    \includegraphics[width=0.45\linewidth]{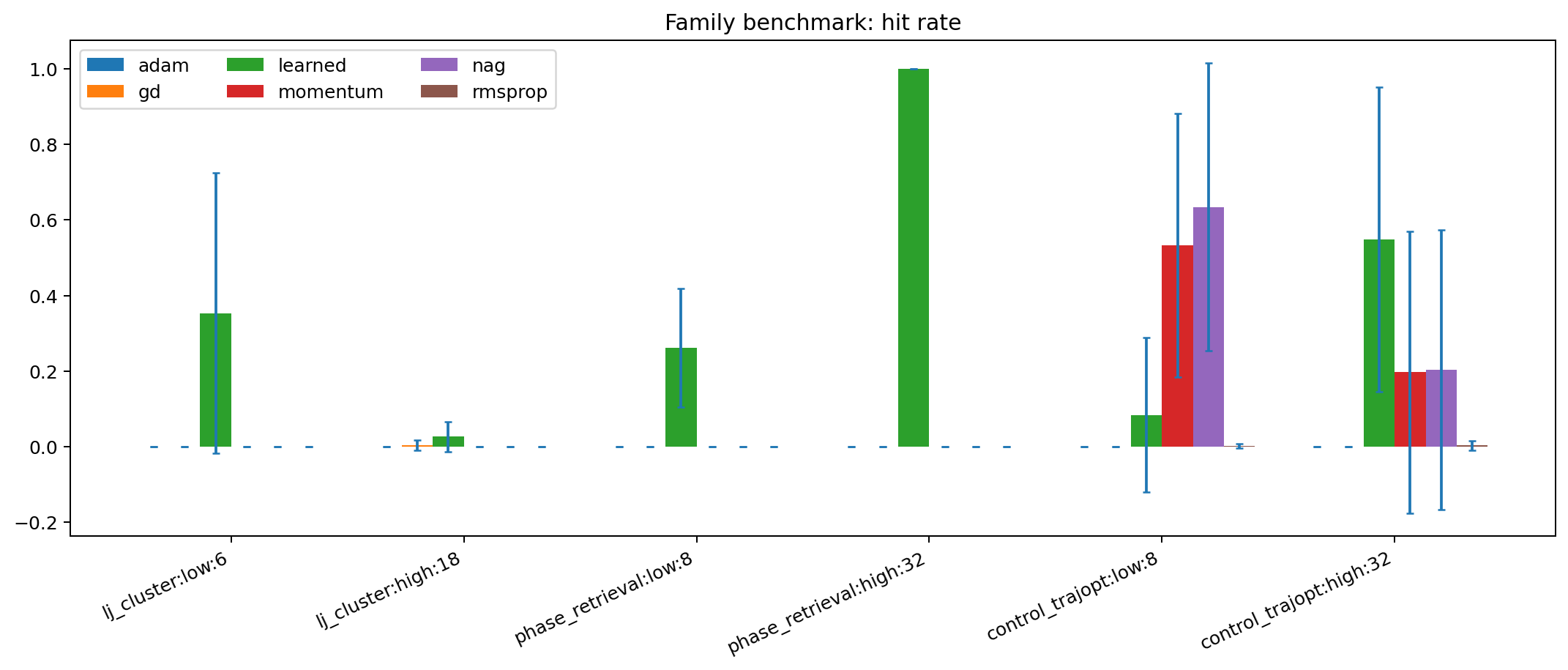}
    \caption{Optimizer performance on each real-world differentiable task. We report the average distance to a minimum and the minimum-hitting rate.}
    \label{fig:real:world:display}
\end{figure}

\subsection{More Ablation Results and Visualization}
\label{subsec:additional:ablation}

\begin{figure}[htbp]
    \centering
    \includegraphics[width=0.999\linewidth]{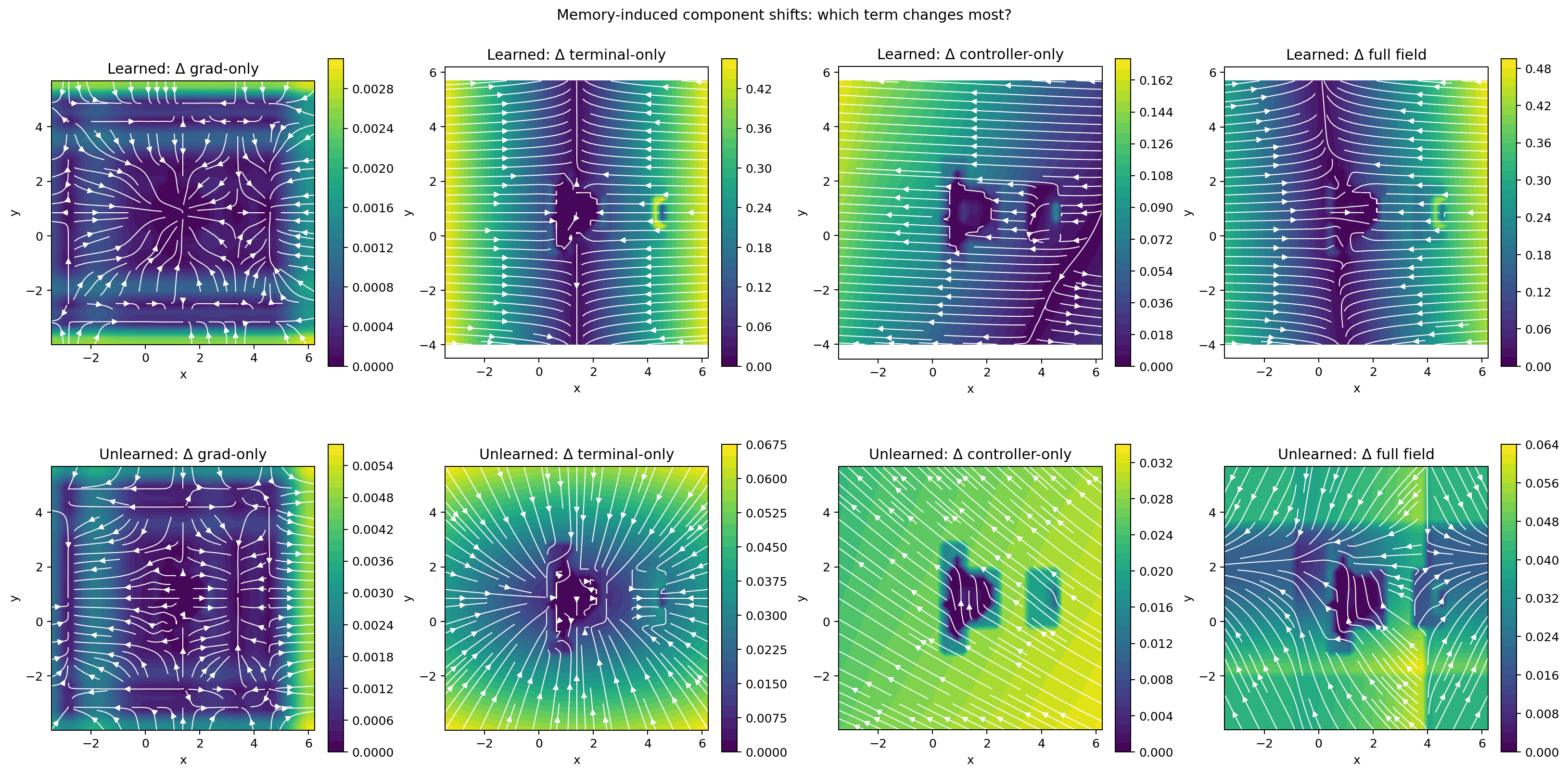}
    \caption{2D descent field decomposition for Levy functional. First column is the comparison regarding the learned velocity field versus oracle gradient field. Second and third columns record force field induced by subgoal and port controller respectively and fourth columns is the addition of second and third column.}
    \label{fig:2d:example:vector:field:display}
\end{figure}

This subsection expands the compact ablation summary in
Section~\ref{subsec:main_ablation_summary}.  The main text reports two
compressed views: Table~\ref{tab:ackley_oracle_ablation_compact_transposed} studies the robustness under different test-time oracle input. Here we report the corresponding method-level and oracle-level details. Table~\ref{tab:ackley_memory_controller_ablation_main} isolates the effect of enabling the local interconnected controller.  Table~\ref{tab:ackley_budget_particle_ablation_full} report the full comparison between performance evaluated under different task/particle/time-stepping budgets. Finally, Table~\ref{tab:ackley_oracle_ablation_full} reports the oracle-input ablation using the same checkpoint trained with first-order gradients.  The compact version in Table~\ref{tab:ackley_oracle_ablation_compact_transposed}
shows that SHAPE remains functional when the clean gradient channel is replaced at test time by stochastic gradients or zeroth-order finite-difference estimates. Under stochastic oracle inputs, SHAPE retains the best final gap, best-seen gap, final hit rate, and trajectory hit rate among the tested methods.  Under zeroth-order inputs, SHAPE still gives the best terminal gap and final hit rate, while NAG obtains the best best-seen gap and trajectory hit rate.  This distinction is useful: it suggests that the trained port-Hamiltonian controller
is robust to oracle perturbations, but zeroth-order exploration can sometimes favor simpler momentum-style baselines for transient basin hits.  Overall, the appendix tables refine the main conclusion rather than replacing it: SHAPE's most reliable advantage is fixed-budget navigational performance measured by best-seen values and hit rates, while terminal gaps and oracle-specific transient hits remain important complementary diagnostics. We also provide an example of vector field display in Figure~\ref{fig:2d:example:vector:field:display} and additional comparison with

\begin{longtable}{@{}lrrrrrr@{}}
\caption{Full method-level ablation results for the Ackley first-order benchmark. Results are grouped by evaluation setting $(N_{\rm task},N_{\rm part},B)$; the learned optimizer is reported as SHAPE and placed first in each group. Bold indicates the best value within a setting, and underline indicates the second-best value. Lower values are better for distances/gaps, while higher values are better for hit rates and minima counts.}\label{tab:ackley_budget_particle_ablation_full}\\
\toprule
Method & Final dist $\downarrow$ & Final gap $\downarrow$ & Best gap $\downarrow$ & Hit-final $\uparrow$ & Hit-traj $\uparrow$ & Minima $\uparrow$ \\
\midrule
\endfirsthead
\toprule
Method & Final dist $\downarrow$ & Final gap $\downarrow$ & Best gap $\downarrow$ & Hit-final $\uparrow$ & Hit-traj $\uparrow$ & Minima $\uparrow$ \\
\midrule
\endhead
\midrule
\multicolumn{7}{r}{\emph{continued on next page}}\\
\endfoot
\bottomrule
\endlastfoot
\multicolumn{7}{@{}l}{\textbf{Setting S1:} $N_{\rm task}=16$, $N_{\rm part}=128$, $B=256$, $N_{\rm part}B=32768$}\\
\textsc{SHAPE} & \best{1.281} & \best{2.997} & \best{$2.2\times 10^{-4}$} & \best{38.5\%} & \best{40.0\%} & \second{57} \\
adam & 3.850 & 7.907 & 4.592 & 0.0\% & 0.0\% & 47 \\
gd & 3.874 & 7.947 & 4.890 & 0.0\% & 0.0\% & 46 \\
lionk & \second{2.043} & \second{5.910} & $3.2\times 10^{-2}$ & \second{1.6\%} & \second{2.9\%} & \best{60} \\
momentum & 3.819 & 7.840 & $1.6\times 10^{-2}$ & 0.4\% & 0.8\% & 55 \\
nag & 3.830 & 7.860 & \second{$3.1\times 10^{-4}$} & 0.6\% & 0.6\% & 56 \\
rmsprop & 3.849 & 7.912 & 4.592 & 0.0\% & 0.0\% & 47 \\
shampoo & 3.880 & 7.959 & 5.130 & 0.0\% & 0.0\% & 45 \\
soap & 3.878 & 7.954 & 4.890 & 0.0\% & 0.0\% & 46 \\
sophia & 4.026 & 9.545 & 6.375 & 0.0\% & 0.0\% & 23 \\
\midrule
\multicolumn{7}{@{}l}{\textbf{Setting S2:} $N_{\rm task}=32$, $N_{\rm part}=128$, $B=256$, $N_{\rm part}B=32768$}\\
\textsc{SHAPE} & \best{1.054} & \best{2.321} & \second{$1.4\times 10^{-4}$} & \best{42.6\%} & \best{45.3\%} & \best{70} \\
adam & 3.979 & 7.492 & 4.623 & 0.0\% & 0.0\% & 58 \\
gd & 3.996 & 7.714 & 4.628 & 0.0\% & 0.0\% & 57 \\
lionk & \second{2.057} & \second{5.901} & 0.542 & 0.0\% & 0.0\% & \second{69} \\
momentum & 3.919 & 7.385 & $1.7\times 10^{-2}$ & 0.2\% & 0.4\% & 68 \\
nag & 3.894 & 7.459 & \best{$1.9\times 10^{-6}$} & \second{0.4\%} & \second{0.6\%} & \second{69} \\
rmsprop & 3.979 & 7.499 & 4.623 & 0.0\% & 0.0\% & 58 \\
shampoo & 4.013 & 7.541 & 4.623 & 0.0\% & 0.0\% & 57 \\
soap & 4.000 & 7.524 & 4.623 & 0.0\% & 0.0\% & 57 \\
sophia & 4.144 & 9.156 & 5.996 & 0.0\% & 0.0\% & 32 \\
\midrule
\multicolumn{7}{@{}l}{\textbf{Setting S3:} $N_{\rm task}=64$, $N_{\rm part}=128$, $B=256$, $N_{\rm part}B=32768$}\\
\textsc{SHAPE} & \best{0.621} & \best{1.645} & \second{$1.3\times 10^{-3}$} & \best{56.1\%} & \best{63.7\%} & \second{105} \\
adam & 4.544 & 9.082 & 6.053 & 0.0\% & 0.0\% & 85 \\
gd & 4.553 & 9.737 & 6.225 & 0.0\% & 0.0\% & 84 \\
lionk & \second{2.123} & 6.623 & 0.519 & 0.0\% & 0.2\% & \best{108} \\
momentum & 4.257 & 8.578 & $4.3\times 10^{-2}$ & 0.8\% & 1.4\% & \second{105} \\
nag & 2.480 & \second{5.808} & \best{$4.4\times 10^{-4}$} & \second{5.3\%} & \second{46.9\%} & \best{108} \\
rmsprop & 4.543 & 9.093 & 6.053 & 0.0\% & 0.0\% & 85 \\
shampoo & 4.587 & 9.150 & 6.053 & 0.0\% & 0.0\% & 85 \\
soap & 4.573 & 9.128 & 6.053 & 0.0\% & 0.0\% & 85 \\
sophia & 4.718 & 10.741 & 7.349 & 0.0\% & 0.0\% & 41 \\
\midrule
\multicolumn{7}{@{}l}{\textbf{Setting S4:} $N_{\rm task}=64$, $N_{\rm part}=64$, $B=512$, $N_{\rm part}B=32768$}\\
\textsc{SHAPE} & \best{0.871} & \best{2.105} & \best{$3.6\times 10^{-5}$} & \best{42.0\%} & \best{52.1\%} & \second{71} \\
adam & 3.627 & 7.441 & 4.056 & 0.0\% & 0.0\% & 60 \\
gd & 3.645 & 7.889 & 4.142 & 0.0\% & 0.0\% & 60 \\
lionk & \second{1.879} & \second{5.704} & $2.5\times 10^{-2}$ & 1.2\% & \second{3.1\%} & \best{73} \\
momentum & 3.384 & 6.959 & $1.4\times 10^{-3}$ & \second{2.0\%} & \second{3.1\%} & \best{73} \\
nag & 3.395 & 7.185 & \second{$4.2\times 10^{-5}$} & 0.6\% & 2.0\% & \best{73} \\
rmsprop & 3.627 & 7.449 & 4.056 & 0.0\% & 0.0\% & 60 \\
shampoo & 3.677 & 7.523 & 4.678 & 0.0\% & 0.0\% & 60 \\
soap & 3.652 & 7.482 & 4.056 & 0.0\% & 0.0\% & 61 \\
sophia & 3.825 & 9.517 & 5.781 & 0.0\% & 0.0\% & 33 \\
\midrule
\multicolumn{7}{@{}l}{\textbf{Setting S5:} $N_{\rm task}=64$, $N_{\rm part}=32$, $B=1024$, $N_{\rm part}B=32768$}\\
\textsc{SHAPE} & \best{0.667} & \best{1.861} & \best{$9.5\times 10^{-6}$} & \best{53.1\%} & \best{62.5\%} & \best{78} \\
adam & 4.204 & 9.310 & 5.734 & 0.0\% & 0.0\% & 59 \\
gd & 4.202 & 9.984 & 5.857 & 0.0\% & 0.0\% & 59 \\
lionk & 1.919 & 6.821 & $1.7\times 10^{-2}$ & 0.0\% & 0.8\% & \best{78} \\
momentum & 3.799 & 8.497 & $5.7\times 10^{-3}$ & 2.3\% & 3.7\% & \second{77} \\
nag & \second{1.796} & \second{4.927} & \second{$5.9\times 10^{-5}$} & \second{8.4\%} & \second{61.1\%} & \best{78} \\
rmsprop & 4.203 & 9.320 & 5.734 & 0.0\% & 0.0\% & 59 \\
shampoo & 4.234 & 9.358 & 5.734 & 0.0\% & 0.0\% & 59 \\
soap & 4.222 & 9.339 & 5.734 & 0.0\% & 0.0\% & 59 \\
sophia & 4.389 & 11.192 & 7.361 & 0.0\% & 0.0\% & 26 \\
\midrule
\multicolumn{7}{@{}l}{\textbf{Setting S6:} $N_{\rm task}=64$, $N_{\rm part}=16$, $B=2048$, $N_{\rm part}B=32768$}\\
\textsc{SHAPE} & \best{0.456} & \best{1.651} & \best{$<10^{-10}$} & \best{66.0\%} & \second{72.7\%} & \second{70} \\
adam & 3.725 & 10.965 & 6.909 & 0.0\% & 0.0\% & 58 \\
gd & 3.728 & 11.612 & 6.910 & 0.0\% & 0.0\% & 58 \\
lionk & 1.649 & 7.346 & 0.298 & 0.8\% & 2.0\% & \best{73} \\
momentum & 3.059 & 9.070 & \second{$4.4\times 10^{-4}$} & 1.2\% & 11.7\% & \best{73} \\
nag & \second{1.274} & \second{4.759} & \best{$<10^{-10}$} & \second{10.0\%} & \best{74.0\%} & \best{73} \\
rmsprop & 3.723 & 10.969 & 6.910 & 0.0\% & 0.0\% & 59 \\
shampoo & 3.757 & 11.023 & 6.909 & 0.0\% & 0.0\% & 58 \\
soap & 3.738 & 10.987 & 6.909 & 0.0\% & 0.0\% & 59 \\
sophia & 3.942 & 13.167 & 8.947 & 0.0\% & 0.0\% & 27 \\
\end{longtable}

\begin{longtable}{lrrrrrr}
\caption{Full gradient-oracle ablation on the Ackley benchmark using the same SHAPE checkpoint trained with first-order oracle inputs.  Within each oracle group, SHAPE is listed first and best/second-best entries are shown in \textbf{bold}/\underline{underline}. Lower gaps are better; higher hit rates and minima counts are better.}\label{tab:ackley_oracle_ablation_full}\\
\toprule
Method & Final gap $\downarrow$ & Best gap $\downarrow$ & Final hit $\uparrow$ & Traj. hit $\uparrow$ & Traj. minima $\uparrow$ & Minima hits $\uparrow$ \\
\midrule
\endfirsthead
\caption[]{Full gradient-oracle ablation on the Ackley benchmark (continued).}\\
\toprule
Method & Final gap $\downarrow$ & Best gap $\downarrow$ & Final hit $\uparrow$ & Traj. hit $\uparrow$ & Traj. minima $\uparrow$ & Minima hits $\uparrow$ \\
\midrule
\endhead
\midrule
\multicolumn{7}{r}{Continued on next page}\\
\endfoot
\bottomrule
\endlastfoot
\midrule
\multicolumn{7}{l}{\textbf{First-order test oracle}}\\
\textbf{SHAPE} & \textbf{2.105} & \textbf{3.6e-5} & \textbf{42.0\%} & \textbf{52.1\%} & \underline{71} & 69386 \\
GD & 7.889 & 4.142 & 0.0\% & 0.0\% & 60 & 47180 \\
Momentum & 6.959 & 0.001 & \underline{2.0\%} & \underline{3.1\%} & \textbf{73} & \underline{111369} \\
NAG & 7.185 & \underline{4.2e-5} & 0.6\% & 2.0\% & \textbf{73} & 102997 \\
RMSProp & 7.449 & 4.056 & 0.0\% & 0.0\% & 60 & \textbf{115169} \\
Adam & 7.441 & 4.056 & 0.0\% & 0.0\% & 60 & 110169 \\
LionK & \underline{5.704} & 0.025 & 1.2\% & \underline{3.1\%} & \textbf{73} & 8596 \\
Shampoo & 7.523 & 4.678 & 0.0\% & 0.0\% & 60 & 109682 \\
SOAP & 7.482 & 4.056 & 0.0\% & 0.0\% & 61 & 101737 \\
Sophia & 9.517 & 5.781 & 0.0\% & 0.0\% & 33 & 7338 \\
\midrule
\multicolumn{7}{l}{\textbf{Stochastic test oracle}}\\
\textbf{SHAPE} & \textbf{1.709} & \textbf{5.7e-6} & \textbf{59.4\%} & \textbf{66.6\%} & \underline{64} & 64322 \\
GD & 8.545 & 5.098 & 0.0\% & 0.0\% & 52 & 24955 \\
Momentum & 7.410 & 0.012 & \underline{1.4\%} & 2.1\% & \textbf{65} & \underline{98591} \\
NAG & 7.572 & \underline{3.2e-4} & 1.0\% & \underline{2.7\%} & \textbf{65} & 73840 \\
RMSProp & 7.994 & 5.025 & 0.0\% & 0.0\% & 53 & \textbf{100124} \\
Adam & 7.984 & 5.025 & 0.0\% & 0.0\% & 53 & 95729 \\
LionK & \underline{6.030} & 0.194 & 0.8\% & 1.6\% & \textbf{65} & 7730 \\
Shampoo & 8.046 & 5.025 & 0.0\% & 0.0\% & 52 & 95177 \\
SOAP & 8.018 & 5.025 & 0.0\% & 0.0\% & 53 & 88025 \\
Sophia & 10.140 & 6.497 & 0.0\% & 0.0\% & 26 & 6679 \\
\midrule
\multicolumn{7}{l}{\textbf{Zeroth-order test oracle}}\\
\textbf{SHAPE} & \textbf{1.574} & \underline{2.5e-4} & \textbf{65.8\%} & \underline{72.7\%} & 81 & 62226 \\
GD & 11.008 & 6.401 & 0.0\% & 0.0\% & 72 & 32550 \\
Momentum & 8.910 & 0.013 & 0.0\% & 5.7\% & 82 & 109147 \\
NAG & \underline{3.800} & \textbf{1.1e-5} & \underline{7.2\%} & \textbf{90.4\%} & \textbf{93} & 33383 \\
RMSProp & 10.203 & 6.392 & 0.0\% & 0.0\% & 73 & \textbf{121749} \\
Adam & 10.191 & 6.392 & 0.0\% & 0.0\% & 73 & 117174 \\
LionK & 7.319 & 0.044 & 0.0\% & 1.2\% & \underline{83} & 9722 \\
Shampoo & 10.245 & 6.392 & 0.0\% & 0.0\% & 72 & \underline{117614} \\
SOAP & 10.226 & 6.392 & 0.0\% & 0.0\% & 72 & 109203 \\
Sophia & 12.321 & 7.922 & 0.0\% & 0.0\% & 37 & 12141 \\
\end{longtable}

\begin{figure}[htbp]
    \centering
    \includegraphics[width=0.55\linewidth]{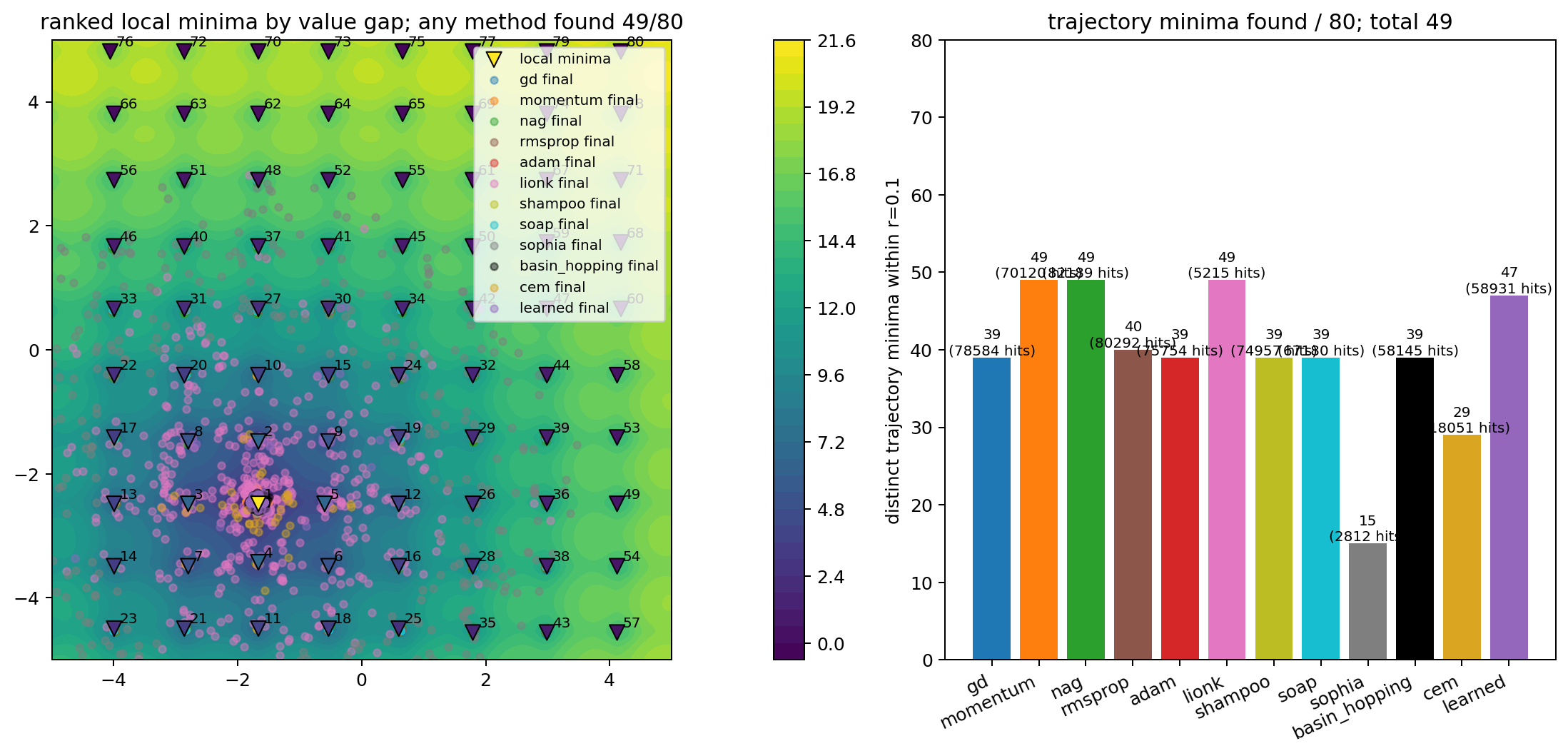}
     \includegraphics[width=0.4\linewidth]{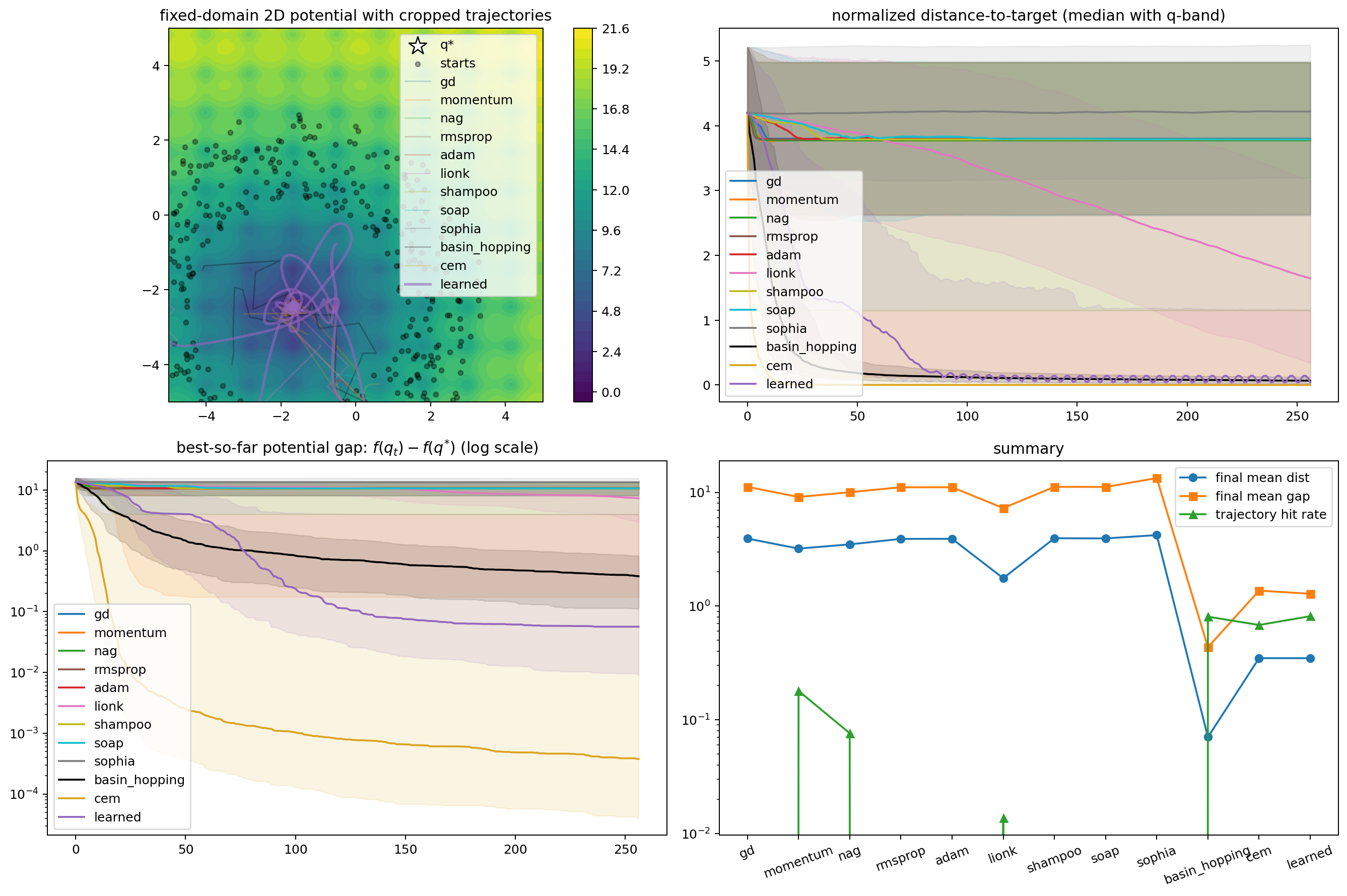}
    \caption{Additional comparison against some search method including basin-hopping and CEM. Notably, CEM has a population size of $40$ therefore it costs $20\times$ more compared with SHAPE, which consumes a single gradient call per query time.}
    \label{fig:ackley:with:search:algo}
\end{figure}
\section{Theoretical analysis and proofs}
\label{app:shape_theory}

This appendix section contains the assumptions, theorem statements, and proofs
supporting the main theoretical summary in Section~\ref{subsec:main_theory_summary}.
Implementation choices are recorded separately in Appendix~\ref{app:shape_implementation}.
The analysis has two layers: frozen-stage estimates for a fixed local
port-Hamiltonian system, and hybrid estimates for the event-triggered memory
mechanism that changes the shaped landscape between stages.

\paragraph{Notation synchronization with the main text.}
The appendix uses $f_\tau$ only to denote a sampled task objective from a task family; for a fixed test problem this is the same object denoted $f$ in the main text.  The symbol $\bar q_s$ denotes the stage anchor or local target, while $g(q)$ or $\widetilde g_{s,n}$ denotes a gradient/force observation.  The frozen shaped potential is written
\[
    U_{\tau,s}(q) = f_\tau(q)+U_s^{\rm shp}(q),
    \qquad
    U_s^{\rm shp}(q)
    =U_{\rm goal,s}(q;\bar q_s)+U_{\rm mem,s}(q;m_s^0)+V_{\rm bar,s}(q;m_s^0,\ell_s).
\]
Here $U_{\rm goal,s}$ is the local anchoring term, $U_{\rm mem,s}$ is the memory-induced reshaping term, and $V_{\rm bar,s}$ is a mode-dependent exclusion/barrier term.  The actual port input is $u_s^{\rm port}=u_s^{\rm shp}-K_s^dy_s$; in proofs where only one input appears, $u_s$ is shorthand for $u_s^{\rm port}$.

\subsection{Stagewise and hybrid convergence guarantees}
\label{subsec:dp_stagewise_theory}

The key distinction is between a \emph{frozen local stage}, in which the planner output and memory snapshot are held fixed and one studies local contraction of the induced second-order system, and the \emph{hybrid inter-stage mechanism}, in which the event-triggered update map $\Pi_\eta$ changes the shaped landscape and may expose a lower-value basin.

\paragraph{Frozen stage model.}
Fix a stage $s$ and freeze the planner and memory variables $(\bar q_s,m_s^0,\mu_s)$. Define the frozen stage potential
\begin{equation}
\label{eq:dp_stage_potential}
U_{\tau,s}(q)
:=
f_\tau(q)+U_s^{\rm shp}(q),
\qquad
U_s^{\rm shp}(q)
:=
U_{\rm goal,s}(q;\bar q_s)+U_{\rm mem,s}(q;m_s^0)+V_{\rm bar,s}(q;m_s^0,\ell_s).
\end{equation}
For the local theory we study the frozen stage dynamics
\begin{equation}
\label{eq:dp_frozen_stage_dynamics}
\dot q = M_s^{-1}p,
\qquad
\dot p = -\nabla U_{\tau,s}(q)+\Omega_s M_s^{-1}p-D_s M_s^{-1}p+B_su_s^{\mathrm{port}},
\end{equation}
with collocated velocity output
\begin{equation}
\label{eq:dp_frozen_stage_output}
y_s=B_s^\top M_s^{-1}p.
\end{equation}
Here $M_s\in\mathbb R^{d\times d}$ is symmetric positive definite, $\Omega_s^\top=-\Omega_s$ is the frozen skew correction, and $D_s\succeq 0$ is the frozen damping block. In the implementation these quantities are produced stagewise by the learned planner--controller pair, but the local proof treats them as frozen coefficients on one contraction neighborhood.

Let $q_s^\star$ be a local minimizer of $U_{\tau,s}$ and set $e=q-q_s^\star$, $v=M_s^{-1}p$. We use the modified Lyapunov function
\begin{equation}
\label{eq:dp_modified_lyapunov}
\mathcal V_s(e,p)
:=
U_{\tau,s}(q)-U_{\tau,s}(q_s^\star)
+\frac12 p^\top M_s^{-1}p
+\varepsilon_s e^\top p,
\end{equation}
where $\varepsilon_s>0$ is chosen sufficiently small.

\paragraph{Local stage assumptions.}
For a fixed stage $s$, assume:
\begin{enumerate}
    \item[(A1)] There exists a neighborhood $\mathcal U_s^q$ of $q_s^\star$ on which $U_{\tau,s}$ is $C^2$, $\mu_s$-strongly convex, and $L_s$-smooth.
    \item[(A2)] There exist constants $0<m_{-,s}\le m_{+,s}<\infty$, $0<d_{0,s}\le \bar d_s<\infty$, and $\bar\omega_s,\bar b_s<\infty$ such that
    \begin{equation}
    \label{eq:dp_stage_operator_bounds}
    m_{-,s}I\preceq M_s\preceq m_{+,s}I,
    \qquad
    d_{0,s}I\preceq D_s\preceq \bar d_s I,
    \qquad
    \|\Omega_s\|\le \bar\omega_s,
    \qquad
    \|B_s\|\le \bar b_s.
    \end{equation}
    \item[(A3)] The stage port input is bounded by
    \begin{equation}
    \label{eq:dp_port_power_bound}
    \|u_s^{\mathrm{port}}(t)\|^2\le \beta_s
    \qquad\text{for all }t\in[0,T_s].
    \end{equation}
    \item[(A4)] The numerical stage update $z_{s,n+1}=\Phi^{h_s}_{\tau,\psi}(z_{s,n};\bar q_s,m_s^0,\mu_s)$, with $z_{s,n}=(q_{s,n},p_{s,n})$, satisfies the one-step Lyapunov defect inequality
    \begin{equation}
    \label{eq:dp_discrete_stage_energy_ineq}
    \mathcal V_s(z_{s,n+1})
    \le
    \mathcal V_s(z_{s,n})
    -h_sc_s\mathcal V_s(z_{s,n})
    +C_{s,1}h_s^2
    +h_s C_{s,2}\beta_{s,n}
    +\delta^{\mathrm{proj}}_{s,n},
    \end{equation}
    where $c_s,C_{s,1},C_{s,2}>0$, $\beta_{s,n}\ge0$ bounds the discrete port/input contribution, and $\delta^{\mathrm{proj}}_{s,n}$ records clipping or projection defects.
\end{enumerate}

\begin{theorem}[Local hypocoercive convergence of a frozen stage]
\label{lem:dp_canonical_frozen_stage_rate}
Assume \emph{(A1)}--\emph{(A3)}. Then there exist constants $\varepsilon_s,c_{1,s},c_{2,s},c_s,C_s>0$ such that the modified Lyapunov function \eqref{eq:dp_modified_lyapunov} satisfies
\begin{equation}
\label{eq:dp_modified_lyapunov_equiv}
c_{1,s}\bigl(\|e\|^2+\|p\|^2\bigr)
\le
\mathcal V_s(e,p)
\le
c_{2,s}\bigl(\|e\|^2+\|p\|^2\bigr)
\qquad\text{on }\mathcal U_s^q\times\mathbb R^d,
\end{equation}
and, along every trajectory of \eqref{eq:dp_frozen_stage_dynamics},
\begin{equation}
\label{eq:dp_modified_lyapunov_dissipation}
\dot{\mathcal V}_s(t)
\le
-c_s\mathcal V_s(t)+C_s\beta_s.
\end{equation}
Consequently,
\begin{equation}
\label{eq:dp_continuous_exp_decay}
\mathcal V_s(t)
\le
\mathrm e^{-c_s t}\mathcal V_s(0)+\frac{C_s}{c_s}\beta_s.
\end{equation}
In particular, if $\beta_s=0$, then
\begin{equation}
\label{eq:dp_continuous_state_decay}
\|q(t)-q_s^\star\|^2+\|p(t)\|^2
\le
\frac{c_{2,s}}{c_{1,s}}\,\mathrm e^{-c_s t}
\bigl(\|q(0)-q_s^\star\|^2+\|p(0)\|^2\bigr).
\end{equation}
Thus the frozen stage converges exponentially to its local equilibrium when the port input vanishes, and to an $\mathcal O(\beta_s)$ neighborhood otherwise.
\end{theorem}

\begin{remark}
The theorem above is the correct replacement for the earlier damping-domination statement. In the implemented architecture, the dissipation is momentum-channel only; it does not directly control $\|\nabla U_{\tau,s}(q)\|^2$. Exponential local convergence is recovered by the cross term in \eqref{eq:dp_modified_lyapunov}, i.e. by a standard hypocoercive argument rather than by a false coercivity identity on the raw Hamiltonian derivative.
\end{remark}

\begin{theorem}[Discrete frozen-stage contraction with numerical defects]
\label{lem:dp_discrete_frozen_stage_contraction}
Assume \emph{(A4)} and $0<c_sh_s<1$. Then for every $n\ge0$,
\begin{equation}
\label{eq:dp_discrete_contraction}
\mathcal V_s(z_{s,n})
\le
(1-c_sh_s)^n\mathcal V_s(z_{s,0})
+\frac{C_{s,1}}{c_s}h_s
+\frac{C_{s,2}}{c_s}\sup_{0\le j<n}\beta_{s,j}
+
\sum_{j=0}^{n-1}(1-c_sh_s)^{n-1-j}\,\delta^{\mathrm{proj}}_{s,j,+},
\end{equation}
where $\delta^{\mathrm{proj}}_{s,j,+}:=\max\{\delta^{\mathrm{proj}}_{s,j},0\}$. In particular, if all projection defects are nonpositive, then the explicit clipping/projection step is energy-nonincreasing and the last term disappears.
\end{theorem}

\begin{remark}
The discrete theorem is intentionally stated as a \emph{defect theorem}. It does not claim that the implemented explicit update automatically inherits a perfect PH energy inequality. Instead it records exactly the three sources of discrepancy that matter in practice: local truncation error, bounded port/input work, and clipping/projection defects.
\end{remark}

\begin{proposition}[Frozen-stage approximate equilibration]
\label{prop:dp_frozen_stage_conv}
Assume \emph{(A1)}--\emph{(A4)} and $0<c_sh_s<1$. Let $z_s^+=(q_s^+,p_s^+)$ denote the terminal state of stage $s$. Define the accumulated numerical defect
\begin{equation}
\label{eq:dp_stage_total_defect}
\mathfrak e_s
:=
\frac{C_{s,1}}{c_s}h_s
+
\frac{C_{s,2}}{c_s}\sup_{0\le n<N_s}\beta_{s,n}
+
\sum_{n=0}^{N_s-1}(1-c_sh_s)^{N_s-1-n}\,\delta^{\mathrm{proj}}_{s,n,+},
\end{equation}
and the total terminal Lyapunov error
\begin{equation}
\label{eq:dp_stage_total_error}
\mathfrak{E}_s
:=(1-c_sh_s)^{N_s}\mathcal V_s(z_{s,0})+\mathfrak{e}_s .
\end{equation}
If the stage terminates at residual level
\begin{equation}
\|\nabla_q U_{\tau,s}(q_s^+)\|+\|p_s^+\|\le \eta_s,
\label{eq:dp_stage_stationarity}
\end{equation}
then, under the usual local nondegeneracy assumptions on $U_{\tau,s}$ near $q_s^\star$, there exists $C_{\mathrm{eq},s}>0$ such that
\begin{equation}
\|q_s^+-q_s^\star\|+\|p_s^+\|
\le C_{\mathrm{eq},s}\bigl(\eta_s+\sqrt{\mathfrak{E}_s}\bigr).
\label{eq:dp_stage_terminal_neighborhood}
\end{equation}
Thus a residual-triggered stage gives the usual $\mathcal O(\eta_s)$ local stationary bound, whereas a budget-limited stage gives the energy bound $\mathcal O(\sqrt{\mathfrak E_s})$. The square root appears because $\mathcal V_s$ is locally equivalent to a squared norm.
\end{proposition}

The preceding statements explain the behavior of a \emph{single frozen stage}. They do not yet explain why lower-potential equilibria may become reachable. This is the role of the inter-stage update map and the acceptance rule.

Let
\begin{equation}
\label{eq:dp_task_potential}
f_{\tau,\mu}(q):=f_\tau(q)+b_\mu(q)
\end{equation}
denote the constrained task potential, where $b_\mu$ is a fixed feasibility or barrier penalty associated with the task constraints.  In unconstrained experiments, $b_\mu\equiv0$. After stage $s$, the planner and persistent memory are updated according to
\begin{equation}
\label{eq:dp_memory_update_rule}
(m_{s+1}^0,\bar q_{s+1},\mu_{s+1})
=
\Pi_\eta(m_s^0,\mathcal S_s).
\end{equation}
Define the within-stage best candidate and the accepted iterate by
\begin{equation}
\label{eq:dp_stage_best_candidate}
\widetilde q_s
\in
\operatorname*{arg\,min}_{0\le n\le N_s} f_\tau(q_{s,n}),
\qquad q_{s,N_s}=q_s^+,
\end{equation}
\begin{equation}
\label{eq:dp_accepted_iterate}
\widehat q_{s+1}
\in
\arg\min\{f_\tau(\widehat q_s),f_\tau(\widetilde q_s)\}.
\end{equation}
This event-time definition is equivalent to the implementation's per-fast-step best-so-far update.
Define the stagewise shaping distortion
\begin{equation}
\label{eq:dp_shaping_distortion}
\Delta^{\tau}_{\psi,s}(q;\bar q_s,m_s^0,\mu_s)
:=
U_{\tau,s}(q)-f_\tau(q).
\end{equation}
Assume:
\begin{enumerate}
    \item[(C1)] The terminal stage residuals satisfy \eqref{eq:dp_stage_stationarity} with $\eta_s\to0$.
    \item[(C2)] The shaping distortion obeys
    \begin{equation}
    \label{eq:dp_shaping_distortion_bound}
    |\Delta^{\tau}_{\psi,s}(q;\bar q_s,m_s^0,\mu_s)|\le \rho_s
    \qquad\text{on the reachable neighborhood of stage }s,
    \end{equation}
    with $\rho_s\to0$.
    \item[(C3)] The accumulated stage defects are summable:
    \begin{equation}
    \label{eq:dp_summable_defects}
    \sum_{s=0}^\infty \sqrt{\mathfrak E_s} < \infty.
    \end{equation}
\end{enumerate}

\begin{proposition}[Hybrid memory-assisted strict improvement]
\label{prop:dp_hybrid_memory_improvement}
Suppose each stage satisfies Proposition~\ref{prop:dp_frozen_stage_conv} and \emph{(C1)}--\emph{(C3)} hold. Then:
\begin{enumerate}
    \item the accepted task values are nonincreasing,
    \begin{equation}
    \label{eq:dp_monotone_acceptance}
    f_\tau(\widehat q_{s+1})\le f_\tau(\widehat q_s),
    \qquad s\ge0,
    \end{equation}
    hence converge to a finite limit;
    \item if the update at the end of stage $s$ produces a next-stage frozen equilibrium $q_{s+1}^\star$ such that
    \begin{equation}
    \label{eq:dp_strictly_better_equilibrium}
    f_\tau(q_{s+1}^\star)\le f_\tau(\widehat q_{s+1})-\delta_{s+1},
    \qquad \delta_{s+1}>0,
    \end{equation}
    then after executing stage $s+1$ one has
    \begin{equation}
    \label{eq:dp_strict_improvement}
    f_\tau(\widehat q_{s+2})
    \le
    f_\tau(\widehat q_{s+1})
    -\delta_{s+1}
    +C_1\eta_{s+1}+C_2\rho_{s+1}+C_3\sqrt{\mathfrak E_{s+1}},
    \end{equation}
    where $C_1,C_2,C_3>0$ depend only on local regularity.
\end{enumerate}
\end{proposition}

\begin{proposition}[General nonconvex convergence under sufficient decrease and relative error]
\label{prop:dp_general_nonconvex}
Assume $f_\tau$ is bounded below and has bounded sublevel sets. Suppose that, for all sufficiently large $s$, the accepted sequence satisfies
\begin{align}
f_\tau(\widehat q_s)-f_\tau(\widehat q_{s+1})
&\ge a\,\|\widehat q_{s+1}-\widehat q_s\|^2-\varepsilon_s,
\label{eq:dp_sufficient_decrease}
\\
\operatorname{dist}\bigl(0,\partial f_\tau(\widehat q_{s+1})\bigr)
&\le b\,\|\widehat q_{s+1}-\widehat q_s\|+\varepsilon_s,
\label{eq:dp_relative_error}
\end{align}
for some $a,b>0$ and some nonnegative summable error sequence $(\varepsilon_s)$. Then:
\begin{enumerate}
    \item the accepted values $f_\tau(\widehat q_s)$ converge;
    \item every accumulation point $\bar q$ of $\{\widehat q_s\}$ is critical for $f_\tau$, i.e.
    \begin{equation}
    \label{eq:dp_critical_cluster_point}
    0\in \partial f_\tau(\bar q);
    \end{equation}
    \item if, in addition, $f_\tau$ satisfies the Kurdyka--\L{}ojasiewicz property on the relevant sublevel set, then the full accepted sequence converges to a single critical point.
\end{enumerate}
\end{proposition}

\paragraph{Stochastic oracle assumption.}
At stage \(s\), the observed local force is \(\mathcal F_{s,n}\)-adapted and satisfies
\[
\widetilde g_{s,n}
=
\nabla_q U_{\tau,s}(q_{s,n})+\xi_{s,n},
\qquad
\mathbb E[\xi_{s,n}\mid\mathcal F_{s,n}]=0,
\qquad
\mathbb E[\xi_{s,n}\xi_{s,n}^{\top}\mid\mathcal F_{s,n}]
\preceq
\Xi_{s,n},
\]
where \(\Xi_{s,n}\in\mathbb S_+^d\) is the conditional oracle-noise covariance.
For the fixed metric weight \(W_\Xi\succeq0\) used in the one-step Lyapunov expansion, assume the weighted second moment is finite, so that
\[
    \mathbb E[\|\xi_{s,n}\|_{W_\Xi}^2\mid\mathcal F_{s,n}]
    \le
    \operatorname{tr}(W_\Xi\Xi_{s,n}),
    \qquad
    \|\xi\|_{W_\Xi}^2:=\xi^\top W_\Xi\xi .
\]

\begin{proposition}[Stochastic-oracle defect bound]
\label{prop:dp_stochastic_oracle_defect}
Suppose the stochastic oracle assumption above holds and the one-step Lyapunov
expansion satisfies the same deterministic second-order remainder estimate as
in \eqref{eq:dp_discrete_stage_energy_ineq}.  Then the discrete frozen-stage
contraction estimate extends to
\begin{equation}
\begin{aligned}   
\label{eq:dp_stochastic_discrete_ineq}
\mathbb E\!&\left[\mathcal V_s(z_{s,n+1})\mid \mathcal F_{s,n}\right]
\le\\
&(1-c_sh_s)\mathcal V_s(z_{s,n})
+
C_{s,1}h_s^2
+
\widetilde C_{s,1}h_s^2\,\operatorname{tr}(W_\Xi\Xi_{s,n})
+
h_s\widetilde C_{s,2}\beta_{s,n}
+
\mathbb E\!\left[\delta^{\mathrm{proj}}_{s,n}\mid \mathcal F_{s,n}\right],
\end{aligned}
\end{equation}
for suitable constants \(\widetilde C_{s,1},\widetilde C_{s,2}>0\).  Thus the
deterministic frozen-stage contraction is preserved up to the deterministic
truncation defect \(C_{s,1}h_s^2\) and an additional one-step stochastic defect
of order \(h_s^2\operatorname{tr}(W_\Xi\Xi_{s,n})\).  After summing the geometric
recursion over one stage, this stochastic one-step term contributes a steady
defect floor of order \(h_s\sup_n\operatorname{tr}(W_\Xi\Xi_{s,n})\).
\end{proposition}

In practice, the covariance \(\Xi_{s,n}\) is typically unknown, exactly as in
SGHMC-type constructions where the stochastic-oracle noise contribution is not
directly available.  In the stochastic extension, one therefore replaces
\(\Xi_{s,n}\) by an online filtered or learned estimate \(\widehat\Xi_{s,n}\).

\begin{remark}[Dimension dependence]
The statements above are not dimension-free in a worst-case complexity sense, but the ambient dimension $d$ does not appear explicitly in the logical form of the theorems. Dimension enters only through local geometric and numerical constants: $(m_{-,s},m_{+,s},\mu_s,L_s,d_{0,s},\bar d_s,\bar\omega_s,\bar b_s)$ and the defect constants $(C_{s,1},C_{s,2})$. In particular, high ambient dimension can deteriorate these constants and shrink the strict-improvement margin, but it does not change the proof architecture itself.
\end{remark}

\subsection{Proof of the local hypocoercive frozen-stage theorem}
\label{app:dp_hypocoercive_stage}

Fix a stage $s$ and write the frozen local system as
\begin{equation}
\label{eq:app_frozen_stage_repeat}
\dot q=M_s^{-1}p,
\qquad
\dot p=-\nabla U_{\tau,s}(q)+\Omega_s M_s^{-1}p-D_s M_s^{-1}p+B_su_s^{\mathrm{port}}.
\end{equation}
Let $q_s^\star$ be the local minimizer from Assumption~(A1), set $e=q-q_s^\star$, and let $v=M_s^{-1}p$. Define the raw stage Hamiltonian gap
\[
H_s(e,p):=U_{\tau,s}(q)-U_{\tau,s}(q_s^\star)+\frac12 p^\top M_s^{-1}p.
\]
Then the modified Lyapunov function is
\[
\mathcal V_s(e,p)=H_s(e,p)+\varepsilon_s e^\top p.
\]

\paragraph{Step 1: equivalence with $\|e\|^2+\|p\|^2$.}
By $\mu_s$-strong convexity and $L_s$-smoothness of $U_{\tau,s}$ on $\mathcal U_s^q$,
\[
\frac{\mu_s}{2}\|e\|^2
\le U_{\tau,s}(q)-U_{\tau,s}(q_s^\star)
\le \frac{L_s}{2}\|e\|^2.
\]
By the bounds on $M_s$,
\[
\frac{1}{2m_{+,s}}\|p\|^2
\le \frac12 p^\top M_s^{-1}p
\le \frac{1}{2m_{-,s}}\|p\|^2.
\]
Finally, Young's inequality gives
\[
|\varepsilon_s e^\top p|
\le \frac{\varepsilon_s\gamma_s}{2}\|e\|^2+\frac{\varepsilon_s}{2\gamma_s}\|p\|^2
\]
for any $\gamma_s>0$. Choosing first $\gamma_s$ and then $\varepsilon_s>0$ sufficiently small yields constants $c_{1,s},c_{2,s}>0$ such that
\begin{equation}
\label{eq:app_modified_lyapunov_equiv}
c_{1,s}(\|e\|^2+\|p\|^2)
\le \mathcal V_s(e,p)
\le c_{2,s}(\|e\|^2+\|p\|^2).
\end{equation}
This proves \eqref{eq:dp_modified_lyapunov_equiv}.

\paragraph{Step 2: derivative of the raw Hamiltonian gap.}
Differentiating $H_s$ along \eqref{eq:app_frozen_stage_repeat} gives
\begin{align}
\dot H_s
&=\nabla U_{\tau,s}(q)^\top \dot q + v^\top \dot p
\nonumber\\
&=\nabla U_{\tau,s}(q)^\top v + v^\top\bigl(-\nabla U_{\tau,s}(q)+\Omega_s v-D_s v+B_su_s^{\mathrm{port}}\bigr)
\nonumber\\
&=-v^\top D_sv+v^\top B_su_s^{\mathrm{port}},
\label{eq:app_raw_hamiltonian_derivative}
\end{align}
because $v^\top \Omega_s v=0$ for every skew-symmetric $\Omega_s$.

Using $D_s\succeq d_{0,s}I$ and $\|B_s\|\le \bar b_s$, Young's inequality yields
\begin{equation}
\label{eq:app_raw_hamiltonian_derivative_bound}
\dot H_s
\le -\frac{d_{0,s}}{2}\|v\|^2 + \frac{\bar b_s^2}{2d_{0,s}}\|u_s^{\mathrm{port}}\|^2.
\end{equation}
By Assumption~(A3), this becomes
\begin{equation}
\label{eq:app_raw_hamiltonian_derivative_beta}
\dot H_s
\le -\frac{d_{0,s}}{2}\|v\|^2 + C_{u,s}\beta_s,
\qquad C_{u,s}:=\frac{\bar b_s^2}{2d_{0,s}}.
\end{equation}

\paragraph{Step 3: derivative of the cross term.}
Differentiating $e^\top p$ gives
\begin{align}
\frac{d}{dt}(e^\top p)
&=\dot e^\top p + e^\top \dot p
\nonumber\\
&=v^\top p - e^\top \nabla U_{\tau,s}(q) + e^\top \Omega_s v - e^\top D_s v + e^\top B_su_s^{\mathrm{port}}.
\label{eq:app_cross_term_derivative}
\end{align}
We estimate the right-hand side term by term.
First,
\[
v^\top p=p^\top M_s^{-1}p\le m_{-,s}^{-1}\|p\|^2.
\]
Second, by strong convexity,
\[
e^\top \nabla U_{\tau,s}(q)\ge \mu_s\|e\|^2.
\]
Third, using $\|\Omega_s\|\le \bar \omega_s$, $\|D_s\|\le \bar d_s$, and $\|B_s\|\le \bar b_s$,
\begin{align*}
|e^\top \Omega_s v|
&\le \bar \omega_s\|e\|\,\|v\|
\le \frac{\mu_s}{8}\|e\|^2 + \frac{2\bar \omega_s^2}{\mu_s}\|v\|^2,
\\
|e^\top D_s v|
&\le \bar d_s\|e\|\,\|v\|
\le \frac{\mu_s}{8}\|e\|^2 + \frac{2\bar d_s^2}{\mu_s}\|v\|^2,
\\
|e^\top B_su_s^{\mathrm{port}}|
&\le \bar b_s\|e\|\,\|u_s^{\mathrm{port}}\|
\le \frac{\mu_s}{8}\|e\|^2 + \frac{2\bar b_s^2}{\mu_s}\|u_s^{\mathrm{port}}\|^2.
\end{align*}
Substituting these bounds into \eqref{eq:app_cross_term_derivative} gives
\begin{equation}
\label{eq:app_cross_term_derivative_bound}
\frac{d}{dt}(e^\top p)
\le -\frac{5\mu_s}{8}\|e\|^2 + C_{p,s}\|p\|^2 + C_{v,s}\|v\|^2 + C_{B,s}\|u_s^{\mathrm{port}}\|^2,
\end{equation}
for constants $C_{p,s},C_{v,s},C_{B,s}>0$ depending only on the local bounds in Assumptions~(A1)--(A3).

\paragraph{Step 4: derivative of the modified Lyapunov function.}
Combining \eqref{eq:app_raw_hamiltonian_derivative_beta} and $\varepsilon_s$ times \eqref{eq:app_cross_term_derivative_bound}, and using the equivalence between $\|v\|^2$ and $\|p\|^2$ induced by $M_s$, we obtain
\[
\dot{\mathcal V}_s
\le -c_{e,s}\|e\|^2 - c_{p,s}\|p\|^2 + C_s\beta_s,
\]
provided $\varepsilon_s>0$ is chosen sufficiently small. Using \eqref{eq:app_modified_lyapunov_equiv} once more gives
\[
\dot{\mathcal V}_s\le -c_s\mathcal V_s + C_s\beta_s,
\]
which is exactly \eqref{eq:dp_modified_lyapunov_dissipation}. The comparison lemma then yields \eqref{eq:dp_continuous_exp_decay}, and \eqref{eq:dp_continuous_state_decay} follows from the norm equivalence \eqref{eq:app_modified_lyapunov_equiv}. This proves Theorem~\ref{lem:dp_canonical_frozen_stage_rate}.

\subsection{Proof of the discrete defect theorem}
\label{app:dp_discrete_defect}

The proof of Theorem~\ref{lem:dp_discrete_frozen_stage_contraction} is a direct iteration of the one-step bound \eqref{eq:dp_discrete_stage_energy_ineq}. Write
\[
\delta^{\mathrm{proj}}_{s,n}=\delta^{\mathrm{proj}}_{s,n,+}-\delta^{\mathrm{proj}}_{s,n,-},
\qquad
\delta^{\mathrm{proj}}_{s,n,\pm}\ge 0.
\]
Discarding the nonpositive part yields
\[
\mathcal V_s(z_{s,n+1})
\le
(1-c_sh_s)\mathcal V_s(z_{s,n})
+C_{s,1}h_s^2+h_sC_{s,2}\beta_{s,n}+\delta^{\mathrm{proj}}_{s,n,+}.
\]
Iterating this recursion gives
\begin{align*}
\mathcal V_s(z_{s,n})
&\le (1-c_sh_s)^n\mathcal V_s(z_{s,0})
+\sum_{j=0}^{n-1}(1-c_sh_s)^{n-1-j}\bigl(C_{s,1}h_s^2+h_sC_{s,2}\beta_{s,j}+\delta^{\mathrm{proj}}_{s,j,+}\bigr)
\\
&\le (1-c_sh_s)^n\mathcal V_s(z_{s,0})
+ C_{s,1}h_s^2\sum_{j=0}^{n-1}(1-c_sh_s)^j
+ h_s C_{s,2}\sup_{0\le j<n}\beta_{s,j}\sum_{j=0}^{n-1}(1-c_sh_s)^j
\\
&\qquad + \sum_{j=0}^{n-1}(1-c_sh_s)^{n-1-j}\delta^{\mathrm{proj}}_{s,j,+}.
\end{align*}
Using the geometric-series bound
\[
\sum_{j=0}^{n-1}(1-c_sh_s)^j\le \frac{1}{c_sh_s},
\]
we obtain
\[
\mathcal V_s(z_{s,n})
\le
(1-c_sh_s)^n\mathcal V_s(z_{s,0})
+\frac{C_{s,1}}{c_s}h_s
+\frac{C_{s,2}}{c_s}\sup_{0\le j<n}\beta_{s,j}
+\sum_{j=0}^{n-1}(1-c_sh_s)^{n-1-j}\delta^{\mathrm{proj}}_{s,j,+},
\]
which is exactly \eqref{eq:dp_discrete_contraction}. This proves Theorem~\ref{lem:dp_discrete_frozen_stage_contraction}.

\subsection{Approximate equilibration of a terminated stage}
\label{app:dp_approximate_equilibration}

We next justify Proposition~\ref{prop:dp_frozen_stage_conv}. Let the stage terminate at $z_s^+=(q_s^+,p_s^+)$ with
\[
\|\nabla_q U_{\tau,s}(q_s^+)\|+\|p_s^+\|\le \eta_s.
\]
By strong convexity of $U_{\tau,s}$ on $\mathcal U_s^q$,
\[
\|q_s^+-q_s^\star\|
\le \mu_s^{-1}\|\nabla_q U_{\tau,s}(q_s^+)\|
\le \mu_s^{-1}\eta_s.
\]
On the other hand, the discrete defect theorem gives
\[
\mathcal V_s(z_s^+)\le (1-c_sh_s)^{N_s}\mathcal V_s(z_{s,0})+\mathfrak e_s = \mathfrak E_s .
\]
By the norm equivalence \eqref{eq:app_modified_lyapunov_equiv}, this implies
\[
\|q_s^+-q_s^\star\|+\|p_s^+\|\le C\sqrt{\mathfrak E_s}.
\]
The residual condition separately yields
\(\|q_s^+-q_s^\star\|+\|p_s^+\|\le C'\eta_s\) under local strong convexity/nondegeneracy. Combining these two estimates gives
\[
\|q_s^+-q_s^\star\|+\|p_s^+\|
\le C_{\mathrm{eq},s}(\eta_s+\sqrt{\mathfrak E_s}),
\]
for a local constant $C_{\mathrm{eq},s}>0$. This proves \eqref{eq:dp_stage_terminal_neighborhood}.

\subsection{Comparison between the shaped and original potentials}
\label{app:dp_shaping_comparison}

By definition,
\[
U_{\tau,s}(q)=f_\tau(q)+\Delta^{\tau}_{\psi,s}(q;\bar q_s,m_s^0,\mu_s).
\]
Assumption~(C2) implies
\begin{equation}
\label{eq:app_shaping_distortion_bound}
|U_{\tau,s}(q)-f_\tau(q)|\le \rho_s
\qquad\text{on the reachable neighborhood of stage }s.
\end{equation}
Applying this bound at $q_s^+$ and $q_s^\star$, and using the local Lipschitz continuity of both $U_{\tau,s}$ and $f_\tau$ together with Proposition~\ref{prop:dp_frozen_stage_conv}, yields constants $C_1,C_2,C_3>0$ such that
\begin{equation}
\label{eq:app_task_potential_compare}
f_\tau(q_s^+)
\le
f_\tau(q_s^\star)+C_1\eta_s+C_2\rho_s+C_3\sqrt{\mathfrak E_s}.
\end{equation}
This is the key bridge from stagewise equilibration in the shaped potential to strict improvement for the original objective.

\subsection{Proof of hybrid strict improvement}
\label{app:dp_hybrid_improvement}

The monotonicity statement follows immediately from the acceptance rule:
\[
f_\tau(\widehat q_{s+1})
=
\min\{f_\tau(\widehat q_s),f_\tau(\widetilde q_s)\}
\le f_\tau(\widehat q_s).
\]
Since $f_\tau$ is bounded below on the relevant sublevel set, the sequence $\{f_\tau(\widehat q_s)\}$ converges.

For the strict-improvement part, suppose that the update performed at the end of stage $s$ produces a next-stage frozen equilibrium $q_{s+1}^\star$ such that
\[
f_\tau(q_{s+1}^\star)\le f_\tau(\widehat q_{s+1})-\delta_{s+1},
\qquad \delta_{s+1}>0.
\]
Applying \eqref{eq:app_task_potential_compare} to stage $s+1$ yields
\[
f_\tau(q_{s+1}^+)
\le
f_\tau(q_{s+1}^\star)+C_1\eta_{s+1}+C_2\rho_{s+1}+C_3\sqrt{\mathfrak E_{s+1}}.
\]
Combining the two displays gives
\[
f_\tau(q_{s+1}^+)
\le
f_\tau(\widehat q_{s+1})-\delta_{s+1}+C_1\eta_{s+1}+C_2\rho_{s+1}+C_3\sqrt{\mathfrak E_{s+1}}.
\]
Finally, because $f_\tau(\widetilde q_{s+1})\le f_\tau(q_{s+1}^+)$ and the accepted iterate after executing stage $s+1$ is
\[
\widehat q_{s+2}\in \arg\min\{f_\tau(\widehat q_{s+1}),f_\tau(\widetilde q_{s+1})\},
\]
we obtain
\[
f_\tau(\widehat q_{s+2})
\le
f_\tau(\widehat q_{s+1})-\delta_{s+1}+C_1\eta_{s+1}+C_2\rho_{s+1}+C_3\sqrt{\mathfrak E_{s+1}},
\]
which is exactly \eqref{eq:dp_strict_improvement}. This proves Proposition~\ref{prop:dp_hybrid_memory_improvement}.

\subsection{Proof of nonconvex convergence under sufficient decrease and relative error}
\label{app:dp_nonconvex_convergence}

Assume the sufficient decrease and relative error conditions \eqref{eq:dp_sufficient_decrease}--\eqref{eq:dp_relative_error}. Since the accepted values are monotone and bounded below, they converge. Summing \eqref{eq:dp_sufficient_decrease} from some index $s_0$ to $N$ gives
\[
a\sum_{s=s_0}^N \|\widehat q_{s+1}-\widehat q_s\|^2
\le
f_\tau(\widehat q_{s_0})-f_\tau(\widehat q_{N+1}) + \sum_{s=s_0}^N \varepsilon_s.
\]
The right-hand side remains bounded as $N\to\infty$, because $f_\tau(\widehat q_s)$ converges and $\sum_s \varepsilon_s<\infty$. Hence
\[
\sum_{s=s_0}^\infty \|\widehat q_{s+1}-\widehat q_s\|^2<\infty,
\qquad\text{so in particular}\qquad
\|\widehat q_{s+1}-\widehat q_s\|\to0.
\]
Now let $\bar q$ be an accumulation point, obtained along a subsequence $\widehat q_{s_j}\to \bar q$. Since $\varepsilon_s\to 0$ and $\|\widehat q_{s+1}-\widehat q_s\|\to0$, the relative-error bound implies
\[
\operatorname{dist}(0,\partial f_\tau(\widehat q_{s_j+1}))\to0.
\]
Because $\widehat q_{s_j+1}-\widehat q_{s_j}\to 0$, the shifted subsequence $\widehat q_{s_j+1}$ converges to the same limit $\bar q$. Closedness of the limiting subdifferential therefore gives
\[
0\in \partial f_\tau(\bar q),
\]
which proves \eqref{eq:dp_critical_cluster_point}. If, in addition, $f_\tau$ satisfies the Kurdyka--\L{}ojasiewicz property on the relevant sublevel set, then the standard KL argument yields finite length of the accepted sequence and convergence of the full sequence to a single critical point.

\subsection{Proof of the stochastic-oracle defect bound}
\label{app:proof_stochastic_oracle_defect}
\label{app:dp_stochastic_extension}

We prove Proposition~\ref{prop:dp_stochastic_oracle_defect}.  Suppose the
frozen local update admits the conditional one-step expansion
\[
\mathcal V_s(z_{s,n+1})
\le
(1-c_sh_s)\mathcal V_s(z_{s,n})
+h_s\langle \zeta_{s,n},\xi_{s,n}\rangle
+C_{s,1}h_s^2
+C h_s^2\|\xi_{s,n}\|_{W_\Xi}^2
+h_s\widetilde C_{s,2}\beta_{s,n}
+\delta^{\mathrm{proj}}_{s,n},
\]
where \(\zeta_{s,n}\) is \(\mathcal F_{s,n}\)-measurable and bounded. Taking
conditional expectation and using
\[
\mathbb E[\xi_{s,n}\mid \mathcal F_{s,n}]=0,
\qquad
\mathbb E[\|\xi_{s,n}\|_{W_\Xi}^2\mid \mathcal F_{s,n}]
\le
\operatorname{tr}(W_\Xi\Xi_{s,n}),
\]
yields
\[
\mathbb E[\mathcal V_s(z_{s,n+1})\mid \mathcal F_{s,n}]
\le
(1-c_sh_s)\mathcal V_s(z_{s,n})
+C_{s,1}h_s^2
+Ch_s^2\operatorname{tr}(W_\Xi\Xi_{s,n})
+h_s\widetilde C_{s,2}\beta_{s,n}
+\mathbb E[\delta^{\mathrm{proj}}_{s,n}\mid \mathcal F_{s,n}],
\]
which is \eqref{eq:dp_stochastic_discrete_ineq} after absorbing constants into
\(\widetilde C_{s,1}\).  Hence the stochastic term changes the one-step defect
by \(\mathcal O(h_s^2\operatorname{tr}(W_\Xi\Xi_{s,n}))\) and the accumulated
stage floor by
\(\mathcal O(h_s\sup_n\operatorname{tr}(W_\Xi\Xi_{s,n}))\).

\subsection{How the finite-budget proposition uses the preceding estimates}
\label{app:dp_budget_dependency_map}

The finite-budget proposition in the main text uses the previous appendix
results as a dependency chain rather than as independent assumptions.  We make
this chain explicit here.

First, Theorem~\ref{lem:dp_canonical_frozen_stage_rate} gives the
continuous-time mechanism for local convergence of a frozen stage.  Because
damping acts in the momentum channel, the proof uses the modified Lyapunov
function $\mathcal V_s$ with a cross term $e^\top p$.  This supports the
claim that, once a stage action has fixed a shaped potential
$U_{\tau,s}=f_\tau+U_s^{\rm shp}$, the induced port-Hamiltonian dynamics can
contract toward the local frozen equilibrium $q_s^\star$.

Second, Theorem~\ref{lem:dp_discrete_frozen_stage_contraction} transfers this
mechanism to the implemented semi-implicit rollout.  It shows that the terminal
Lyapunov error is bounded by a decaying initialization term plus the explicit
defect floor
\[
    \mathfrak e_s
    =
    \frac{C_{s,1}}{c_s}h_s
    +
    \frac{C_{s,2}}{c_s}\sup_{0\le n<N_s}\beta_{s,n}
    +
    \sum_{n=0}^{N_s-1}
    (1-c_sh_s)^{N_s-1-n}\delta^{\rm proj}_{s,n,+}.
\]
Thus the local solver success probability $r_s$ in
Proposition~\ref{prop:outer_budgeted_progress} can be interpreted as the
probability that the stage budget, step size, oracle quality, and projection
defects make this terminal error small enough to enter the desired basin.

Third, Proposition~\ref{prop:dp_frozen_stage_conv} converts the terminal
Lyapunov error into an approximate-equilibration estimate:
\[
    \|q_s^+-q_s^\star\|+\|p_s^+\|
    \le
    C_{\rm eq,s}\bigl(\eta_s+\sqrt{\mathfrak E_s}\bigr).
\]
This is the local bridge from ``the rollout dissipates energy'' to ``the stage
actually terminates near the frozen-stage equilibrium.''  It is the main
technical justification behind the $r_s$ factor.

Fourth, Proposition~\ref{prop:dp_hybrid_memory_improvement} explains how a
good frozen-stage equilibrium gives improvement in the original task objective,
not only in the shaped potential.  If the planner--memory update creates a
next-stage equilibrium $q_{s+1}^{\star}$ satisfying
\[
    f_\tau(q_{s+1}^{\star})
    \le
    f_\tau(\widehat q_{s+1})-\delta_{s+1},
\]
then after executing the next stage,
\[
    f_\tau(\widehat q_{s+2})
    \le
    f_\tau(\widehat q_{s+1})
    -\delta_{s+1}
    +C_1\eta_{s+1}+C_2\rho_{s+1}+C_3\sqrt{\mathfrak E_{s+1}}.
\]
Therefore a strict event-level improvement of size $\delta$ is certified
whenever the proposed lower basin has margin larger than the residual,
shaping-distortion, and numerical-defect terms.  In the main proposition this
combined margin is denoted by $\Delta_s$.

Finally, Proposition~\ref{prop:dp_stochastic_oracle_defect} modifies the same defect floor by an
additional covariance-weighted term of order
\[
    h_s \sup_n \operatorname{tr}(W_\Xi\Xi_{s,n})
\]
after summing over a stage.  Hence stochastic gradients do not change the
logical structure of the finite-budget proposition; they change the achievable
value of $r_s$ by increasing the stage error budget.

In summary, $\vartheta_s$ is the probability that the slow planner--memory mechanism
selects a useful basin, while $r_s$ is the probability that the local IDA-PBC
rollout reaches that basin with a small enough residual and defect floor.  The
preceding theorems do not prove a universal positive lower bound on $\vartheta_s$;
that quantity depends on the learned planner and the task distribution.  They
do, however, give explicit sufficient conditions under which a useful proposal
is converted into a certified best-so-far improvement.

\subsection{Proof of Proposition~\ref{prop:outer_budgeted_progress}}
\label{app:dp_finite_budget_extension}

Let $A_s$ denote the event that stage $s$ both proposes a useful lower basin
and reaches it with sufficiently small residual and defect floor.  More
concretely, $A_s$ is the event that the planner--memory update proposes a
frozen-stage equilibrium satisfying
\[
    f_\tau(q_{s+1}^{\star})
    \le
    F_s-\delta-\Delta_s,
\]
and that the local rollout enters the corresponding basin within the allocated
budget.  The preceding dependency map explains how the local contraction,
discrete-defect, approximate-equilibration, and hybrid-improvement results make
this event sufficient for
\[
    F_{s+1}\le F_s-\delta .
\]
By the definitions of $\vartheta_s$ and $r_s$,
\[
    \mathbb P(A_s\mid \mathcal F_s)\ge \vartheta_s r_s .
\]
If no stage up to $S-1$ succeeds, then $F_S>F_0-\delta$.  Therefore
\[
\mathbb P(F_S>F_0-\delta)
\le
\mathbb P\Bigl(\bigcap_{s=0}^{S-1} A_s^c\Bigr).
\]
Conditioning recursively on the stage history gives
\[
\mathbb P\Bigl(\bigcap_{s=0}^{S-1} A_s^c\Bigr)
\le
\prod_{s=0}^{S-1}(1-\vartheta_s r_s),
\]
which proves the product bound in
\eqref{eq:outer_no_improvement_probability}.  If $\vartheta_s r_s\ge\rho>0$
uniformly, then
\[
\prod_{s=0}^{S-1}(1-\vartheta_s r_s)
\le
(1-\rho)^S
\le
\mathrm e^{-\rho S},
\]
which proves \eqref{eq:outer_exponential_budget_progress}.


\end{document}